\documentclass[10pt,twocolumn,letterpaper]{article}

\usepackage[pagenumbers]{cvpr} %

\newcommand{\methodname}{ArtReveal}
\newcommand{\anchorfm}{Exterior-Anchored Flow Matching}
\newcommand{\stateest}{Latent Articulation State Estimation}
\newcommand{\pixalign}{Kinematics-Aware Pixel-to-3D Alignment}
\newcommand{\papertitle}{
  \methodname: Articulated Object Generation with Progressive Updates to Newly Revealed Structure%
}
\definecolor{rankfirst}{HTML}{F4A6A6}
\definecolor{ranksecond}{HTML}{F8CFA0}
\definecolor{rankthird}{HTML}{FBEBA6}
\newcommand{\rfirst}[1]{\cellcolor{rankfirst}\textbf{#1}}
\newcommand{\rsecond}[1]{\cellcolor{ranksecond}#1}
\newcommand{\rthird}[1]{\cellcolor{rankthird}#1}

\usepackage{xcolor}
\usepackage{booktabs}   %
\usepackage{placeins}
\usepackage{colortbl}   %
\usepackage{graphicx}   %
\usepackage{tikz}       %
\usetikzlibrary{arrows.meta,calc}
\usepackage{fontawesome5} %
\definecolor{panelA}{HTML}{EEF0F3}   %
\definecolor{panelB}{HTML}{E6F0FA}   %
\definecolor{panelC}{HTML}{FCEEE3}   %
\definecolor{accA}{HTML}{5A6270}
\definecolor{accB}{HTML}{2F6FB0}
\definecolor{accC}{HTML}{C9601F}
\definecolor{frozen}{HTML}{8A8F98}
\definecolor{slotA}{HTML}{4C78A8}    %
\definecolor{slotB}{HTML}{F2A33A}    %
\definecolor{slotC}{HTML}{54A24B}    %

\tikzset{
  panel/.style={rounded corners=6pt, fill=#1, draw=none},
  ptitle/.style={anchor=north west, font=\footnotesize\bfseries, text=#1, inner sep=0pt},
  box/.style={draw=#1!75!black, fill=#1!15, rounded corners=3pt, align=center,
              font=\scriptsize, inner sep=2pt, line width=0.4pt},
  lbl/.style={font=\tiny, align=center, text=black!75, inner sep=1pt},
  arr/.style={-{Stealth[length=3pt,width=2.5pt]}, line width=0.5pt, draw=black!65},
  bus/.style={line width=0.8pt, draw=#1, -{Stealth[length=3.5pt,width=3pt]}},
  hop/.style={preaction={draw=white, line width=2.8pt}},
}

\newcommand{\figimg}[4]{%
  \IfFileExists{#4}%
    {\includegraphics[width=#1,height=#2,keepaspectratio]{#4}}%
    {\fcolorbox{black!30}{white}{\parbox[c][#2][c]{#1}{\centering\tiny\itshape\color{black!55}#3}}}}

\newcommand{\latcube}[4]{%
  \begin{scope}[shift={(#1,#2)}]
    \pgfmathsetmacro\lcH{#3/2}\pgfmathsetmacro\lcD{0.35*#3}
    \filldraw[draw=#4!60!black, fill=#4!45, line width=0.3pt]
      (-\lcH,\lcH) -- ++(\lcD,\lcD) -- ++(#3,0) -- (\lcH,\lcH) -- cycle;
    \filldraw[draw=#4!60!black, fill=#4!65, line width=0.3pt]
      (\lcH,-\lcH) -- ++(\lcD,\lcD) -- ++(0,#3) -- (\lcH,\lcH) -- cycle;
    \filldraw[draw=#4!60!black, fill=#4!20, line width=0.3pt] (-\lcH,-\lcH) rectangle (\lcH,\lcH);
  \end{scope}}

\newcommand{\anchorcube}[3]{%
  \latcube{#1}{#2}{#3}{frozen}%
  \begin{scope}[shift={(#1,#2)}]
    \pgfmathsetmacro\lcH{#3/2}\pgfmathsetmacro\lcC{#3/4}
    \pgfmathsetseed{7}
    \foreach \i in {0,...,3}{\foreach \j in {0,...,3}{%
      \ifnum\i<2\relax
        \ifnum\j<3\relax \def\lcCol{slotA!60}\else\def\lcCol{slotB!70}\fi
      \else
        \pgfmathsetmacro\lcG{int(25+50*rnd)}\edef\lcCol{black!\lcG}%
      \fi
      \filldraw[fill=\lcCol, draw=white, line width=0.2pt]
        (-\lcH+\i*\lcC,-\lcH+\j*\lcC) rectangle ++(\lcC,\lcC);}}
  \end{scope}}

\definecolor{cvprblue}{rgb}{0.21,0.49,0.74}
\usepackage[pagebackref,breaklinks,colorlinks,allcolors=cvprblue]{hyperref}

\def\confName{CVPR}
\def\confYear{2027}

\title{\papertitle}

\author{Jim Yun-Jin Li$^{1}$ \quad Benjamin Bogenberger$^{1}$ \quad Niklas Schl\"{u}ter$^{1}$ \quad Angela P. Schoellig$^{1}$\\
$^{1}$Learning Systems and Robotics Lab, Technical University of Munich\\
{\tt\small \{jim.li, benjamin.bogenberger, niklas.schlueter, angela.schoellig\}@tum.de}\\
{\small\url{https://learnsyslab.github.io/artreveal/}}
}

\begin{document}
\maketitle

\begin{abstract}
    Generating articulated 3D assets is essential for embodied agents, real-to-sim modeling, and AR/VR content creation.
    Humans guess how the parts of a closed object move and what is inside, and revise the guess as interaction reveals more. An agent needs the same, a model that generates an initial articulated asset from limited visual evidence and updates its geometry, appearance, and joints as new evidence arrives.
    Existing generators produce an asset once, and neither 3D inpainting nor multi-state reconstruction refines it progressively.
    We present \methodname{}, to our knowledge the first feed-forward model for articulated object generation with progressive updates.
    From a single rest-state image and part masks, it generates an initial asset with part-wise flow matching. Given a later image with parts opened, it estimates how far each part was opened, starts generation from the asset's visible exterior to preserve it, and samples image features at the projections of the posed parts so that the revealed interior matches the observation.
    Since no benchmark evaluates this setting, we further build RevealTrain and RevealBench with an agentic modeling pipeline. \methodname{} outperforms eight baselines on single-image generation and, on RevealBench, reaches an F-score of 86.7 on the revealed geometry against 79.8 for the strongest baseline while correcting earlier joint predictions. We also show updates on real objects captured with a phone.
\end{abstract}

\begin{figure}[t]
  \centering
  \includegraphics[width=\linewidth]{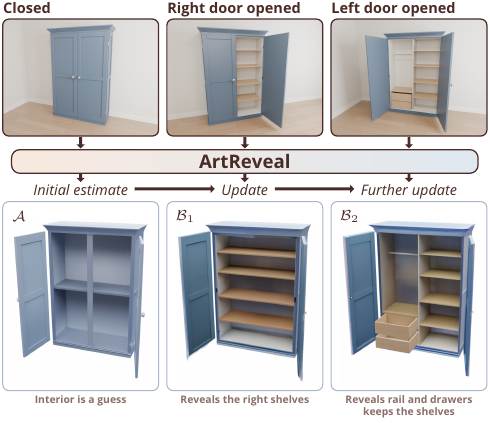}
  \vspace{-4mm}
  \caption{\textbf{\methodname{} updates an articulated asset as its interior is observed.} From a single image of the closed wardrobe, \methodname{} generates an initial estimate $\mathcal{A}$ whose interior is plausible but unobserved. Each close-up taken after a door is opened updates the previous result. $\mathcal{B}_1$ reveals the shelves behind the right door, and $\mathcal{B}_2$ reveals the rail and drawers behind the left door while keeping the shelves of $\mathcal{B}_1$. Doors and drawers are shown open to visualize the interiors.}
  \label{fig:teaser}
  \vspace{-3mm}
\end{figure}

\section{Introduction}
\label{sec:intro}

Articulated objects such as cabinets, ovens, dishwashers, and refrigerators are common in households, and interacting with them requires knowing not only their geometry but also how their parts move.
Creating such digital assets can support interactive content creation in VR/AR, physics-based simulation~\cite{xiang2020sapien, jin2026artvip}, robot learning~\cite{Chen-RSS-24,Villasevil-RSS-24}, and robot manipulation~\cite{chen2025g3flow,gauthier2026percepttwin,xiang2026real}.
Recent advances in 3D generative models enable the creation of high-fidelity assets from a single image~\cite{xiang2025structured,xiang2026native,chen2026sam,huang2026cupid,zadaianchuk2026reconstruction}.
However, these assets are typically monolithic and lack the part structure and joint kinematics needed for fine-grained interaction.
To address this limitation, methods for articulated asset generation combine part geometry with predicted joint parameters. These methods either assemble retrieved part meshes~\cite{liu2025singapo,zheng2026pwm,wu2026dipo,le2025articulate} or generate part geometry using large pretrained 3D models~\cite{chen2025artilatent,cao2026physx,cao2025physxanything,wu2026urdf,liu2026pact}.

Despite these advances, a single rest-state image leaves much of the interior of an object unobserved~\cite{lee2026articulated}.
Cabinets with nearly identical exteriors may contain shelves or hidden drawers, or be entirely empty.
A generative model must therefore rely on learned priors to infer a plausible interior, which may not match the actual object (Fig.~\ref{fig:teaser}, $\mathcal{A}$).
Furthermore, methods that ``imagine'' the open state with image or video generators~\cite{he2026spark,boudjoghra2026unfoldart} face the same ambiguity, since these views are generated rather than observed.
Methods using real observations across multiple articulation states provide evidence about the actual interior, but they assume that these ``privileged'' observations are all available at once~\cite{liu2023paris,wu2026dipo,li2026art,yuan2025larm,wang2026artic, chen2025freeart3d}.
In practice, such evidence arrives progressively, as a robot interacts with an object~\cite{yan2023interaction,nie2023structure} or a person captures it with a phone or egocentric glasses~\cite{engel2023project,kong2025aria}.
This requires a model that generates an initial articulated asset from the first observation and refines it with every new one, giving a robot the best estimate so far and capturing digital twins progressively.

To this end, we present \methodname{}. From a single rest-state image and 2D part masks, it first generates an initial articulated asset through part-wise flow matching on structured 3D latents~\cite{xiang2025structured}.
To update the asset from later observations, \methodname{} addresses three key challenges.

\textbf{\textit{How can we preserve previously observed exterior structure?}}
Regenerating from a new observation can change regions that earlier observations already settled. We introduce exterior-anchored flow matching, which starts the flow in the visible exterior from the current asset's latents instead of noise. Generation therefore begins at the current exterior, so its geometry and appearance stay close to it while the hidden structure is free to change.

\textbf{\textit{How can we determine which parts have moved and by how much?}}
We introduce latent articulation state estimation, which predicts the state of each part from the cached part tokens of the asset and the new observation, without an external pose estimator. The joints of the moved parts are then re-predicted on the updated asset, which corrects their type and axis.

\textbf{\textit{How can we update the interior to match the newly observed structure?}}
We introduce kinematics-aware pixel-to-3D alignment, which registers the asset to metric camera poses for direct use in robotics, poses each part's latents in its state, and samples image features at their projections.

To train and evaluate progressive updates, we build an agentic pipeline that crafts families of articulated objects that share an exterior but differ inside, yielding the \textit{RevealTrain} dataset, and the \textit{RevealBench} benchmark. \methodname{} outperforms eight baselines in rest-state articulated object generation on PartNet-Mobility (PM)~\cite{xiang2020sapien} and ACD~\cite{iliash2026s2o}, and on RevealBench it recovers revealed interiors more accurately than reconstruction, generation, and 3D inpainting baselines while correcting earlier joint predictions.
Our contributions are summarized as follows:
\begin{itemize}
      \item We propose \methodname{}, to our knowledge the first feed-forward model for single-image articulated object generation with progressive updates to newly revealed structure.
      \item We develop an update mechanism combining exterior-anchored flow matching, latent articulation state estimation, and kinematics-aware pixel-to-3D alignment to reconcile new observations with the existing asset.
      \item We introduce \textit{RevealBench} and the \textit{RevealTrain} dataset, built by an agentic pipeline, for evaluating and training progressive updates.
\end{itemize}

\section{Related Works}
\label{sec:related_works}

\subsection{3D Asset and Part Generation}
Image-to-3D generation has moved from lifting generated multi-view images into 3D~\cite{liu2024syncdreamer,liu2023zero,long2024wonder3d} to native 3D generation, supported by compact and expressive latent representations such as vector sets~\cite{zhang3dshape2vecset2023} and structured 3D latents~\cite{xiang2025structured} trained on large-scale 3D asset collections~\cite{Deitke2023Objaverse,deitke2023objaversexl}.
These advances have enabled high-fidelity asset generation from a single image or text description~\cite{clay2024zhang,xiang2025structured,li2025triposg,zhao2025hunyuan3d,xiang2026native}.
Recent methods such as CUPID~\cite{huang2026cupid} and Pixal3D~\cite{li2026pixal3d} further introduce explicit 2D-3D alignment to improve consistency with the conditioning image.
Part-aware 3D generative models extend this paradigm by representing an object as separate components, either by generating multiple parts jointly~\cite{lin2026partcrafter,yang2025omnipart} or by decomposing an existing shape~\cite{liu2025partfield}. However, these methods do not explicitly model articulation or joint kinematics, and geometry hidden behind a closed door must still be inferred from learned priors.

\subsection{Articulated Object Modeling}
Early generative models learn priors over the geometry and kinematics of articulated objects and generate plausible assets unconditionally, from structural representations, or from text~\cite{liu2025survey,lei2023nap,liu2024cage,gao2025meshart,su2025artformer}, but modeling a specific object requires grounding it in visual evidence.

\noindent\textbf{Optimization and multi-stage pipeline.}
Optimization-based methods fit geometry and motion parameters to multi-view observations across articulation states, using neural fields~\cite{liu2023paris} or 3DGS~\cite{guo2025articulatedgs,yu2025artgs}, while FreeArt3D~\cite{chen2025freeart3d} optimizes both by score distillation from a frozen TRELLIS~\cite{xiang2025structured}.
Multi-stage pipelines combine pretrained VLMs and image or video generators.
Articulate-Anything~\cite{le2025articulate} combines mesh retrieval with a VLM actor--critic loop, Articraft~\cite{zhou2026articraft} lets an LLM coding agent write programs that build the asset, and Kinematify~\cite{wang2026kinematify} combines VLM predictions, search, and geometric refinement.
SPARK~\cite{he2026spark}, DreamArt~\cite{lu2025dreamart}, Rest2Art~\cite{lee2026articulated}, and UnfoldArt~\cite{boudjoghra2026unfoldart} use generated images or videos as extra observations to infer articulation or expose hidden geometry.

\noindent\textbf{Feed-forward generation.}
Feed-forward methods predict the asset directly from the input without per-object optimization, which makes inference efficient.
Multi-state methods take observations of several articulation states, all provided before inference.
DIPO~\cite{wu2026dipo} conditions on a rest-state and articulated-state image pair,
and ART~\cite{li2026art}, LARM~\cite{yuan2025larm}, and Artic-O~\cite{wang2026artic} use sparse multi-view images of two or more states.
Single-image methods infer the asset from one rest-state image.
SINGAPO~\cite{liu2025singapo} and PWM-ArtGen~\cite{zheng2026pwm} predict structure and articulation and retrieve part meshes from a database, while others generate geometry natively by finetuning TRELLIS~\cite{xiang2025structured,chen2025artilatent,cao2026physx,cao2025physxanything} or TripoSG~\cite{li2025triposg,wu2026urdf}.
Particulate~\cite{li2026particulate} articulates a generated mesh, and PAct~\cite{liu2026pact} conditions on 2D part masks through learned part-identity embeddings.
However, all these methods generate the asset once and do not update it when later observations reveal hidden structure or contradict the inferred articulation. Interactive perception reconstructs the interiors that a robot exposes by manipulating an object from RGB-D observations~\cite{yan2023interaction,nie2023structure, jiang2022ditto, hsu2023ditto}, rather than generating the asset from a single image.

\begin{figure*}[t]
  \centering
  \includegraphics[width=\textwidth]{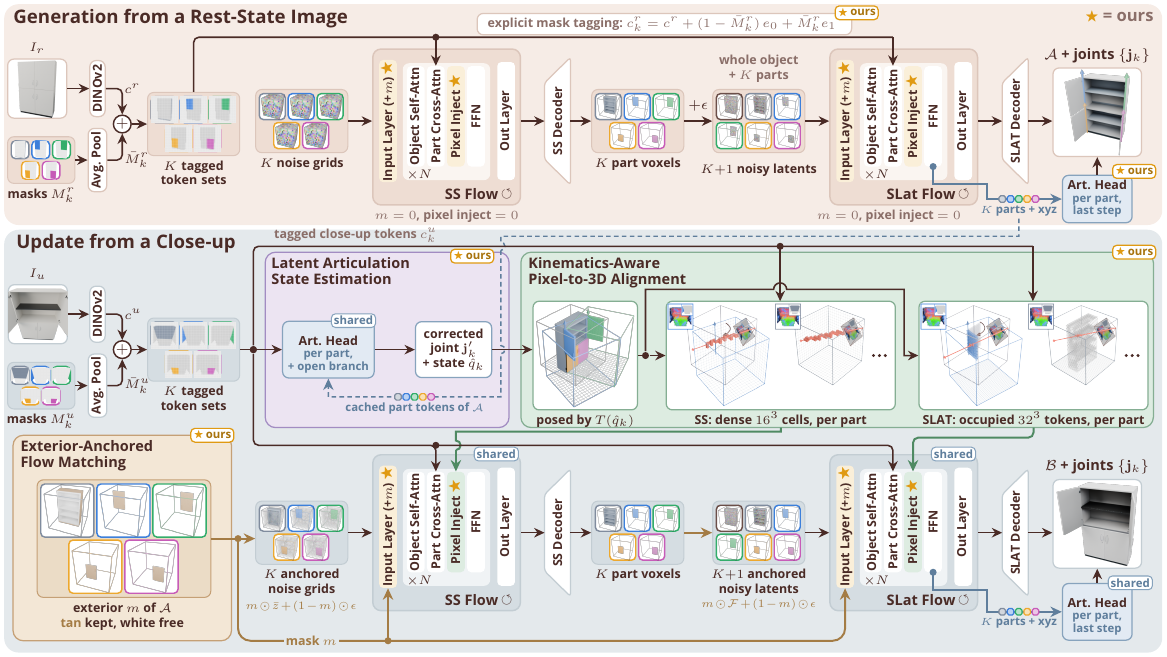}
  \vspace{-4mm}
  \caption{\textbf{Overview of \methodname{}.}
    (Top) DINOv2 tokens are tagged with each part mask, the SS and SLat flows generate all parts of $\mathcal{A}$ jointly, and the articulation head (\cref{fig:arthead}) predicts the joints from the last-step part tokens.
    (Bottom) Updates from a close-up use the same network.
    \textbf{\stateest{}} predicts corrected joints $\mathbf{j}'_k$ and states $\hat{q}_k$ from the cached part tokens of $\mathcal{A}$,
    \textbf{\pixalign{}} poses the moved parts by $T(\hat{q}_k)$ and samples close-up features at their projections,
    and \textbf{\anchorfm{}} starts both flows from the visible exterior of $\mathcal{A}$ (\cref{eq:anchor}) to produce the updated asset $\mathcal{B}$.}
  \label{fig:overview}
  \vspace{-3mm}
\end{figure*}

\subsection{3D Inpainting and Editing}
TRELLIS~\cite{xiang2025structured} supports training-free local 3D editing by adapting RePaint~\cite{lugmayr2022repaint} to its two-stage latent generation. VoxHammer~\cite{li2026voxhammer} edits locally in TRELLIS latent space by inverting an existing asset and reusing cached latents and attention features to preserve unedited regions during denoising. Points-to-3D~\cite{xia2026points} incorporates visible-region point clouds into the TRELLIS sparse structure representation and learns to inpaint the remaining geometry.

\noindent\textbf{Ours.}
Prior articulated object methods generate an asset once, and inpainting and editing methods modify static assets within a given region. \methodname{} instead learns from the visible exterior which content of an articulated asset to keep, grounds the regenerated geometry in a new observation, and refines the joints, so that a single-image estimate can be updated as observations arrive progressively.

\section{Method}
\label{sec:method}
\noindent\textbf{Problem formulation.}
Given a rest-state image $I_r$ of an object and 2D masks $\mathcal{M}_r=\{M_k^r\}_{k=1}^{K}$ of its $K$ visible parts, \methodname{} generates an articulated asset $\mathcal{A}$ with part-level geometry, appearance, and articulation, namely a part type and a joint with its type, axis, origin, and range.
When a later close-up $I_u$ with known camera pose shows the object in a different articulation state, 
together with the masks $\mathcal{M}_u$ of its $K^\prime$ visible parts, the same network updates $\mathcal{A}$ to a new asset $\mathcal{B}$ and estimates the articulation state $\hat q_k$ of parts in $\mathcal{A}$. 
$\mathcal{B}$ then becomes the current asset for subsequent observations.
The masks resolve the ambiguity of the part decomposition, such as one wide door or two narrow ones.
Sharing indices across images, the masks identify previously generated and newly revealed parts. Parts outside the close-up remain represented in $\mathcal{B}$.
A drawer behind a closed door, for example, gets a new index once $I_u$ reveals it.
They can also come from our automatic mask pipeline.

\noindent\textbf{Overview.}
We implement \methodname{} as a part-wise extension of TRELLIS that generates all parts jointly (\cref{sec:prelim,sec:base}).
An update estimates the part states with \stateest{} (\cref{sec:state}) and regenerates the asset in its rest frame, where \anchorfm{} preserves the observed exterior (\cref{sec:anchor}) and \pixalign{} grounds the revealed interior in $I_u$ (\cref{sec:align}), as shown in \cref{fig:overview}.

\subsection{Preliminaries on TRELLIS}
\label{sec:prelim}
TRELLIS~\cite{xiang2025structured} is a native-3D generative model pretrained on large-scale 3D asset collections and provides us with a strong shape prior to transfer to articulated assets.
TRELLIS generates a 3D asset in two stages, each with a rectified flow transformer conditioned on DINOv2~\cite{oquab2024dinov} image tokens $c$ through cross-attention.
In the first stage, the flow model $v_\theta^{\mathrm{ss}}$ denoises a compact sparse structure (SS) latent
$z \in \mathbb{R}^{16^3 \times 8}$, which is decoded by the sparse structure VAE decoder $\mathcal{D}_{\mathrm{ss}}$ into a binary occupancy grid $O \in \{0,1\}^{64^3}$.
In the second stage, another rectified flow model $v_\theta^{\mathrm{SLat}}$ denoises an 8-dimensional feature $\mathbf{f}_i$ at each occupied voxel $\mathbf{p}_i$, forming the structured latent (SLat) sparse tensor
$\mathcal{F} = \{(\mathbf{p}_i,\mathbf{f}_i)\}_{i=1}^{L}$,
where $L$ is the number of occupied voxels.
The SLat is decoded by the VAE decoders $\mathcal{D}_{\mathrm{SLat}}$ into 3D Gaussians~\cite{kerbl20233d} or a mesh~\cite{shen2023flexible}.
Both flow models are trained with conditional flow matching (CFM)~\cite{lipman2022flow}. A clean latent $x_0$, \ie $z$ or $\mathcal{F}$, is interpolated with Gaussian noise $\epsilon$ as $x_t = (1-t)\,x_0 + t\,\epsilon$, and $v_\theta$ is
trained to predict the velocity.
\begin{equation}
  \mathcal{L}_{\mathrm{CFM}} = \mathbb{E}_{t,\,x_0,\,\epsilon}\,\big\|\,v_\theta(x_t, t, c) - (\epsilon - x_0)\,\big\|_2^2
  \label{eq:cfm}
\end{equation}

\subsection{Part-Wise Articulated Object Generation}
\label{sec:base}
TRELLIS generates an object as one monolithic shape, with no understanding of the parts it consists of. An articulated object, however, must represent every movable part as a separate shape, while all parts together still form a plausible assembly.
We therefore adopt a part-wise paradigm that generates all parts of an object jointly in both the sparse structure (SS) and the structured latent (SLat) stages (\cref{fig:overview}, top).
Our part-wise SS flow model $v_\theta^{\mathrm{ss}}$ denoises one latent $z_k$ per part.
We concatenate the latents of all parts of an object $[\,z_1;\dots;z_K\,]$ as input to the transformer, where each $z_k$ is projected into a sequence of tokens $h_k$.
The tokens $[\,h_1;\dots;h_K\,]$ of all parts are processed as one sequence in self-attention, so that all parts attend to each other in every block (\cref{eq:block}).
Our part-wise SLat flow model $v_\theta^{\mathrm{SLat}}$ then denoises the structured latent $\mathcal{F}_k$ of every part on the voxels decoded from the generated $z_k$. Following the spatially-conditioned part synthesis stage of OmniPart~\cite{yang2025omnipart}, it also denoises the structured latent $\mathcal{F}_0$ of the whole object on the union of these voxels, and processes the tokens of $\mathcal{F}_0, \mathcal{F}_1, \dots, \mathcal{F}_K$ as one sequence in self-attention.
We initialize $v_\theta^{\mathrm{ss}}$ from TRELLIS and $v_\theta^{\mathrm{SLat}}$ from the pretrained model of OmniPart, inheriting the object-level prior of the former and the part-level prior of the latter.

\noindent\textbf{Binding parts to masks.}
The central problem when generating all parts jointly is how each latent $z_k$ knows which part of the image it should generate.
PAct~\cite{liu2026pact} addresses this problem by using one shared map of part IDs added to the image tokens, such that the $k$-th latent is expected to generate the region labelled $k$.
We find this binding to be weak in practice, as the evaluated model sometimes duplicates or misses parts.
To solve this problem, we instead condition every part on the image directly by tagging each image token with the binary mask of that part.
Let $c^r \in \mathbb{R}^{N\times d}$ be the $N$ DINOv2 patch tokens of $I_r$ and $\bar{M}_k^r \in [0,1]^{N}$ be the binary mask $M_k^r$ average-pooled to the token grid.
Each part $k$ receives its own tagged tokens $c_k^r$ (\cref{eq:tag}) and attends to it through cross-attention, while all parts attend to each other through self-attention (\cref{eq:block}).
\begin{equation}
  c_k^r = c^r + (1-\bar{M}_k^r)\,e_0 + \bar{M}_k^r\,e_1
  \label{eq:tag}
\end{equation}
\begin{equation}
  \begin{aligned}
    [\,h_1;\dots;h_K\,] & \leftarrow \mathrm{SelfAttn}\big([\,h_1;\dots;h_K\,]\big), \\
    h_k                 & \leftarrow \mathrm{CrossAttn}\big(h_k,\, c_k^r\big)
  \end{aligned}
  \label{eq:block}
\end{equation}
Here, $e_0, e_1 \in \mathbb{R}^{d}$ are learned embeddings, zero-initialized so that training starts from the behavior of the pretrained model.
Since $v_\theta^{\mathrm{ss}}$ carries no part index, the tag is the only signal that distinguishes the parts, and the model cannot fall back on their order.
$v_\theta^{\mathrm{SLat}}$ uses the same tagging, with $\mathcal{F}_0$ tagged by the foreground mask.

\begin{figure}[t]
  \centering
  \includegraphics[width=\columnwidth]{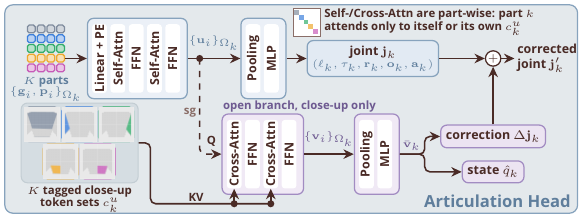}
  \vspace{-4mm}
  \caption{\textbf{Articulation head.} The last-step part tokens $\mathbf{g}_i$ at their token positions $\mathbf{p}_i$ are refined per part into $\mathbf{u}_i$ and pooled into the joint $\mathbf{j}_k$ (\cref{eq:head}). With a close-up, the open branch (purple) cross-attends to the tagged close-up tokens $c_k^u$ and predicts the correction $\Delta\mathbf{j}_k$ and the state $\hat{q}_k$ (\cref{eq:open}).}
  \label{fig:arthead}
  \vspace{-3mm}
\end{figure}

\noindent\textbf{Articulation prediction.}
Since the parts attend to each other in $v_\theta^{\mathrm{SLat}}$, the tokens of each part already encode its shape and its relation to the rest of the object.
An articulation head (Art.\ Head as shown in~\cref{fig:overview}) therefore predicts the articulation of every part directly from these tokens.
Let $\Omega_k$ be the tokens of part $k$, where token $\mathbf{h}_{k,i}$ sits at the center $\mathbf{p}_i$ of its cell on the $16^3$ (SS) or $32^3$ (SLat) token grid, and $\mathbf{g}_i$ be its output from the last transformer block of $v_\theta^{\mathrm{SLat}}$ in the last sampling step, which encodes the generated geometry and appearance of the part around $\mathbf{p}_i$.
The head embeds each token as $W\mathbf{g}_i + \mathrm{PE}(\mathbf{p}_i)$ and refines the tokens of every part with two transformer blocks whose self-attention is restricted to that part, which gives the part tokens $\mathbf{u}_i$ as shown in~\cref{fig:arthead}.
It then predicts the part type, joint type, and range of part $k$ from its pooled feature, and the axis origin and direction by attending to its tokens (\cref{eq:head}).
\begin{equation}
  \begin{aligned}
    (\ell_k,\tau_k,\mathbf{r}_k) & = \mathrm{MLP}\big(\big[\,\mathrm{mean}_{i \in \Omega_k}\mathbf{u}_i\,;\,\mathrm{max}_{i \in \Omega_k}\mathbf{u}_i\,\big]\big), \\
    \mathbf{o}_k                 & = \textstyle\sum_{i\in\Omega_k}\alpha_i\,\mathbf{p}_i,                                                              \\
    \mathbf{a}_k                 & = \mathrm{normalize}\big(W_a \textstyle\sum_{i\in\Omega_k}\beta_i\,\mathbf{u}_i\big)
  \end{aligned}
  \label{eq:head}
\end{equation}
Here, $\ell_k \in \mathbb{R}^{6}$ and $\tau_k \in \mathbb{R}^{4}$ are the logits of the part type (base, door, drawer, shelf, tray, knob) and of the joint type (fixed, revolute, prismatic, continuous).
The remaining outputs describe how the part moves.
Its joint axis passes through the origin $\mathbf{o}_k \in \mathbb{R}^{3}$ along the unit direction $\mathbf{a}_k \in \mathbb{R}^{3}$, and $\mathbf{r}_k \in \mathbb{R}^{2}$ holds the lower and upper limits of the joint.
Together, these outputs form the joint $\mathbf{j}_k = (\ell_k, \tau_k, \mathbf{r}_k, \mathbf{o}_k, \mathbf{a}_k)$ of part $k$.
Finally, $\alpha_i$ and $\beta_i$ are softmax weights over the tokens of the part, computed from two learned scores $\mathbf{w}_\alpha^{\top}\mathbf{u}_i$ and $\mathbf{w}_\beta^{\top}\mathbf{u}_i$, so that the origin is a weighted average of the token positions of the part.
The head is trained with \cref{eq:artloss}, where $^{*}$ marks the ground truth, $\mathrm{CE}$ is the cross-entropy, and $\mathcal{P}$ holds the movable parts.
\begin{equation}
  \begin{gathered}
    \mathcal{L}_{\mathrm{art}} = \tfrac{1}{K}\textstyle\sum_{k}[\mathrm{CE}(\ell_k,\ell_k^{*}) + \mathrm{CE}(\tau_k,\tau_k^{*})] \\
    + \tfrac{1}{|\mathcal{P}|}\textstyle\sum_{k\in\mathcal{P}}[\|\mathbf{o}_k{-}\mathbf{o}_k^{*}\|^2 + \|\mathbf{r}_k{-}\mathbf{r}_k^{*}\|^2 + (1{-}\mathbf{a}_k^{\top}\mathbf{a}_k^{*})]
  \end{gathered}
  \label{eq:artloss}
\end{equation}
\subsection{\stateest{}}
\label{sec:state}
To pose the asset as shown in a close-up $I_u$, we estimate the state of every part from the cached part tokens of $\mathcal{A}$, natively without an external pose estimator or optimization. Newly revealed parts have no cached tokens and are assumed closed.
We add an open branch to the articulation head, illustrated by the purple block in \cref{fig:arthead}.
It lets the tokens of part $k$ attend to the close-up tokens $c_k^u$, tagged as in \cref{eq:tag} with a separate pair of learnable embeddings, and predicts a correction of $\mathbf{j}_k$ together with the state $\hat{q}_k$ of the part as shown in \cref{eq:open}.
\begin{equation}
  \begin{aligned}
    \bar{\mathbf{v}}_k & = \mathrm{MLP}\big(\big[\,\mathrm{mean}_{i \in \Omega_k}\mathbf{v}_i\,;\,\mathrm{max}_{i \in \Omega_k}\mathbf{v}_i\,\big]\big), \\
    \mathbf{j}'_k      & = \mathrm{sg}(\mathbf{j}_k) + W_{\Delta}\,\bar{\mathbf{v}}_k, \quad \hat{q}_k = \mathbf{w}_q^{\top}\bar{\mathbf{v}}_k
  \end{aligned}
  \label{eq:open}
\end{equation}
The tokens $\mathbf{v}_i$ are obtained from $\mathrm{sg}(\mathbf{u}_i)$ by two transformer blocks, each with a cross-attention to $c_k^u$ as shown in \cref{fig:arthead}, and $\mathrm{sg}$ denotes the stop-gradient.
The state prediction $\hat{q}_k$ is a normalized opening value in the units of the range $\mathbf{r}_k$, where $\hat{q}_k = 0$ means that the part is in the rest state. The corrected axis direction is renormalized to unit length.
$W_{\Delta}$ and $\mathbf{w}_q$ are zero-initialized, so that the branch starts from $\mathbf{j}'_k = \mathbf{j}_k$ and $\hat{q}_k = 0$. We apply stop-gradients here so that training the branch does not change $\mathbf{j}_k$.
The moved parts are posed with $\mathbf{j}'_k$ and $\hat{q}_k$ during the update (\cref{sec:align}), and their joints are re-predicted on $\mathcal{B}$ afterwards.
We train with \cref{eq:openloss}, applying \cref{eq:artloss} to visible corrected joints. $\mathcal{L}_{\mathrm{state}}$ sums the mean squared state errors computed separately over moved and closed movable parts.
\begin{equation}
    \mathcal{L}_{\mathrm{open}}
    = \mathcal{L}_{\mathrm{art}}(\mathbf{j}'_k)
    + \mathcal{L}_{\mathrm{state}}
    \label{eq:openloss}
  \end{equation}

\subsection{\anchorfm{}}
\label{sec:anchor}
An update should keep what has already been observed, and change the asset only where the close-up shows something new.
We therefore adopt an anchoring strategy inspired by Points-to-3D~\cite{xia2026points}, which starts its SS flow from an encoded point cloud prior. Instead of passing $\mathcal{A}$ to the network as an additional condition, we start the flow from the exterior surface of $\mathcal{A}$ as an anchor.
For every part, an exterior mask $m_k$ marks the voxels of the decoded rest-state geometry of $\mathcal{A}$ that are visible from any of 100 directions around the asset (\cref{sec:supp_model}), whether or not $I_r$ shows them.
For a chained update, such as $\mathcal{B}_1 \rightarrow \mathcal{B}_2$ in \cref{fig:teaser}, $m_k$ is extended by the voxels that earlier close-ups saw, with the asset posed in their predicted states from~\cref{sec:state}, and the movable parts that the close-up does not move are kept.
In the SS stage, $m_k$ marks the cells of the $16^3$ grid of part $k$ that contain a visible voxel, and the anchor $\bar{z}_k$ is obtained by re-encoding the visible voxels of the part with the SS VAE encoder of TRELLIS.
In the SLat stage, $m_k$ marks the voxels of the updated structure that coincide with visible voxels of $\mathcal{A}$, and the anchor is the structured latent $\mathcal{F}_k$ of $\mathcal{A}$ at these voxels.
We replace the noise in the anchored cells with the anchor $x_{\mathcal{A}}$, \ie $\bar{z}_k$ or $\mathcal{F}_k$, as shown in \cref{eq:anchor}.
\begin{equation}
  \epsilon' = m_k \odot x_{\mathcal{A}} + (1-m_k) \odot \epsilon
  \label{eq:anchor}
\end{equation}
We train with \cref{eq:cfm}, using $\epsilon'$ in place of $\epsilon$ in both $x_t$ and the target velocity, so that sampling starts from $\epsilon'$ at $t=1$. With $m_k = 0$, $\epsilon' = \epsilon$, and the initial generation from $I_r$ is the special case without an anchor.
To tell the network which cells hold the anchor, we additionally feed $m_k$ as an extra input channel in the SS stage and through a zero-initialized linear layer after the input layer in the SLat stage.

\subsection{\pixalign{}}
\label{sec:align}
The anchor keeps the exterior, but the newly revealed interior must come from $I_u$, and at the right place.
Cross-attention to the close-up tokens tells the network what $I_u$ shows but leaves the 2D-3D correspondence implicit~\cite{li2026pixal3d,huang2026cupid}, so a revealed shelf can appear at the wrong height.
We therefore give every 3D location the image features at its projection in $I_u$.
This requires the camera of $I_u$ in the frame of $\mathcal{A}$.
We assume metric camera poses of $I_r$ and $I_u$, \eg from the SLAM of a robot or the AR tracking of a phone, and register $\mathcal{A}$ to them at metric scale with a 7-DoF similarity transform $(s,\mathbf{R},\mathbf{t})$, initialized by aligning the part silhouettes of $\mathcal{A}$ with the masks and refined by PnP over both photos on RoMa~\cite{edstedt2024roma} matches between renders of $\mathcal{A}$ and the photos, using only matches in the parts that have not moved in $I_u$ (details in \cref{sec:supp_inference}).
With $\mathcal{A}$ registered, every voxel can be projected into $I_u$, but the moved parts are not in their rest state anymore. We therefore pose the canonical positions of all voxels of every moved part by the rigid transform $T(\hat{q}_k)$ about its corrected joint $\mathbf{j}'_k$ (\cref{eq:open}), as illustrated in the alignment panel of \cref{fig:overview}.
Every token $i$ of part $k$ then samples its tagged close-up tokens $c_k^u$ (\cref{eq:tag}) at its projection, and every transformer block adds the sampled feature after its cross-attention, as shown in \cref{eq:pix}.
\begin{equation}
  \mathbf{s}_i = c_k^u\big[\,\pi_u\big(T(\hat{q}_k)\,\mathbf{p}_i\big)\big], \qquad \mathbf{h}_{k,i} \leftarrow \mathbf{h}_{k,i} + W_b\,\mathbf{s}_i
  \label{eq:pix}
\end{equation}
Here, $\pi_u$ projects into $I_u$ with its registered camera, $[\cdot]$ is bilinear sampling on the token grid of $c_k^u$, and $W_b$ is a zero-initialized linear layer of block $b$.
Projecting voxel centers into an image for an explicit 2D-3D correspondence has been used for image-LiDAR place recognition~\cite{li2025vxp} and to ground 3D generation~\cite{huang2026cupid,li2026pixal3d}, and we extend it to parts that have moved between $I_r$ and $I_u$ by posing their voxels by their estimated state to sample features, while the asset is still generated in its canonical rest state.

\section{Experiments}
\label{sec:exp}
\begin{figure*}[t]
  \centering
  \setlength{\tabcolsep}{0pt}
  \renewcommand{\arraystretch}{0}
  \newcommand{\qr}[1]{\raisebox{-0.5\height}{\includegraphics[width=0.080\textwidth]{figures/qual_rest/#1.jpg}}}
  \newcommand{\qh}[1]{\scriptsize\renewcommand{\arraystretch}{0.9}\begin{tabular}[b]{@{}c@{}}#1\end{tabular}}
  \newcommand{\ql}[1]{\rotatebox[origin=c]{90}{\scriptsize #1}\hspace{2pt}}
  \begin{tabular}{c@{\hspace{2pt}}cccccccc@{\hspace{2pt}}c@{\hspace{2pt}}c}
    \qh{Input}         & \qh{SINGAPO}                & \qh{Articulate-                                                                                                                                                                                                                                                                                             \\Anything} & \qh{PWM-\\ArtGen} & \qh{URDF-\\Anything+} & \qh{PhysX-\\Anything} & \qh{TRELLIS+\\Particulate} & \qh{SPARK} & \qh{PAct} & \qh{Ours} & \qh{GT} \\[3pt]
    \qr{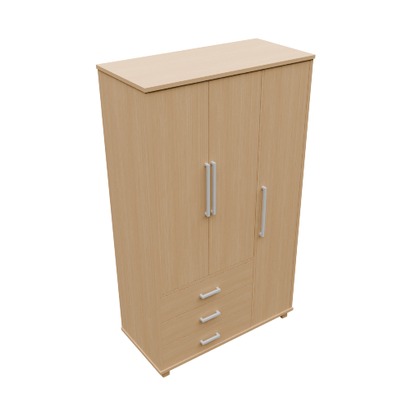} & \qr{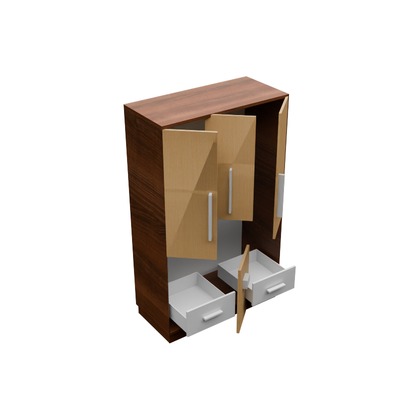} & \qr{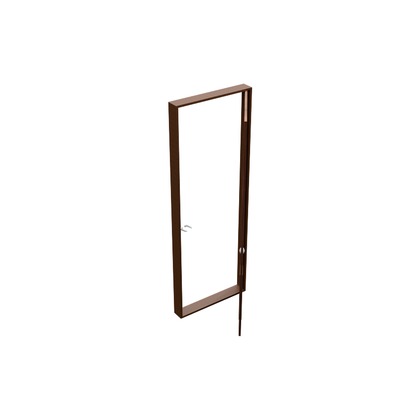} & \qr{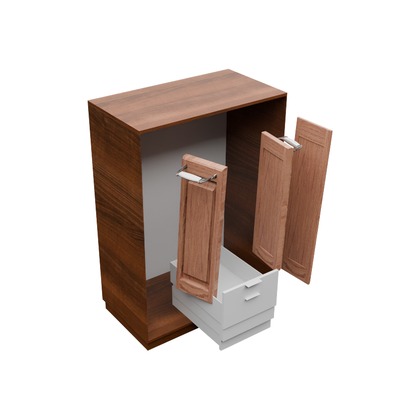} & \qr{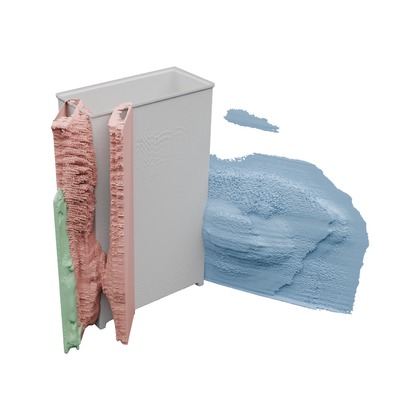} & \qr{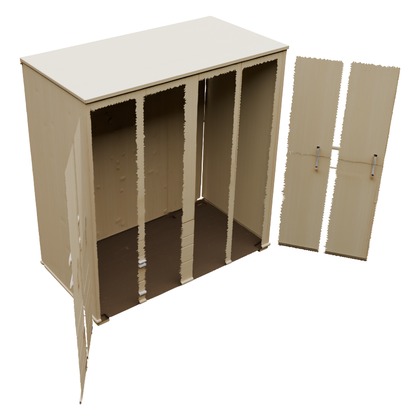} & \qr{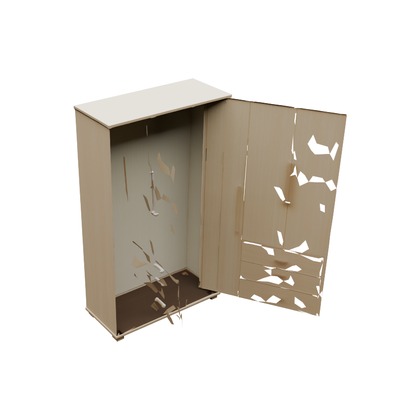} & \qr{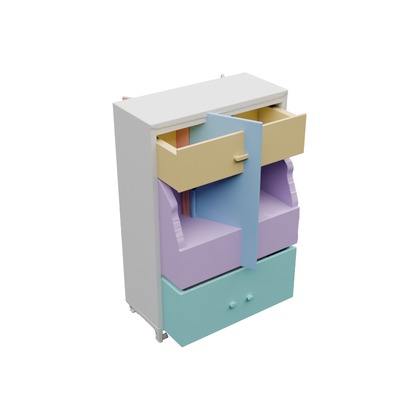} & \qr{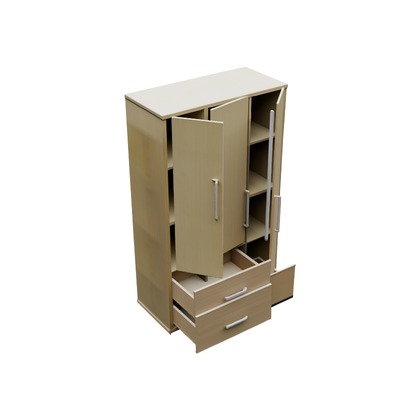} & \qr{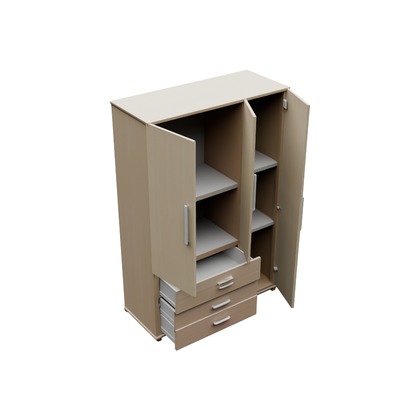} & \qr{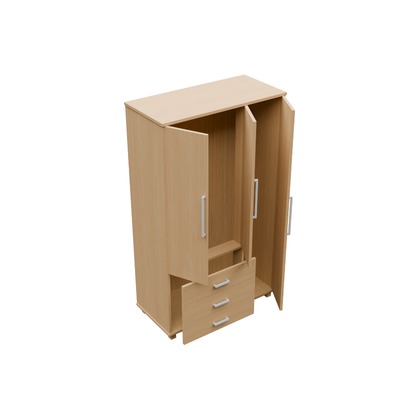} \\[2pt]
    \qr{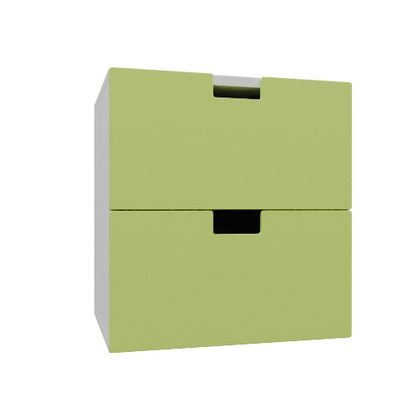}   & \qr{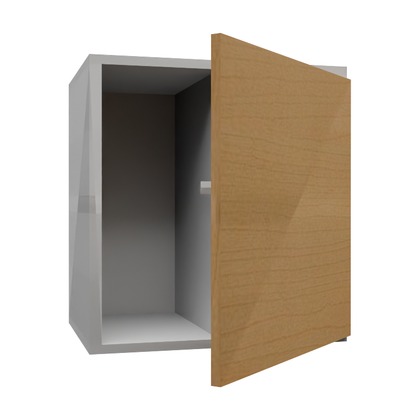}   & \qr{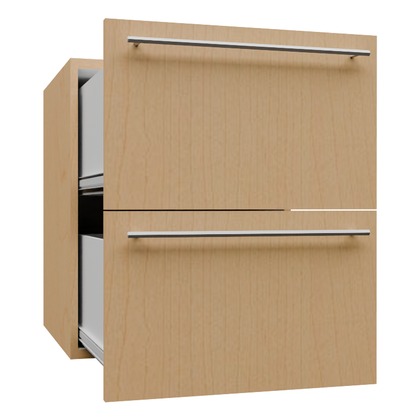}   & \qr{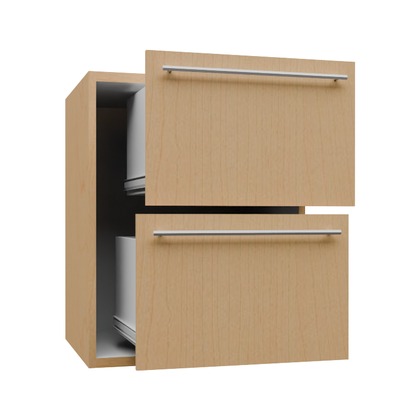}   & \qr{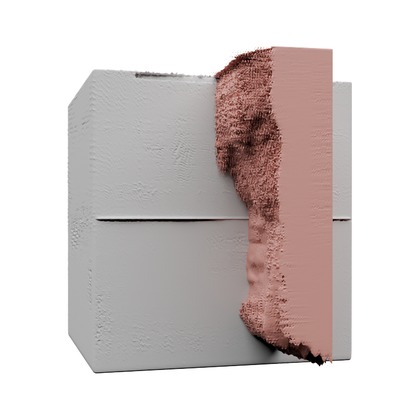}   & \qr{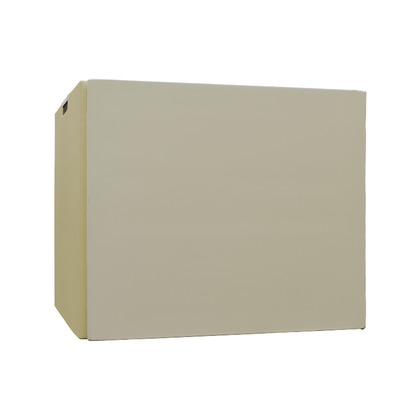}   & \qr{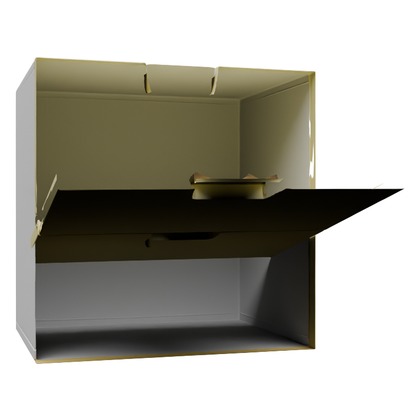}   & \qr{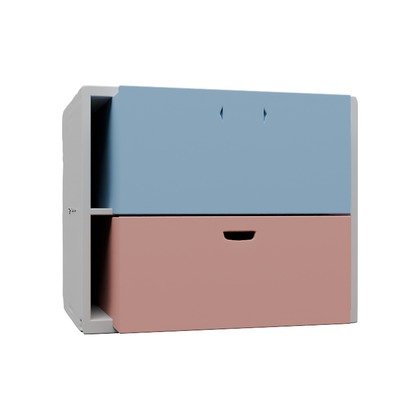}   & \qr{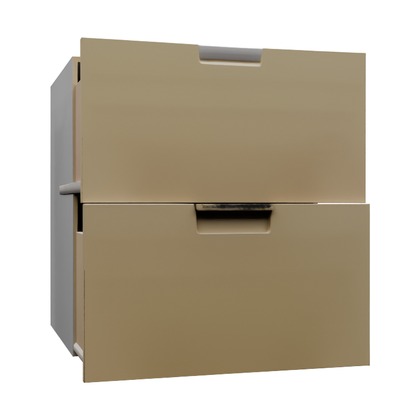}   & \qr{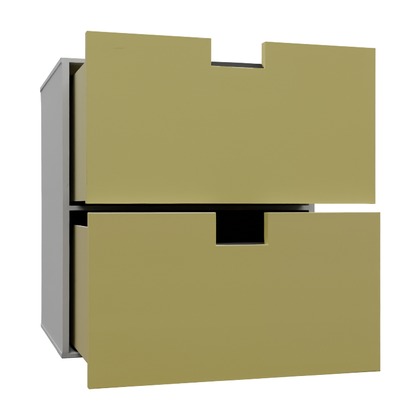}   & \qr{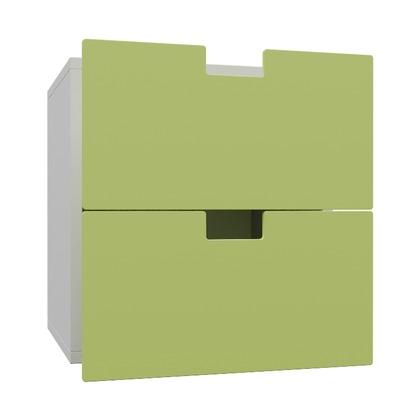}   \\
  \end{tabular}
  \vspace{-3mm}
  \caption{Rest-state generation on ACD~\cite{iliash2026s2o} with all movable parts opened to 70\% of their range. SPARK and URDF-Anything+ have no texture (one color per part). All models are trained without using any ACD data. }
  \label{fig:qual_rest}
  \vspace{-4mm}
\end{figure*}

\subsection{Evaluation Protocol and Baselines}
\label{sec:protocol}
\noindent\textbf{Implementation.}
We train in two phases and provide more details in \cref{sec:supp_impl}. For rest-state generation, we finetune the SS flow of TRELLIS~\cite{xiang2025structured} for 40k iterations and the SLat flow from OmniPart~\cite{yang2025omnipart} with LoRA~\cite{hu2022lowrank} for 80k iterations, together with the mask embeddings (\cref{eq:tag}) and the articulation head, on the SINGAPO training split of PM (\methodname{}-PM). For the update, we train both stages for another 20k iterations on PM and RevealTrain, adding the zero-initialized anchor and pixel-feature layers and the open branch of the articulation head. RevealTrain contains 600 objects modeled by our agentic pipeline, in families that share the exterior but differ inside (\cref{sec:supp_revealtrain}). On one consumer RTX 5090, generating an asset takes a median of 12~s on PM and 19~s on ACD, whose objects have 3.0 and 4.4 parts on average. On the timed RevealBench cases, with 6.5 parts on average, generating takes 32~s and updating 27~s after a 5~s registration, each before a texture bake for export (\cref{sec:supp_runtime}).

\noindent\textbf{Single-image rest-state generation.}
We evaluate on the SINGAPO~\cite{liu2025singapo} test splits of PM (77 objects) and ACD (135 objects), each rendered from two random views.
Following SINGAPO, dgIoU is $1-\mathrm{gIoU}$~\cite{rezatofighi2019generalized} between the predicted and ground-truth part bounding boxes and dCD is the Chamfer distance between the meshes, in the rest state (RS) and averaged over five states opening all parts (AS).
We compare with retrieval-based methods (SINGAPO, Articulate-Anything~\cite{le2025articulate}, PWM-ArtGen~\cite{zheng2026pwm}), image-only generation (URDF-Anything+~\cite{wu2026urdf}, PhysX-Anything~\cite{cao2025physxanything}, TRELLIS~\cite{xiang2025structured} with Particulate~\cite{li2026particulate}, SPARK~\cite{he2026spark}), and part-conditioned generation (PAct~\cite{liu2026pact}), all with released code and weights.
\methodname{}, PAct, and PWM-ArtGen also receive the part masks, SINGAPO and PWM-ArtGen the ground-truth part graph, PWM-ArtGen also the object category, and retrieval excludes test objects.
For a fair comparison, we evaluate \methodname{}-PM, trained in the first phase on the SINGAPO training split only, besides the final \methodname{} trained on PM and RevealTrain and used for updates.

\noindent\textbf{RevealBench.}
To evaluate updates, we introduce RevealBench.
Each case consists of a rest-state image and a close-up of the object with parts opened, both with 2D part masks and known camera poses and intrinsics.
RevealBench-ArtiVerse contains 163 storage furniture, table, and kitchen appliance objects from the ArtiVerse~\cite{iliash2026artiverse} test split, after manually removing objects with incorrect articulation, solid-filled parts, or broken textures.
RevealBench-Astra contains 34 object families built by the pipeline of RevealTrain from ABO~\cite{collins2022abo}, PM, and captured real furniture, each with four interior variants (136 objects) disjoint from RevealTrain in families and source objects.
Each object is rendered in Blender with two cases that open different parts, or the same parts to different extents, to 70 to 100\% of the joint range, giving 326 and 272 cases.
Updates use the given cameras and our registration (\cref{sec:align}). For scoring only, $\mathcal{A}$ and the updated asset are aligned to the ground truth and evaluated on the same ground-truth regions (\cref{sec:supp_revealbench}).
Unless noted, opened parts are posed with the ground-truth joint and state in \cref{tab:main_rb} only for scoring, so that geometry and appearance are scored separately from joint accuracy.
On the ground-truth surface, the revealed region is the part seen in the close-up but hidden from outside the closed object, and the unmoved exterior is the outside surface of parts that did not move.
For each region, we report the F-score at 2\% of the bounding-box diagonal and the Chamfer distance.
For appearance, we report LPIPS and DINO-I~\cite{ruiz2023dreambooth,li2026voxhammer}, the cosine similarity of DINOv2~\cite{oquab2024dinov} class embeddings, from two novel views on the revealed region (whole object in \cref{sec:supp_fullrb}).
We compare with training-free inpainting, RePaint~\cite{lugmayr2022repaint} and VoxHammer~\cite{li2026voxhammer} with or without its K/V replacement (+KV), on a base model trained with the same data (PM and RevealTrain), initialization, and number of steps as \methodname{}, but without the anchor, the pixel features, and the state estimation.
Conditioned on the close-up, they regenerate the voxels behind the opened part except the visible exterior of the initial asset and are given the opened part.
We further compare with DIPO~\cite{wu2026dipo}, given both images and the ground-truth part graph. FreeArt3D~\cite{chen2025freeart3d} receives six renders and the joint type, and LARM~\cite{yuan2025larm} and Artic-O~\cite{wang2026artic} both end states.
For the states, we compare with FoundationPose~\cite{wen2024foundationpose} and CAP-Net~\cite{huang2025cap}, given privileged inputs (\cref{tab:main_rb_state}).

\subsection{Single-Image Rest-State Generation Results}
\cref{tab:main_rest} and \cref{fig:qual_rest} compare single-image generation. Both \methodname{} models give the lowest errors on all reported metrics on both datasets.
On PM, SINGAPO, which retrieves parts from the same dataset, comes closest, and \methodname{}-PM still lowers RS-dCD from 0.49 to 0.39.
On ACD, which \methodname{}-PM never sees in training, the margin to the strongest baseline, PAct, grows to an RS-dgIoU of 32.84 against 44.19.
PAct conditions on the same part masks but sometimes duplicates or misses parts (\cref{sec:supp_binding}), while neither of our models duplicates any part.
The final model performs on par with \methodname{}-PM, so learning the update does not cost rest-state generation quality.

\begin{table}[t]
    \centering
    \caption{Rest-state generation on PM~\cite{xiang2020sapien} and ACD~\cite{iliash2026s2o} (SINGAPO~\cite{liu2025singapo} test split). Part-level dgIoU and dCD ($\times 100$) in the rest state (RS) and averaged over articulated states (AS). All metrics in \cref{sec:supp_rest}.}
    \vspace{-3mm}
    \label{tab:main_rest}
    \scriptsize
    \setlength{\tabcolsep}{1.6pt}
    \begin{tabular}{lcccccccc}
        \toprule
                                                     & \multicolumn{4}{c}{PM}                & \multicolumn{4}{c}{ACD}                                                                                                                                                                 \\
        \cmidrule(lr){2-5} \cmidrule(lr){6-9}
                                                     & \multicolumn{2}{c}{dgIoU$\downarrow$} & \multicolumn{2}{c}{dCD$\downarrow$} & \multicolumn{2}{c}{dgIoU$\downarrow$} & \multicolumn{2}{c}{dCD$\downarrow$}                                                                       \\
        \cmidrule(lr){2-3} \cmidrule(lr){4-5} \cmidrule(lr){6-7} \cmidrule(lr){8-9}
        Method                                       & RS                                    & AS                                  & RS                                    & AS                                  & RS              & AS              & RS             & AS             \\
        \midrule
        SINGAPO~\cite{liu2025singapo}                & \rthird{17.12}                        & \rthird{19.41}                      & \rthird{0.49}                         & \rthird{2.20}                       & 62.67           & 63.60           & 4.78           & 6.78           \\
        Artic.-Anything~\cite{le2025articulate}      & 28.93                                 & 31.22                               & 2.99                                  & 6.68                                & 76.19           & 77.19           & 7.06           & 10.46          \\
        PWM-ArtGen~\cite{zheng2026pwm}               & 48.58                                 & 49.58                               & 1.88                                  & 4.86                                & 67.29           & 68.43           & 6.20           & 7.23           \\
        \midrule
        URDF-Anything+~\cite{wu2026urdf}             & 61.38                                 & 62.05                               & 5.63                                  & 11.72                               & 75.05           & 75.60           & 7.75           & 11.52          \\
        PhysX-Anything~\cite{cao2025physxanything}   & 45.53                                 & 46.56                               & 8.98                                  & 14.64                               & 67.90           & 68.50           & 15.63          & 24.69          \\
        TRELLIS+Particulate~\cite{li2026particulate} & 51.94                                 & 52.93                               & 7.40                                  & 11.33                               & 75.69           & 76.29           & 8.98           & 13.58          \\
        SPARK~\cite{he2026spark}                     & 34.43                                 & 35.56                               & 1.96                                  & 11.45                               & 69.31           & 70.18           & 5.27           & 11.00          \\
        \midrule
        PAct~\cite{liu2026pact}                      & 18.65                                 & 20.53                               & 0.82                                  & 2.82                                & \rthird{44.19}  & \rthird{45.91}  & \rthird{2.91}  & \rthird{4.51}  \\
        \methodname{}-PM                             & \rsecond{15.01}                       & \rfirst{16.70}                      & \rsecond{0.39}                        & \rsecond{1.96}                      & \rfirst{32.84}  & \rfirst{35.14}  & \rsecond{1.19} & \rsecond{2.09} \\
        \methodname{} (Ours)                         & \rfirst{14.97}                        & \rsecond{16.92}                     & \rfirst{0.38}                         & \rfirst{1.79}                       & \rsecond{33.82} & \rsecond{35.57} & \rfirst{1.14}  & \rfirst{1.80}  \\
        \bottomrule
    \end{tabular}
    \vspace{-5mm}
\end{table}

\subsection{Reveal and Update Generation Results}
\begin{figure*}[t]
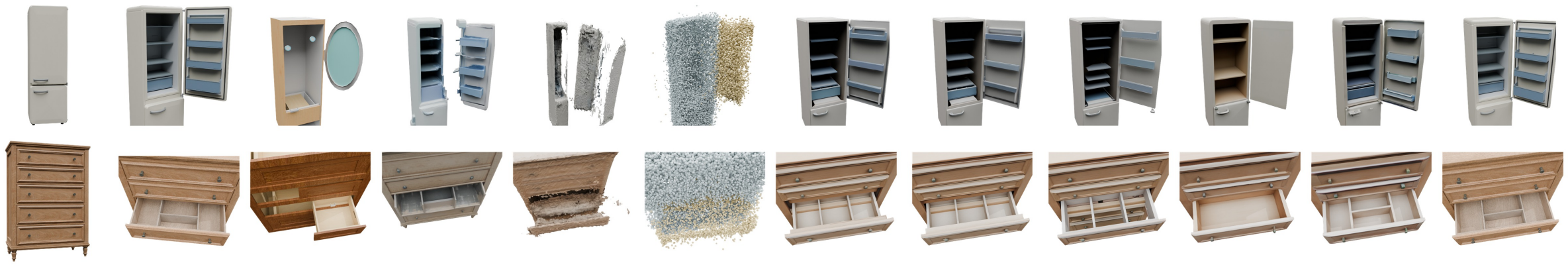

  \centering
  \setlength{\tabcolsep}{0.3pt}
  \renewcommand{\arraystretch}{0.5}
  \newcommand{\rbcell}[2]{\includegraphics[width=0.074\textwidth]{figures/revealbench/rb_#1_#2.jpg}}
  \begin{tabular}{cccccccccccc}
    \scriptsize Rest      & \scriptsize Close-up     & \scriptsize DIPO      & \scriptsize FreeArt3D$^\dagger$ & \scriptsize LARM$^\dagger$ & \scriptsize Artic-O$^\dagger$ & \scriptsize RePaint$^\ddagger$ & \scriptsize VoxHammer$^\ddagger$ & \scriptsize\shortstack{VoxHammer                                                                    \\+KV$^\ddagger$} & \scriptsize $\mathcal{A}$ (before) & \scriptsize Ours & \scriptsize Ground truth \\[2pt]
    \rbcell{fridge}{rest} & \rbcell{fridge}{closeup} & \rbcell{fridge}{dipo} & \rbcell{fridge}{freeart3d}      & \rbcell{fridge}{larm}      & \rbcell{fridge}{artico}       & \rbcell{fridge}{repaint}       & \rbcell{fridge}{vh}              & \rbcell{fridge}{vhkv}            & \rbcell{fridge}{A} & \rbcell{fridge}{ours} & \rbcell{fridge}{gt} \\
    \rbcell{drawer}{rest} & \rbcell{drawer}{closeup} & \rbcell{drawer}{dipo} & \rbcell{drawer}{freeart3d}      & \rbcell{drawer}{larm}      & \rbcell{drawer}{artico}       & \rbcell{drawer}{repaint}       & \rbcell{drawer}{vh}              & \rbcell{drawer}{vhkv}            & \rbcell{drawer}{A} & \rbcell{drawer}{ours} & \rbcell{drawer}{gt} \\
  \end{tabular}
  \vspace{-4mm}
  \caption{Updates on RevealBench.
    Top, a refrigerator with shelves and door bins seen only in the close-up.
    Bottom, a chest whose opened drawer holds a divider.
    $^\dagger$ privileged inputs, $^\ddagger$ opened part given.
    Artic-O outputs points (body blue, opened part gold).}
  \label{fig:qual_rb}
  \vspace{-4mm}
\end{figure*}

\cref{tab:main_rb} reports updates on RevealBench. The close-up raises the revealed F@2\% of the initial asset from 79.0 to 86.7, lowers its Chamfer distance from 0.0184 to 0.0130, and improves its appearance, while the unmoved exterior keeps its score (94.0 against 94.1).
Posed with our predicted joints and states instead of the ground truth for scoring, the revealed F@2\% is still 81.2, above all baselines and $\mathcal{A}$, which use the ground-truth pose for scoring as per the evaluation protocol, so the drop from 86.7 comes from articulation errors (type, axis, pivot, state), not from the generated interior.
Inpainting keeps the exterior almost as well, but barely improves the revealed geometry of its initial asset (F@2\% from 76.7 to at most 79.8). Constraining the exterior is therefore not enough, and the interior must be grounded in the close-up, as our anchor and pixel-aligned features do.
DIPO and the multi-state methods, although the latter receive several states, fall well behind on both regions.
In \cref{fig:qual_rb}, inpainting produces plausible compartments but misses the observed layout, such as the drawer divider, while \methodname{} reproduces what the close-up shows.
Seeing the opened part also corrects the joints (\cref{tab:main_rb_state}). The update raises type accuracy from 91.9 to 97.1\%, the highest of all methods, and lowers the mean axis error from 8.8 to 5.8\textdegree{}, close to the best inpainting baseline (5.3\textdegree{}), while the pivot error rises from 1.8 to 2.2\%, as for the inpainting baselines, and is 2.1\% with the ground-truth states as input.
From RGB alone, \methodname{} has the lowest mean revolute state error, below pose estimators with perfect depth. For prismatic parts, FoundationPose with the ground-truth mesh and rest pose of the part is more accurate.

\noindent\textbf{Progressive updates.} We chain up to four close-ups on the 10 RevealBench-Astra families that are not drawer-only (40 objects, \cref{tab:progressive}). Although trained only on single updates, chaining beats updating $\mathcal{A}$ with each close-up alone, with an F@2\% of 94.2 against 90.8 after the second close-up. Regions revealed earlier keep their gain until the last close-up, and the unmoved exterior does not drift (\cref{sec:supp_progressive}).

\begin{figure}[t]
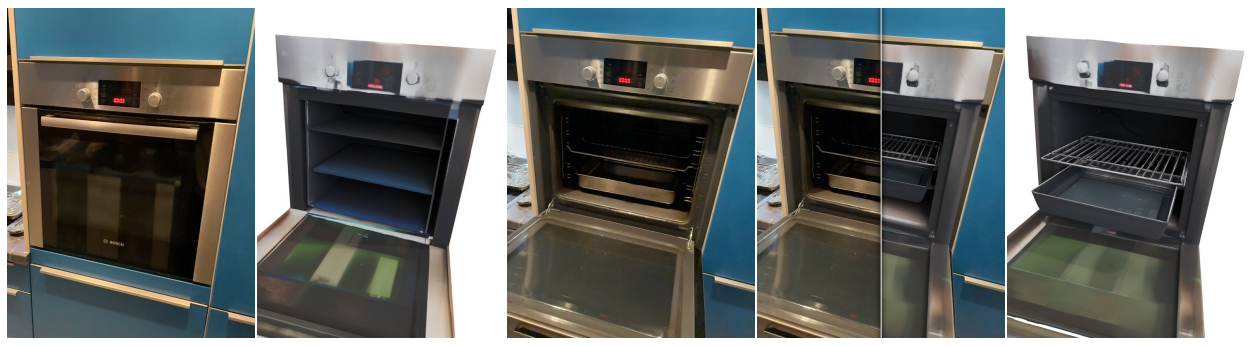

  \centering
  \setlength{\tabcolsep}{0.3pt}
  \renewcommand{\arraystretch}{0.5}
  \newcommand{\realcell}[2]{\includegraphics[width=0.196\columnwidth]{figures/real/real3_#1_#2.jpg}}
  \begin{tabular}{ccccc}
    \realcell{oven}{rest} & \realcell{oven}{A_open}          & \realcell{oven}{closeup} & \realcell{oven}{overlay}               & \realcell{oven}{B_new}           \\[1pt]
    \scriptsize Rest      & \scriptsize $\mathcal{A}$ opened & \scriptsize Close-up     & \scriptsize Close-up $|$ $\mathcal{B}$ & \scriptsize $\mathcal{B}$ opened
  \end{tabular}
  \vspace{-3mm}
  \caption{Update of a real oven from two phone photos, with ARKit poses and part masks from our automatic pipeline. The last column pulls out the revealed rack and pan.
  }
  \label{fig:real}
  \vspace{-2mm}
\end{figure}

\noindent\textbf{Real-world examples.}
\cref{fig:real} updates a real built-in oven from two phone photos, with camera poses from ARKit, registration as in \cref{sec:align}, and part masks from our automatic pipeline built on a VLM and SAM~3~\cite{carion2026sam}.
The update replaces the guessed empty cavity with the observed one and adds the movable rack and baking pan that only the close-up shows.
On RevealBench, the automatic masks reach a mean IoU of 0.79 and keep the identity of 95\% of the parts found in both photos (details in \cref{sec:supp_automask}).
On the cases where they succeed, ground-truth masks give a revealed F@2\% of 86.3 for the update and 78.7 for $\mathcal{A}$ (86.7 and 79.0 on all cases), and the automatic masks 83.6 and 77.1 (\cref{tab:main_rb}), so the update tolerates imperfect masks.

\begin{table}[t]
    \centering
    \caption{Updates on RevealBench (598 cases), appearance from two novel views. $^\dagger$ privileged inputs (several articulation states), $^\ddagger$ opened part given. FreeArt3D, LARM, and our update with automatic masks fail on 27.6\%, 2.3\%, and 9.4\% of the cases (excluded), ours mostly where the masks merge closed doors with the body. Gray rows (not ranked) show the initial asset or use ground-truth states as input, the predicted pose for scoring, or our automatic part masks as input.}
    \vspace{-2mm}
    \label{tab:main_rb}
    \footnotesize
    \setlength{\tabcolsep}{2pt}
    \begin{tabular}{lccccc}
        \toprule
                                                       & \multicolumn{2}{c}{Revealed geo.} & Unmoved                  & \multicolumn{2}{c}{Revealed app.}                                                     \\
        \cmidrule(lr){2-3} \cmidrule(lr){4-4} \cmidrule(lr){5-6}
        Method                                         & F@2\%$\uparrow$                   & CD$\downarrow$           & F@2\%$\uparrow$                   & LPIPS$\downarrow$       & DINO$\uparrow$          \\
        \midrule
        FreeArt3D$^\dagger$~\cite{chen2025freeart3d}   & 56.8                              & 0.0354                   & 76.1                              & 0.562                   & 0.653                   \\
        LARM$^\dagger$~\cite{yuan2025larm}             & 37.3                              & 0.0509                   & 30.6                              & 0.705                   & 0.223                   \\
        Artic-O$^\dagger$~\cite{wang2026artic}         & 69.2                              & 0.0246                   & 78.0                              & --                      & --                      \\
        \midrule
        DIPO~\cite{wu2026dipo}                         & 53.2                              & 0.0351                   & 78.3                              & 0.651                   & 0.565                   \\
        \midrule
        \textcolor{gray}{Inpainting base, before}      & \textcolor{gray}{76.7}            & \textcolor{gray}{0.0191} & \textcolor{gray}{93.5}            & \textcolor{gray}{0.435} & \textcolor{gray}{0.795} \\
        RePaint$^\ddagger$~\cite{lugmayr2022repaint}   & 76.9                              & 0.0183                   & \rthird{93.2}                     & \rthird{0.398}          & \rthird{0.877}          \\
        VoxHammer$^\ddagger$~\cite{li2026voxhammer}    & \rsecond{79.8}                    & \rsecond{0.0168}         & \rsecond{93.5}                    & \rsecond{0.388}         & \rsecond{0.885}         \\
        VoxHammer+KV$^\ddagger$~\cite{li2026voxhammer} & \rthird{77.3}                     & \rthird{0.0175}          & \rsecond{93.5}                    & 0.404                   & 0.862                   \\
        \midrule
        \textcolor{gray}{\methodname{} before update}  & \textcolor{gray}{79.0}            & \textcolor{gray}{0.0184} & \textcolor{gray}{94.1}            & \textcolor{gray}{0.430} & \textcolor{gray}{0.790} \\
        \methodname{} (Ours)                           & \rfirst{86.7}                     & \rfirst{0.0130}          & \rfirst{94.0}                     & \rfirst{0.350}          & \rfirst{0.903}          \\
        \midrule
        \multicolumn{6}{l}{\textit{Ablations}}                                                                                                                                                                \\
        Anchor only                                    & 82.6                              & 0.0154                   & 94.2                              & 0.367                   & 0.902                   \\
        Pixel-aligned only                             & 79.3                              & 0.0172                   & 88.9                              & 0.367                   & 0.902                   \\
        First variant only                             & 83.6                              & 0.0152                   & 93.6                              & 0.382                   & 0.860                   \\
        \textcolor{gray}{Ours, GT states as input}          & \textcolor{gray}{87.3}            & \textcolor{gray}{0.0128} & \textcolor{gray}{94.1}            & \textcolor{gray}{0.343} & \textcolor{gray}{0.906} \\
                \textcolor{gray}{Ours, predicted pose}         & \textcolor{gray}{81.2}            & \textcolor{gray}{0.0179} & \textcolor{gray}{93.6}            & \textcolor{gray}{--}    & \textcolor{gray}{--}    \\
                \textcolor{gray}{Auto masks, before}           & \textcolor{gray}{77.1}            & \textcolor{gray}{0.0193} & \textcolor{gray}{93.7}            & \textcolor{gray}{0.440} & \textcolor{gray}{0.787} \\
        \textcolor{gray}{Auto masks}                   & \textcolor{gray}{83.6}            & \textcolor{gray}{0.0147} & \textcolor{gray}{93.1}            & \textcolor{gray}{0.368} & \textcolor{gray}{0.879} \\
        \bottomrule
    \end{tabular}
    \vspace{-2mm}
\end{table}

\begin{table}[t]
    \centering
    \caption{Articulation of the opened parts on RevealBench (652 parts). Pivot error in \% of the diagonal, state error in degrees (revolute) and \% of travel (prismatic). $^\dagger$ perfect depth, ground-truth cameras, and for FoundationPose the ground-truth part mesh and mask. $^\ddagger$ opened part given. Gray rows are not ranked.}
    \vspace{-2mm}
    \label{tab:main_rb_state}
    \footnotesize
    \setlength{\tabcolsep}{2pt}
    \resizebox{\columnwidth}{!}{%
        \begin{tabular}{lccccc}
            \toprule
                                                                            & \multicolumn{3}{c}{Joint} & \multicolumn{2}{c}{State}                                                                                             \\
            \cmidrule(lr){2-4} \cmidrule(lr){5-6}
            Method                                                          & Type\,(\%)$\uparrow$      & Axis\,(\textdegree)$\downarrow$ & Pivot\,(\%)$\downarrow$ & Rev.\,(\textdegree)$\downarrow$ & Pris.\,(\%)$\downarrow$ \\
            \midrule
            FoundationPose$^\dagger$~\cite{wen2024foundationpose} + GT rest & --                        & --                              & --                      & \rthird{43.4}                   & \rfirst{0.8}            \\
            CAP-Net$^\dagger$~\cite{huang2025cap} + GT rest                 & --                        & --                              & --                      & \rsecond{32.9}                  & \rthird{14.6}           \\
            \midrule
            \textcolor{gray}{Inpainting base, before update}                & \textcolor{gray}{92.2}    & \textcolor{gray}{8.9}           & \textcolor{gray}{1.6}   & \textcolor{gray}{--}            & \textcolor{gray}{--}    \\
            RePaint$^\ddagger$~\cite{lugmayr2022repaint}                    & \rthird{95.7}             & \rthird{6.5}                    & \rfirst{2.2}            & --                              & --                      \\
            VoxHammer$^\ddagger$~\cite{li2026voxhammer}                     & \rsecond{96.3}            & \rfirst{5.3}                    & \rsecond{2.3}           & --                              & --                      \\
            VoxHammer+KV$^\ddagger$~\cite{li2026voxhammer}                  & 92.4                      & 10.7                            & \rfirst{2.2}            & --                              & --                      \\
            \midrule
            \textcolor{gray}{\methodname{} before update}                   & \textcolor{gray}{91.9}    & \textcolor{gray}{8.8}           & \textcolor{gray}{1.8}   & \textcolor{gray}{--}            & \textcolor{gray}{--}    \\
            \methodname{} (Ours)                                            & \rfirst{97.1}             & \rsecond{5.8}                   & \rfirst{2.2}            & \rfirst{10.6}                   & \rsecond{13.3}          \\
            \textcolor{gray}{\methodname{}, GT states as input} & \textcolor{gray}{97.1} & \textcolor{gray}{5.8} & \textcolor{gray}{2.1} & \textcolor{gray}{--} & \textcolor{gray}{--} \\
            \bottomrule
        \end{tabular}%
    }
\end{table}

\begin{table}[t]
    \centering
    \caption{Progressive updates on all regions that the full sequence of close-ups reveals, scored with all parts closed, 10 RevealBench-Astra families. One-step updates $\mathcal{A}$ with only the $k$-th close-up. Columns average 10, 10, 10, 6, 3 families.}
    \vspace{-2mm}
    \label{tab:progressive}
    \footnotesize
    \setlength{\tabcolsep}{4pt}
    \begin{tabular*}{\columnwidth}{@{\extracolsep{\fill}}llccccc@{}}
            \toprule
            &            & $\mathcal{A}$ & $\mathcal{B}_1$ & $\mathcal{B}_2$ & $\mathcal{B}_3$ & $\mathcal{B}_4$ \\
            \midrule
                    F@2\%$\uparrow$ & One-step & -- & 93.2 & 90.8 & 92.4 & 89.3 \\
            & Progressive & 89.0 & 93.2 & \textbf{94.2} & \textbf{94.4} & \textbf{95.5} \\
            \midrule
                    CD$\downarrow$ & One-step & -- & 0.0088 & 0.0095 & 0.0081 & 0.0093 \\
            & Progressive & 0.0106 & 0.0088 & \textbf{0.0083} & \textbf{0.0076} & \textbf{0.0070} \\
            \bottomrule
    \end{tabular*}
    \vspace{-2mm}
\end{table}

\subsection{Ablation Studies}
\label{sec:ablation}
The ablation rows of \cref{tab:main_rb} vary one component each, with every variant trained like \methodname{}. The anchor alone keeps the exterior and reaches a revealed F@2\% of 82.6, while the pixel features alone lose the exterior (88.9), since nothing anchors the generation to the observed surface of the existing asset. Only both together reach 86.7 with the exterior intact.
Training on only the first variant of each RevealTrain family lowers the revealed F@2\% from 86.7 to 83.6, indicating that the interior variants help the model generate the layout seen in the close-up.
Running the update with the ground-truth states as input raises the revealed F@2\% only from 86.7 to 87.3, so the state errors presented in \cref{tab:main_rb_state} barely affect the generated geometry.

\section{Conclusion}
\label{sec:conclusion}
We presented \methodname{}, a feed-forward model that generates an articulated asset from a single rest-state image and part masks and updates it when a later image shows parts opened, preserving the observed exterior and grounding the revealed interior in the new image. 
To train and evaluate updates, we built RevealTrain, a dataset of object families that look the same outside but differ inside, and RevealBench, a benchmark in which a close-up reveals what the rest image hides.
\methodname{} outperforms eight baselines in single-image generation, recovers revealed interiors better than reconstruction, generation, and inpainting baselines, corrects earlier joints, and works on real phone captures.
We see this as a step toward closed-loop asset generation, in which an asset is updated as interaction reveals more.

\noindent\textbf{Acknowledgement.} 
This work was supported through the Bavarian AI Foundation Model Initiative, funded by the Bavarian State Ministry of Science and the Arts (StMWK).
We acknowledge EuroHPC Joint Undertaking for awarding us access to Vega at IZUM, Slovenia; 
Karolina at IT4Innovations, Czech Republic; 
MeluXina at LuxProvide, Luxembourg; 
Leonardo at CINECA, Italy; 
Deucalion GPU at MACC, Portugal; 
and Discoverer GPU at SofiaTech, Bulgaria. 
These resources were provided under projects EHPC-BEN-2026B03-046, EHPC-DEV-2026D07-196, EHPC-BEN-2026B07-106, and EHPC-BEN-2026B07-147. 
We also thank Oliver Harrison for setting up the robot.

{
    \small
    \bibliographystyle{ieeenat_fullname}
    \bibliography{main}
}

\clearpage
\setcounter{section}{0}\setcounter{figure}{0}\setcounter{table}{0}\setcounter{equation}{0}
\renewcommand{\thesection}{S\arabic{section}}
\renewcommand{\thefigure}{S\arabic{figure}}
\renewcommand{\thetable}{S\arabic{table}}
\renewcommand{\theequation}{S\arabic{equation}}
\renewcommand{\theHsection}{S\arabic{section}}
\renewcommand{\theHfigure}{S\arabic{figure}}
\renewcommand{\theHtable}{S\arabic{table}}
\renewcommand{\theHequation}{S\arabic{equation}}
\clearpage
\newpage
\twocolumn[{%
            \centering
            {\Large\textbf{\thetitle}\\
                \vspace{0.5em}Supplementary Material\\}
            \vspace{1.0em}
            \includegraphics[width=0.97\textwidth]{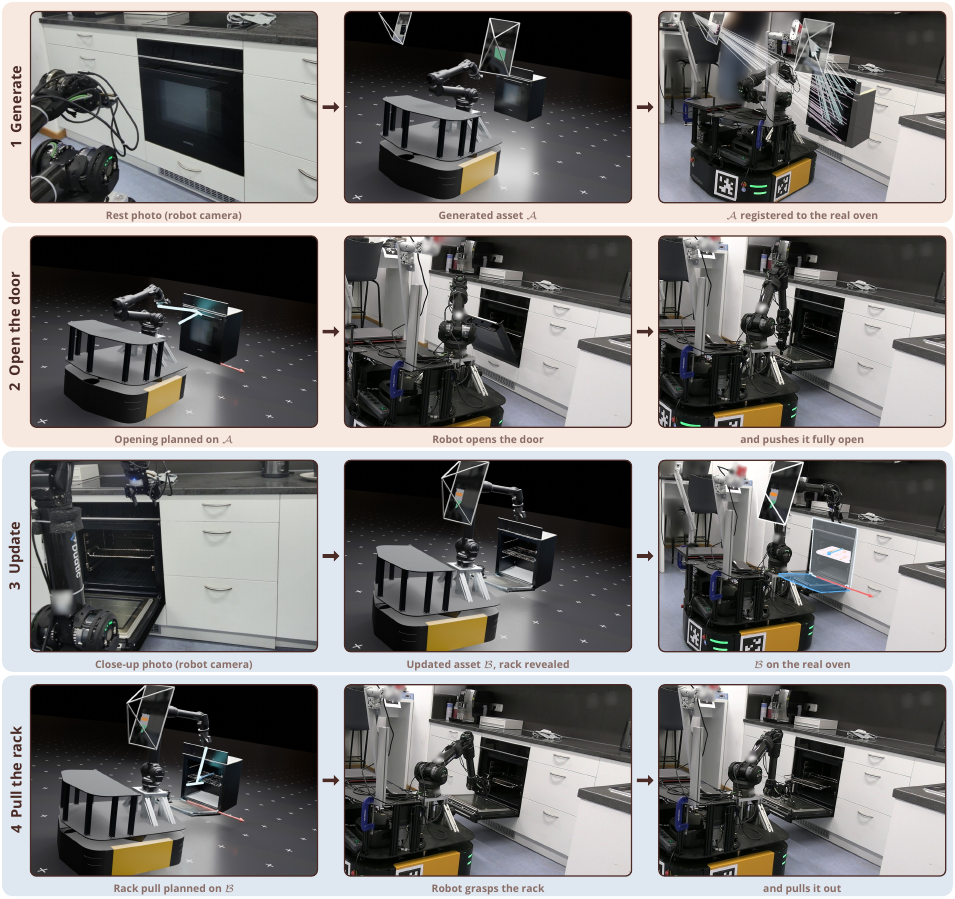}
            \vspace{-2mm}
            \captionof{figure}{\textbf{\methodname{} in the loop of a mobile manipulator.} The robot opens a real oven with the asset generated by \methodname{} and then pulls out the rack that only the update reveals. (1) \methodname{} generates $\mathcal{A}$ from a rest photo of the robot's camera, and $\mathcal{A}$ is registered to the oven with the robot's camera poses. (2) The door opening is planned on $\mathcal{A}$ and executed. (3) A close-up of the opened oven updates $\mathcal{A}$ to $\mathcal{B}$, which reveals the rack. (4) The rack pull is planned on $\mathcal{B}$ and executed. The full run is shown in the robot video on the project page.}
            \label{fig:supp_robot}
            \vspace{4mm}
        }]

This supplementary material provides additional implementation details (\cref{sec:supp_impl}), the construction and metrics of RevealBench (\cref{sec:supp_revealbench}), the full update results with the inpainting region used by the baselines, the opening states, the joint correction, and the progressive updates (\cref{sec:supp_update}), the real-world examples with the automatic part masks and a robot experiment (\cref{sec:supp_real}, \cref{fig:supp_robot}), the construction of RevealTrain (\cref{sec:supp_revealtrain}), the rest-state metrics, baselines, and full results with a study of how parts bind to their masks (\cref{sec:supp_rest}), ablations on the runtime and the input of the articulation head (\cref{sec:supp_ablations}), and additional results and failure cases (\cref{sec:supp_results}). Code and videos of the results and the robot experiment are available on the project page.

\section{Implementation Details}
\label{sec:supp_impl}

\subsection{Training}
\label{sec:supp_training}

\paragraph{Training on rest-state generation.}
We follow TRELLIS~\cite{xiang2025structured} to obtain the target latents and encode every part separately with the frozen VAE encoders of TRELLIS.
The SS latent $z_k$ is encoded from the occupancy of the part voxelized at $64^3$.
For the structured latent $\mathcal{F}_k$, the part alone is rendered from 150 views distributed over a sphere, the DINOv2 patch features of every view are projected onto the occupied voxels of the part and averaged over the views, and the SLat VAE encoder maps these aggregated features to $\mathcal{F}_k$.
The whole-object latent $\mathcal{F}_0$ is not encoded separately. It is formed on the union of the part voxels, and a voxel shared by several parts takes the mean of their latents.
To enable diverse single-view articulated object generation, we render each asset in its rest state with Blender from 20 views in front of it, with elevation from $0^\circ$ to $30^\circ$ and azimuth within $\pm60^\circ$ of the front, together with its part masks.
Each training sample randomly draws one of the 20 views and its part masks as the condition, and we train with $\mathcal{L}_{\mathrm{CFM}}$ (\cref{eq:cfm}) on the per-part SS and SLat targets. We train the articulation head jointly with the SLat flow with $\mathcal{L}_{\mathrm{art}}$ (\cref{eq:artloss}). The head reads the part tokens of the flow at the sampled time step $t$, so it is trained on all noise levels, while at inference it reads those of the last sampling step. We study the effect of averaging the part tokens over several noise levels at inference in \cref{sec:supp_agg}.

\paragraph{Training on update.}
The update is learned in a second training phase that adds RevealTrain to PM and mixes rest-state renders with close-ups.
A sample is a rest-state render with probability 0.5 and a close-up otherwise. 
We define two kinds of close-ups, namely opened close-ups and reveal close-ups.
Opened close-ups show either each movable part fully opened, each movable part opened to a fraction of its range drawn between 60 and 100\%, or two adjacent doors opened one after the other, with 10 views per state taken from where a person would stand to open the part. 
A view of a door or flap is kept only if at least 70\% of the opening it uncovers is visible, and is otherwise drawn again up to 12 times.
Reveal close-ups open the doors in front of a hidden part to 55--100\% of their range, leave the hidden part at rest, and have 16 views aimed at it.
The target latents are always those of the whole object at rest, encoded as for the rest-state generation, including the hidden parts that the close-up shows.
Since $\mathcal{A}$ does not exist during training, the inputs that an update takes from $\mathcal{A}$, namely the anchor, the part tokens of the open branch, and the poses for the pixel-aligned features, are built from the ground truth.
The anchor of a close-up sample is the exterior of the ground-truth object, computed as for $\mathcal{A}$ at inference.
The SS anchor re-encodes the exterior voxels with the SS VAE encoder, and the SLat anchor takes the structured latent of the target at these exterior voxels. 
For a RevealTrain object, the anchor comes from the first variant of its family.
A rest-state sample provides the part tokens needed to train the open branch (purple in \cref{fig:arthead}). 
The open branch reads the part tokens after the last transformer block of the SLat flow at the sampled time step $t$, so it is trained across all noise levels.
Its input is a randomly drawn close-up of the same object in an opened or reveal state, with the joints and states of that close-up as the targets of $\mathcal{L}_{\mathrm{open}}$.
On RevealTrain, this close-up may show any variant of the family, whose joints can differ from those of the rest-state render, so the open branch learns to correct a joint from the close-up.
For the pixel-aligned features, each moved part is posed with its ground-truth transform, perturbed along its own joint by up to $\pm5^\circ$ (revolute) or $\pm5\%$ of its range (prismatic), so that the network tolerates errors of the estimated state.
A close-up sample is trained without pixel-aligned features with probability 0.15, and a sample whose image condition is dropped for classifier-free guidance has none either.
The SS flow is trained with $\mathcal{L}_{\mathrm{CFM}}$ alone, and $\mathcal{L}_{\mathrm{art}}$ and $\mathcal{L}_{\mathrm{open}}$ are added to the loss of the SLat flow.

\paragraph{Additional training settings.}
Beyond the schedule of \cref{sec:protocol}, the SS flow is fully finetuned in both phases, and the LoRA of the SLat flow has rank 16 and $\alpha{=}32$ on all attention layers. The first phase uses the 3063 objects of the SINGAPO training split, and the second adds RevealTrain (600 objects of 100 families).
Both phases use 4 NVIDIA A100 GPUs, an effective batch size of 32, AdamW with a learning rate of $10^{-4}$ and no weight decay, and an image-condition dropout of 0.1 for classifier-free guidance.
We keep an exponential moving average of the weights with a rate of 0.9999 in the first phase and 0.9995 in the second and use it at inference.

\subsection{Model}
\label{sec:supp_model}

\paragraph{Articulation head.}
The head and its open branch each have 2 transformer blocks with 8 heads and 256 channels, and the [mean $\|$ max] pooled feature passes through three linear layers with SiLU and dropout 0.1.
The axis and origin live in the normalized frame of the voxel grid, and the range and the state $\hat{q}_k$ are in degrees divided by 180 for a rotational joint and in units of this normalized frame for a prismatic one.
At inference, the origin of a rotational joint is moved to the nearest point on the bounding box of its part, as in PAct~\cite{liu2026pact}.

\begin{figure*}[t]
    \centering
    \includegraphics[width=\textwidth]{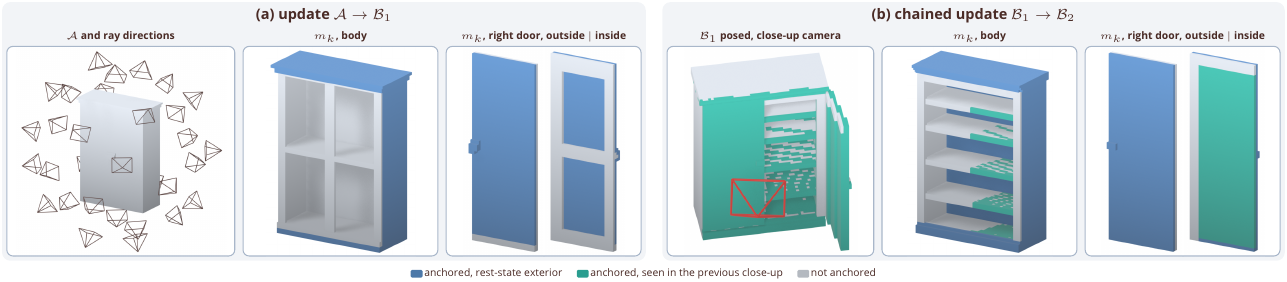}
    \caption{Exterior mask $m_k$ of the body and the right door of the wardrobe in \cref{fig:teaser}. (a) For $\mathcal{A}$, the shelves behind the closed doors and the inner face of the door are not anchored. (b) The voxels of the posed $\mathcal{B}_1$ that the previous close-up camera (red) sees (teal) are added to the rest-state exterior (blue), so the right compartment and the inner face of the opened door are anchored, while the unseen left compartment is generated again.}
    \label{fig:supp_exterior}
\end{figure*}

\paragraph{Exterior mask.}
Visibility is tested at $64^3$. A voxel of a part is exterior if it is the first occupied voxel of the whole asset along a ray from one of the 100 directions, which lie on a Fibonacci lattice (\cref{fig:supp_exterior}a).
For a chained update, a voxel counts as seen by an earlier close-up if it lies within 1.5 voxels of the depth buffer of the posed asset, computed at $128\times128$ over the conditioning crop (\cref{fig:supp_exterior}b). Each moved part is posed with its exported joint (\cref{sec:supp_inference}).

\paragraph{Pixel-aligned features.}
The close-up is cropped and resized to $518\times518$ as for the image condition, with the intrinsics adjusted to the crop, and cells that project outside it receive no feature. The SS and SLat flows have separate layers $W_b$ and tagging embeddings. In the SS stage, every cell of the dense $16^3$ grid of a part samples a feature, including empty ones, since the occupancy is only known after decoding, and in the SLat stage every token of the $32^3$ transformer grid does.
No occlusion test is applied, so a cell behind a visible surface receives the feature of that surface, and the network learns from the context which cells the observation constrains.

\subsection{Inference}
\label{sec:supp_inference}

\paragraph{Registration.}
The part mask fit of \cref{sec:align} maximizes the mean IoU between the part silhouettes of $\mathcal{A}$ and the part masks. 
$\mathcal{A}$ is first fitted to $I_r$ from a grid of azimuths ($-60^\circ$ to $60^\circ$) and elevations ($0^\circ$ to $40^\circ$) in steps of $10^\circ$, and the five best starts are refined with a simplex search~\cite{nelder1965simplex} in SciPy~\cite{virtanen2020scipy}. Since one photo cannot separate the size of $\mathcal{A}$ from its distance, the scale $s$ is then found by a one-dimensional search on the IoU of the unmoved parts in $I_u$.
Finally, the transform is refined over both photos jointly by minimizing
\begin{equation}
    \textstyle\sum_{v\in\{r,u\}}\sum_{j}\rho\big(\|\pi_v(s\mathbf{R}\mathbf{X}_j+\mathbf{t})-\mathbf{u}_j\|_2\big),
    \label{eq:reg}
\end{equation}
a perspective-n-point (PnP) problem in which each RoMa~\cite{edstedt2024roma} match pairs a pixel $\mathbf{u}_j$ of a photo with the 3D point $\mathbf{X}_j$ of $\mathcal{A}$ seen at the matched pixel of a render of $\mathcal{A}$, $\pi_v$ projects with the camera of view $v$, and $\rho$ is the Huber loss with a scale of 4 pixels.
Rendering, matching within the masks, and solving alternate four times, and in the close-up only the parts that have not moved are rendered and matched.
Registration only estimates the transform from the frame of $\mathcal{A}$ to that of the cameras, so any second photo with a known camera pose can take the place of $I_u$. 
In the robot experiment (\cref{sec:supp_robot}), it is a second rest-state photo from another camera of the robot.

\paragraph{Sampling.}
Both stages use 25 steps of the TRELLIS schedule (time rescaling 3) with classifier-free guidance only in the high-noise half ($t\in[0.5,1]$), with strength 7 for the SS stage and 3 for the SLat stage.

\paragraph{Moved parts and exported joints.}
A part is considered as moved when its joint type is not fixed and its estimated state $\hat{q}_k$ exceeds 0.1. The moved and the revealed parts of $\mathcal{B}$ take the joints that the articulation head predicts on $\mathcal{B}$, and the other parts keep those of $\mathcal{A}$. All movable parts are attached to the body. A newly revealed part has no anchor, so its slot starts from pure noise.

\section{RevealBench}
\label{sec:supp_revealbench}

\paragraph{RevealBench-ArtiVerse.}
We select objects from the test split of ArtiVerse~\cite{iliash2026artiverse} in two steps. We first collect the 389 test objects of storage furniture, tables, and kitchen appliances, and remove 14 objects whose only moving parts are lids and 3 objects without a door or drawer, which leaves 372 candidates. Visual inspection for the criteria of \cref{sec:protocol} then leaves 163 objects, each with at least one door or drawer.
The selection covers chests of drawers (25), nightstands (25), refrigerators (23), wall cabinets (18), ovens (15), sideboards (14), TV stands (14), desks (8), armoires (7), dishwashers (4), microwaves (2), storage benches (2), washing machines (2), and one each of coffee table, display cabinet, locker, and tool cabinet.

\paragraph{RevealBench-Astra.}
The 34 families of RevealBench-Astra are built with our agentic modeling pipeline (\cref{sec:supp_revealtrain}), 21 from ABO references~\cite{collins2022abo}, 12 from our own photographs of real furniture, and 1 from PM~\cite{xiang2020sapien}. Because the variants of a family share their exterior and their cases, this split additionally tests whether a method distinguishes siblings from their close-ups.

\paragraph{Cases.}
The first case of an object opens its main door, the one with the largest front area, or its middle drawer. The second case may also open a pair of adjacent doors, which 54 cases do (42 in RevealBench-ArtiVerse and 12 in RevealBench-Astra), and it reopens the same part only for objects with a single movable part. Doors are opened to 85--100\% of their range and drawers and trays to 70--100\%. The close-up camera follows a fixed rule per opening type. For a door, it looks into the cavity from the side of the free edge, and for a drawer, it is aimed at the drawer and frames it from edge to edge. Every placement is checked for the visibility of the opened part and of the revealed region.
Both views are rendered at $512\times512$ pixels with a field of view of $40^\circ$ in Blender Cycles under the lighting of TRELLIS, with part masks rasterized from the same meshes. 
The cameras are given in the normalized frame of the object, in which the rest-state bounding box is centered at the origin and has a largest extent of 1.

\begin{figure*}[t]
    \centering
    \includegraphics[width=\textwidth]{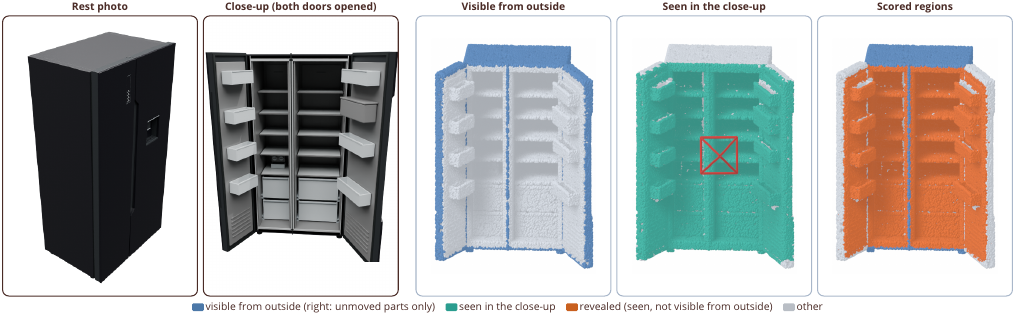}
    \caption{Regions of the RevealBench metrics on a RevealBench-Astra refrigerator with both doors opened. From left, the rest photo, the close-up, and the posed ground truth with the points visible from outside the closed object, the points seen by the close-up camera (red), and the resulting revealed region and unmoved exterior.}
    \label{fig:supp_reveal}
\end{figure*}

\paragraph{Metrics.}
For scoring, prediction and ground truth are centered on their rest-state bounding boxes, and the prediction is rescaled per axis to the ground-truth extents. The opened part is posed as in \cref{sec:protocol} based on ground-truth joints and states. For methods without our part masks, the opened part is the predicted movable part whose rest-state bounding box overlaps that of the ground-truth part most, and none is opened if the overlap is negligible.
Both surfaces are sampled with 200k points. A ground-truth point is visible from outside if it passes a depth test in one of 150 orthographic views at $768\times768$ pixels on a Fibonacci sphere around the closed object, and it is seen in the close-up if it lies in the frame and passes a depth test through the close-up camera with the object in the state of the case. These two tests give the revealed region and the unmoved exterior of \cref{sec:protocol}, shown in \cref{fig:supp_reveal}.
For a region and a tolerance $\tau$ of 1, 2, 5, or 10\% of the diagonal $D$ of the ground-truth rest-state bounding box, the recall $R$ is the fraction of ground-truth points of the region that have a predicted point within $\tau$, and the precision $P$ is the fraction of the predicted points whose nearest ground-truth point lies in the region that are within $\tau$ of it. The F-score is $2PR/(P+R)$, and the Chamfer distance is the bidirectional Chamfer distance between the ground-truth points of the region and these predicted points, averaged over both directions, in the normalized frame.
For appearance, the posed prediction and ground truth are rendered in the same way from two novel views. LPIPS and DINO-I, here with DINOv2 ViT-L/14, compare the two renders around the whole object and around the pixels onto which the revealed region projects.
Cases without output are left out. The metrics are averaged per object in RevealBench-ArtiVerse and per family in RevealBench-Astra, then over each split, and the overall value is the mean of the two splits.

\section{Full Update Results}
\label{sec:supp_update}
This section gives the full results and baseline details on RevealBench (\cref{sec:supp_fullrb}), the inpainting region of the training-free baselines (\cref{sec:supp_inpaint}), the opening-state estimation (\cref{sec:supp_state}), the joint correction (\cref{sec:supp_joint}), and the progressive updates (\cref{sec:supp_progressive}).

\subsection{Results and Baselines}
\label{sec:supp_fullrb}
\cref{tab:supp_rb} extends \cref{tab:main_rb} with the F-scores at 1, 5, and 10\% of the object diagonal and the appearance of the whole object, and \cref{tab:supp_rb_split} gives the geometry per split.
DIPO~\cite{wu2026dipo} expects a full view of the articulated object, so we also run it with an image of the opened object taken from the same viewpoint as the rest image, in place of the close-up, which changes its revealed F@2\% only from 53.2 to 54.3. Its ground-truth part graph also gives it the number of hidden parts, and its part meshes are retrieved from PM outside the test split, as in SINGAPO.
FreeArt3D runs once per opened part, with its multi-joint mode for door pairs, and its six renders move the opened part from closed to the opening of the case. Its failures (27.6\% of the cases) are runs that separate no moving part, and a run takes 30 minutes on an A100.
LARM and Artic-O receive three front views of the closed object and three with the part fully opened, and LARM also the camera poses. A door pair is run once per door and the results are merged. Artic-O outputs a point cloud, so it has no appearance scores.

\begin{table*}[t]
    \centering
    \caption{Update results on RevealBench (598 cases). $^\dagger$ privileged inputs (several articulation states), $^\ddagger$ opened part given. Means are over the cases a method produces, and FreeArt3D and LARM fail on 27.6\% and 2.3\% of the cases. The ablations and the gray rows are not ranked. The predicted-pose row is scored for the revealed F@2\%, the Chamfer distance, and the exterior only, and the auto-mask rows on the 542 cases where these masks succeed.}
    \label{tab:supp_rb}
    \setlength{\tabcolsep}{3.5pt}
    \scalebox{0.85}{%
        \begin{tabular}{lcccccccccc}
            \toprule
                                                           & \multicolumn{5}{c}{Revealed geometry} & \multicolumn{1}{c}{Unmoved ext.} & \multicolumn{2}{c}{Whole object} & \multicolumn{2}{c}{Revealed region}                                                                                                                                                             \\
            \cmidrule(lr){2-6} \cmidrule(lr){7-7} \cmidrule(lr){8-9} \cmidrule(lr){10-11}
            Method                                         & F@1\% $\uparrow$                      & F@2\% $\uparrow$                 & F@5\% $\uparrow$                 & F@10\% $\uparrow$                   & CD $\downarrow$          & F@2\% $\uparrow$       & LPIPS $\downarrow$      & DINO $\uparrow$         & LPIPS $\downarrow$      & DINO $\uparrow$         \\
            \midrule
            \multicolumn{11}{l}{\textit{Multi-state methods (several states given)}}                                                                                                                                                                                                                                                                                       \\
            FreeArt3D$^\dagger$~\cite{chen2025freeart3d}   & 37.2 & 56.8 & 82.7 & 94.9 & 0.0354 & 76.1 & 0.5104 & 0.7369 & 0.5623 & 0.6530  \\
            LARM$^\dagger$~\cite{yuan2025larm}             & 20.8 & 37.3 & 72.2 & 91.8 & 0.0509 & 30.6 & 0.6607 & 0.2816 & 0.7046 & 0.2228  \\
            Artic-O$^\dagger$~\cite{wang2026artic}         & 42.5 & 69.2 & 92.7 & 98.5 & 0.0246 & 78.0 & -- & -- & -- & --  \\
            \midrule
            \multicolumn{11}{l}{\textit{Two-image articulated generation}}                                                                                                                                                                                                                                                                                                 \\
            DIPO~\cite{wu2026dipo} (close-up)              & 35.6 & 53.2 & 83.6 & 96.9 & 0.0351 & 78.3 & 0.6121 & 0.5743 & 0.6510 & 0.5650  \\
            DIPO~\cite{wu2026dipo} (opened, rest view)     & 35.7 & 54.3 & 85.5 & 97.5 & 0.0329 & 78.6 & 0.6082 & 0.5847 & 0.6418 & 0.5890  \\
            \midrule
            \multicolumn{11}{l}{\textit{Training-free inpainting on our base model}}                                                                                                                                                                                                                                                                                       \\
            \textcolor{gray}{Inpainting base, before} & \textcolor{gray}{59.0} & \textcolor{gray}{76.7} & \textcolor{gray}{93.4} & \textcolor{gray}{98.6} & \textcolor{gray}{0.0191} & \textcolor{gray}{93.5} & \textcolor{gray}{0.3763} & \textcolor{gray}{0.8262} & \textcolor{gray}{0.4351} & \textcolor{gray}{0.7953} \\  %
            RePaint$^\ddagger$~\cite{lugmayr2022repaint}   & 57.7 & 76.9 & 95.8 & \rthird{99.5} & 0.0183 & \rthird{93.2} & \rthird{0.3548} & \rthird{0.8983} & \rthird{0.3976} & \rthird{0.8769}  \\
            VoxHammer$^\ddagger$~\cite{li2026voxhammer}    & \rsecond{60.1} & \rsecond{79.8} & \rsecond{96.7} & \rsecond{99.6} & \rsecond{0.0168} & \rsecond{93.5} & \rsecond{0.3486} & \rsecond{0.9046} & \rsecond{0.3878} & \rsecond{0.8854}  \\
            VoxHammer+KV$^\ddagger$~\cite{li2026voxhammer} & \rthird{58.3} & \rthird{77.3} & \rthird{95.9} & \rsecond{99.6} & \rthird{0.0175} & \rsecond{93.5} & 0.3566 & 0.8885 & 0.4042 & 0.8624  \\
            \midrule
            \multicolumn{11}{l}{\textit{Update (rest image + close-up)}}                                                                                                                                                                                                                                                                                                   \\
            \textcolor{gray}{\methodname{} before update} & \textcolor{gray}{61.2} & \textcolor{gray}{79.0} & \textcolor{gray}{93.6} & \textcolor{gray}{98.7} & \textcolor{gray}{0.0184} & \textcolor{gray}{94.1} & \textcolor{gray}{0.3690} & \textcolor{gray}{0.8236} & \textcolor{gray}{0.4300} & \textcolor{gray}{0.7896} \\
            \methodname{} (Ours)                           & \rfirst{68.6} & \rfirst{86.7} & \rfirst{98.7} & \rfirst{99.8} & \rfirst{0.0130} & \rfirst{94.0} & \rfirst{0.3250} & \rfirst{0.9156} & \rfirst{0.3496} & \rfirst{0.9028}  \\
            \midrule
            \multicolumn{11}{l}{\textit{Ablations of \methodname{}}} \\
            Anchor only & 63.6 & 82.6 & 97.7 & 99.8 & 0.0154 & 94.2 & 0.3365 & 0.9163 & 0.3672 & 0.9022 \\
            Pixel-aligned only & 60.1 & 79.3 & 96.5 & 99.5 & 0.0172 & 88.9 & 0.3461 & 0.9175 & 0.3670 & 0.9015 \\
            First variant only & 63.8 & 83.6 & 97.2 & 99.6 & 0.0152 & 93.6 & 0.3442 & 0.8838 & 0.3822 & 0.8601 \\
            \midrule
            \textcolor{gray}{Ours, GT states as input}     & \textcolor{gray}{69.0}                & \textcolor{gray}{87.3}           & \textcolor{gray}{98.8}           & \textcolor{gray}{99.8}              & \textcolor{gray}{0.0128} & \textcolor{gray}{94.1} & \textcolor{gray}{0.3216} & \textcolor{gray}{0.9183} & \textcolor{gray}{0.3434} & \textcolor{gray}{0.9063} \\
            \textcolor{gray}{Ours, predicted pose for scoring}         & \textcolor{gray}{--}                  & \textcolor{gray}{81.2}           & \textcolor{gray}{--}             & \textcolor{gray}{--}                & \textcolor{gray}{0.0179} & \textcolor{gray}{93.6} & \textcolor{gray}{--}    & \textcolor{gray}{--}    & \textcolor{gray}{--}    & \textcolor{gray}{--}    \\
            \textcolor{gray}{Auto masks as input, before} & \textcolor{gray}{59.3} & \textcolor{gray}{77.1} & \textcolor{gray}{93.6} & \textcolor{gray}{98.7} & \textcolor{gray}{0.0193} & \textcolor{gray}{93.7} & \textcolor{gray}{0.3801} & \textcolor{gray}{0.8195} & \textcolor{gray}{0.4398} & \textcolor{gray}{0.7868} \\
            \textcolor{gray}{Auto masks as input} & \textcolor{gray}{65.0} & \textcolor{gray}{83.6} & \textcolor{gray}{97.7} & \textcolor{gray}{99.6} & \textcolor{gray}{0.0147} & \textcolor{gray}{93.1} & \textcolor{gray}{0.3419} & \textcolor{gray}{0.8949} & \textcolor{gray}{0.3682} & \textcolor{gray}{0.8793} \\
            \bottomrule
        \end{tabular}
    }
\end{table*}

\begin{table}[t]
    \centering
    \caption{Geometry of \cref{tab:supp_rb} per split of RevealBench, the F@2\% (Rev.) and Chamfer distance of the revealed region and the F@2\% of the unmoved exterior (Ext.).}
    \label{tab:supp_rb_split}
    \footnotesize
    \setlength{\tabcolsep}{1.6pt}
    \begin{tabular}{lcccccc}
        \toprule
        & \multicolumn{3}{c}{ArtiVerse} & \multicolumn{3}{c}{Astra} \\
        \cmidrule(lr){2-4} \cmidrule(lr){5-7}
        Method & Rev.$\uparrow$ & CD$\downarrow$ & Ext.$\uparrow$ & Rev.$\uparrow$ & CD$\downarrow$ & Ext.$\uparrow$ \\
        \midrule
        \multicolumn{7}{l}{\textit{Multi-state methods (several states given)}} \\
        FreeArt3D$^\dagger$~\cite{chen2025freeart3d} & 57.2 & 0.0356 & 76.3 & 56.5 & 0.0353 & 75.9  \\
        LARM$^\dagger$~\cite{yuan2025larm} & 36.0 & 0.0529 & 33.3 & 38.6 & 0.0489 & 28.0  \\
        Artic-O$^\dagger$~\cite{wang2026artic} & 68.3 & 0.0259 & 76.9 & 70.2 & 0.0233 & 79.2  \\
        \midrule
        \multicolumn{7}{l}{\textit{Two-image articulated generation}} \\
        DIPO (close-up) & 54.6 & 0.0351 & 77.9 & 51.7 & 0.0351 & 78.6  \\
        DIPO (opened, rest view) & 54.9 & 0.0340 & 77.9 & 53.8 & 0.0319 & 79.3  \\
        \midrule
        \multicolumn{7}{l}{\textit{Training-free inpainting on our base model}} \\
        RePaint$^\ddagger$ & \rthird{73.0} & 0.0212 & 91.0 & 80.9 & 0.0155 & \rthird{95.5}  \\
        VoxHammer$^\ddagger$ & \rsecond{76.3} & \rsecond{0.0194} & \rthird{91.3} & \rsecond{83.3} & \rsecond{0.0143} & \rsecond{95.6}  \\
        VoxHammer+KV$^\ddagger$ & 72.7 & \rthird{0.0200} & \rsecond{91.5} & \rthird{81.8} & \rthird{0.0151} & \rthird{95.5}  \\
        \midrule
        \multicolumn{7}{l}{\textit{Update (rest image + close-up)}} \\
        \textcolor{gray}{\methodname{} before update} & \textcolor{gray}{73.3} & \textcolor{gray}{0.0217} & \textcolor{gray}{92.2} & \textcolor{gray}{84.6} & \textcolor{gray}{0.0150} & \textcolor{gray}{96.1} \\
        \methodname{} (Ours) & \rfirst{84.4} & \rfirst{0.0144} & \rfirst{92.0} & \rfirst{88.9} & \rfirst{0.0116} & \rfirst{96.0}  \\
        \bottomrule
    \end{tabular}
\end{table}

\begin{figure}[!htb]
    \centering
    \includegraphics[width=0.9\columnwidth]{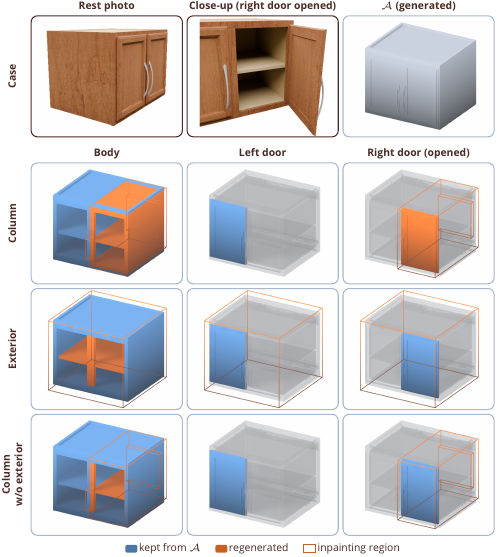}
    \caption{Regenerated region of the inpainting baselines for each part of $\mathcal{A}$, on a RevealBench cabinet whose right door is opened. Top, the rest photo, the close-up, and $\mathcal{A}$, generated from the rest photo alone. Below, the inpainting region of each part (orange outline) for the three regions of \cref{sec:supp_inpaint}, in which everything except the blue voxels is regenerated.}
    \label{fig:supp_inpaint}
\end{figure}

\subsection{Inpainting Region}
\label{sec:supp_inpaint}
Only the inpainting of these baselines is training-free, not the model it runs on. TRELLIS never sees a close-up of an opened object in training, so inpainting with it could not show how the close-up helps. We therefore run all inpainting baselines on a base model trained with the same data (PM and RevealTrain, close-ups included), initialization, and number of steps as \methodname{}, without the anchor, the pixel-aligned features, and the state estimation. The baselines and \methodname{} thus see the same training data and differ only in how the update uses $\mathcal{A}$ and the close-up, so the comparison is fair.
The inpainting baselines regenerate a region of $\mathcal{A}$ and keep the rest, and they differ only in how the kept voxels are held during sampling. RePaint~\cite{lugmayr2022repaint} replaces them at every step with the final latents of $\mathcal{A}$, noised to that step. VoxHammer~\cite{li2026voxhammer} replaces them with the latents that $\mathcal{A}$ had at the same step of its own sampling, which need no inversion here since the base model generated $\mathcal{A}$, and VoxHammer+KV also copies their self-attention keys and values.
Which region is regenerated matters more than this choice (\cref{fig:supp_inpaint} and \cref{tab:supp_inpaint}). We compare the column, the space behind the opened part along the direction in which it opens, the whole asset except the voxels visible from outside, and the column without these visible voxels. Regenerating everything except the visible exterior is the worst region for every method, since it also regenerates the closed parts and damages the unmoved exterior (85.3--88.7 against 90.9--93.5 for the other regions). The column without the visible exterior is the best region for RePaint and VoxHammer and is close to the full column for VoxHammer+KV, so \cref{tab:main_rb} uses it for all three.

\begin{table}[t]
    \centering
    \caption{Regenerated region of the training-free inpainting baselines on RevealBench (598 cases), with the metrics of \cref{tab:main_rb} and the revealed F@1\%. All rows are given the opened part. The base model row is the initial asset without update.}
    \label{tab:supp_inpaint}
    \footnotesize
    \setlength{\tabcolsep}{2pt}
    \resizebox{\columnwidth}{!}{%
        \begin{tabular}{llcccccc}
            \toprule
                         &          & \multicolumn{3}{c}{Revealed geo.} & Unmoved         & \multicolumn{2}{c}{Revealed app.}                                                        \\
            \cmidrule(lr){3-5} \cmidrule(lr){6-6} \cmidrule(lr){7-8}
            Method       & Region   & F@1\%$\uparrow$                   & F@2\%$\uparrow$ & CD$\downarrow$                    & F@2\%$\uparrow$ & LPIPS$\downarrow$ & DINO$\uparrow$ \\
            \midrule
            \textcolor{gray}{Base model} & \textcolor{gray}{--} & \textcolor{gray}{59.0} & \textcolor{gray}{76.7} & \textcolor{gray}{0.0191} & \textcolor{gray}{93.5} & \textcolor{gray}{0.4351} & \textcolor{gray}{0.7953} \\
            \midrule
            RePaint      & column   & 54.9 & 73.5 & 0.0206 & 90.9 & 0.4136 & 0.8654  \\
                         & exterior & 50.5 & 70.6 & 0.0218 & 85.3 & 0.4572 & 0.8431  \\
                         & col.\ w/o ext. & 57.7 & 76.9 & 0.0183 & \rsecond{93.2} & 0.3976 & \rthird{0.8769}  \\
            \midrule
            VoxHammer    & column   & \rsecond{59.4} & \rsecond{78.2} & 0.0180 & 92.2 & \rsecond{0.3932} & \rsecond{0.8825}  \\
                         & exterior & 54.3 & 73.5 & 0.0204 & 86.7 & 0.4358 & 0.8565  \\
                         & col.\ w/o ext. & \rfirst{60.1} & \rfirst{79.8} & \rfirst{0.0168} & \rfirst{93.5} & \rfirst{0.3878} & \rfirst{0.8854}  \\
            \midrule
            VoxHammer+KV & column   & \rthird{59.3} & \rthird{78.0} & \rthird{0.0176} & \rthird{92.9} & \rthird{0.3970} & 0.8757  \\
                         & exterior & 55.5 & 74.5 & 0.0203 & 88.7 & 0.4221 & 0.8558  \\
                         & col.\ w/o ext. & 58.3 & 77.3 & \rsecond{0.0175} & \rfirst{93.5} & 0.4042 & 0.8624  \\
            \bottomrule
        \end{tabular}%
    }
\end{table}

\subsection{Opening-State Estimation}
\label{sec:supp_state}
The pose estimators of \cref{tab:main_rb_state} predict a 6D pose $(\hat{R}_r,\hat{t}_r)$ of the opened part in the rest image and $(\hat{R}_c,\hat{t}_c)$ in the close-up, and with the ground-truth rest pose given, only the close-up pose is estimated and the rest pose is the identity. 
The opening state is the size of the motion between the two poses,
\begin{gather*}
    \hat{\theta} = \arccos\frac{\operatorname{tr}(\hat{R}_c\hat{R}_r^\top)-1}{2}, \\
    \hat{d} = \bigl\|(\hat{R}_c c_0+\hat{t}_c)-(\hat{R}_r c_0+\hat{t}_r)\bigr\|,
\end{gather*}
the rotation angle for a revolute part and the displacement of the rest part center $c_0$ for a prismatic part. 
CAP-Net predicts oriented part boxes, and a box looks the same after a 180\textdegree{} rotation about any of its axes, so an opening of $\theta$ cannot be told from one of $180^\circ-\theta$.
Among these candidate rotations, we keep the one whose motion leaves an edge of the rest box most nearly fixed, as a hinge does. 
We match CAP-Net's detections to the opened part by their overlap with its ground-truth mask. 
\methodname{} predicts the joint and its state directly, and a part counts as estimated only when its predicted joint type is correct. 
With the ground-truth rest pose, over the doors whose error stays below 30\textdegree{}, the mean errors are 2.7\textdegree{} for FoundationPose and 5.8\textdegree{} for CAP-Net, against 8.9\textdegree{} for \methodname{}. 
The pose estimators are thus more precise when they succeed, and their means in \cref{tab:main_rb_state} come from a few large errors, such as flat doors that FoundationPose estimates about 180\textdegree{} from their true pose (68 doors).

\subsection{Joint Correction}
\label{sec:supp_joint}
\cref{tab:supp_rb_joint} gives the joint accuracy of all methods on the opened parts. 
The pivot error is the distance between the predicted and the true axis line, computed for revolute joints with a correctly predicted type only, since a translation has no pivot.
FreeArt3D, LARM, and Artic-O are given the joint type, so their type accuracy is 100\% by construction. FreeArt3D also selects its axis among the coordinate axes, which explains its low axis error.
The inpainting baselines start from the initial asset of the base model and \methodname{} from our final model trained with anchors, state estimation, and pixel-aligned features, so the two rows before the update differ slightly.
Of the 652 opened parts, the update of \methodname{} fixes 36 of the 48 wrong joint types of $\mathcal{A}$ and breaks 6, while the inpainting variants on the same articulation head fix 23 to 25. 
Counted over all parts, the type accuracy rises from 92.6 to 97.2\%. 
\cref{tab:supp_rb_joint} averages per object or family instead, which gives 91.9 and 97.1\%.
The number of axes off by more than 45\textdegree{} drops from 55 to 36, and the median axis error stays at about 0.5\textdegree{}. 
On the parts whose type is correct both before and after the update, the mean axis error rises from 2.6 to 4.4\textdegree{}, so the update fixes joints with large errors at a small cost in precision on joints that were already right.
The pivot error rises for two reasons. On the 328 hinges typed correctly before and after the update, the median pivot error grows only from 0.8 to 1.0\% of the diagonal. The 15 hinges whose type the update corrects enter the pivot error only after the update, with a mean of 4.4\%.
Keeping the joint of the open branch instead of predicting it again on $\mathcal{B}$ lowers the pivot error to 2.0\% but drops the type accuracy to 94.4\% and raises the axis error to 7.6\textdegree{}.

\begin{table}[t]
    \centering
    \caption{Joint accuracy on the opened parts of RevealBench (652 parts). Matched is the fraction of opened parts a method recovers. Type, axis, and pivot are computed over the recovered parts. $^\dagger$ marks privileged inputs (several articulation states) and $^\ddagger$ a given opened part. Only the inpainting baselines and \methodname{} are ranked.}
    \label{tab:supp_rb_joint}
    \footnotesize
    \setlength{\tabcolsep}{2pt}
    \resizebox{\columnwidth}{!}{%
        \begin{tabular}{lcccc}
            \toprule
            Method                                         & Matched\,(\%)$\uparrow$ & Type\,(\%)$\uparrow$ & Axis\,(\textdegree)$\downarrow$ & Pivot\,(\%)$\downarrow$ \\
            \midrule
            \multicolumn{5}{l}{\textit{Multi-state methods (several states given)}}                                                                                     \\
            FreeArt3D$^\dagger$~\cite{chen2025freeart3d}   & 70.6                    & 100.0                & 1.3                             & 9.3                     \\
            LARM$^\dagger$~\cite{yuan2025larm}             & 65.6                    & 100.0                & 20.5                            & 16.2                    \\
            Artic-O$^\dagger$~\cite{wang2026artic}         & 61.9                    & 100.0                & 3.6                             & 3.3                     \\
            \midrule
            \multicolumn{5}{l}{\textit{Two-image articulated generation}}                                                                                               \\
            DIPO~\cite{wu2026dipo} (close-up)              & 83.9                    & 92.9                 & 4.8                             & 3.7                     \\
            DIPO~\cite{wu2026dipo} (opened, rest view)     & 86.2                    & 94.5                 & 4.5                             & 3.4                     \\
            \midrule
            \multicolumn{5}{l}{\textit{Training-free inpainting on our base model}}                                                                                     \\
            \textcolor{gray}{Base model before update} & \textcolor{gray}{99.8} & \textcolor{gray}{92.2} & \textcolor{gray}{8.9} & \textcolor{gray}{1.6} \\
            RePaint$^\ddagger$~\cite{lugmayr2022repaint}   & \rsecond{99.3} & \rthird{95.7} & \rthird{6.5} & \rfirst{2.2}  \\
            VoxHammer$^\ddagger$~\cite{li2026voxhammer}    & 99.0 & \rsecond{96.3} & \rfirst{5.3} & \rsecond{2.3}  \\
            VoxHammer+KV$^\ddagger$~\cite{li2026voxhammer} & \rthird{99.2} & 92.4 & 10.7 & \rfirst{2.2}  \\
            \midrule
            \multicolumn{5}{l}{\textit{Update (rest image + close-up)}}                                                                                                 \\
            \textcolor{gray}{\methodname{} before update} & \textcolor{gray}{100.0} & \textcolor{gray}{91.9} & \textcolor{gray}{8.8} & \textcolor{gray}{1.8} \\
            \methodname{} (Ours)                           & \rfirst{100.0} & \rfirst{97.1} & \rsecond{5.8} & \rfirst{2.2}  \\
            \bottomrule
        \end{tabular}%
    }
\end{table}

\subsection{Progressive Updates}
\label{sec:supp_progressive}
\begin{figure*}[t]
    \centering
    \includegraphics[width=\textwidth]{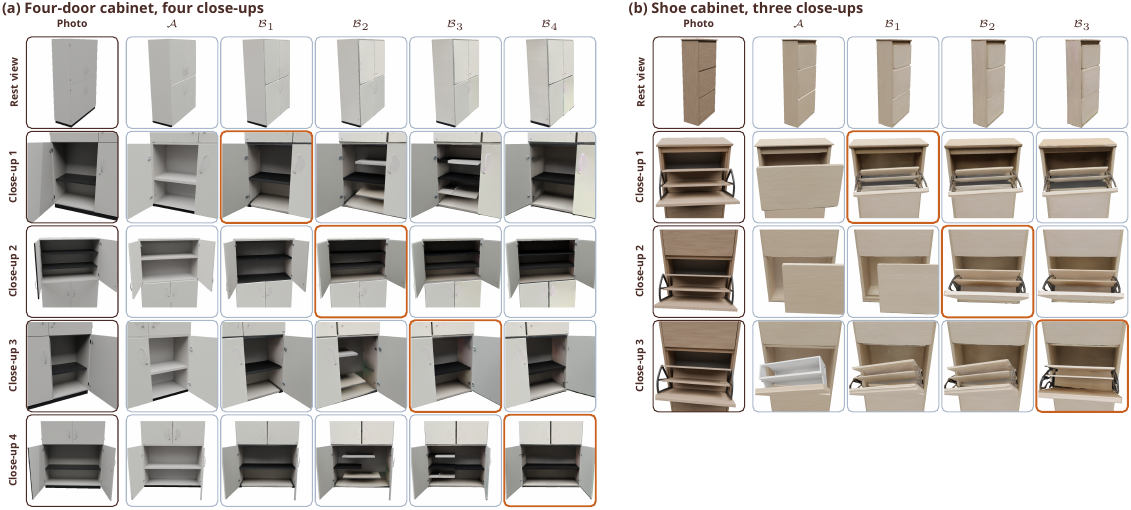}
    \caption{Progressive updates of two RevealBench-Astra objects. Each row shows every stage from the view of one photo, and the stage that receives the close-up of the row is framed in orange. In (a), the second close-up shows the upper compartment, yet its update adds loose shelf pieces in the lower one, since the anchor keeps the surfaces that earlier close-ups saw but not the empty space between them. Later close-ups of the lower compartment remove them. In (b), $\mathcal{A}$ predicts two of the three flaps as drawers and tilts the hinge of the third by $19^\circ$, and each joint is corrected by the update that sees its flap and kept afterwards.}
    \label{fig:supp_chain}
\end{figure*}

Each of the 40 objects of \cref{tab:progressive} receives two to four close-ups, one per visible movable part. 
The first two are its two RevealBench cases, and the later ones open other visible parts or pairs of doors. 
From the second close-up on, the movable parts that a close-up does not move are kept from the current asset instead of being regenerated.
Of the 34 RevealBench-Astra families, the 7 with a single visible movable part cannot be chained, and the 17 whose movable parts are all drawers are left out, since their identical drawers are largely predictable from the exterior. 
All stages of a run are scored with one alignment computed from $\mathcal{A}$. At $\mathcal{B}_1$, both rows of \cref{tab:progressive} are the same update, since the first progressive step is a one-step update of $\mathcal{A}$.
Since the columns of \cref{tab:progressive} average different families, \cref{tab:supp_progressive} follows fixed sets of families. 
In each set, the revealed F@2\% never drops over the close-ups and the Chamfer distance never grows. 
Compared object by object (third block of \cref{tab:supp_progressive}), each of the first three close-ups improves the revealed F@2\% of more objects than it worsens, and the fourth adds nothing on average.
Regenerating all parts at every close-up, instead of keeping those it does not move, gives about the same geometry, but their appearance drifts over the sequence, with doors turning pink or green. A likely reason is that the anchor only sets the start of the flow, so the anchored voxels are denoised again at every update and can change slightly each time, and since \methodname{} is trained on single updates only, these changes add up. Keeping the unmoved parts removes this drift for them, while the body and the moved parts are still regenerated and can change in appearance, as in \cref{fig:supp_chain}(a). 
The regions revealed before the last close-up keep almost all of their gain (bottom of \cref{tab:supp_progressive}), and \cref{fig:supp_chain} shows two sequences from the view of every close-up.

\begin{table}[t]
    \centering
    \caption{Progressive updates on the 10 families of \cref{tab:progressive}. The first block follows the same families through every stage, with the Chamfer distance $\times100$. The second regenerates every part at each close-up instead of keeping the parts it does not move, on the 3 families of four close-ups. The third gives the mean per-object change of the revealed F@2\% from the previous stage and the number of objects it improves. The last shows whether later updates keep what an earlier close-up revealed, scored on the 76 regions revealed before the last close-up in $\mathcal{A}$, right after the close-up that revealed them, and at the end, and the same for the unmoved exterior.}
    \label{tab:supp_progressive}
    \footnotesize
    \setlength{\tabcolsep}{2pt}
    \begin{tabular*}{\columnwidth}{@{\extracolsep{\fill}}llccccc@{}}
        \toprule
        Families & & $\mathcal{A}$ & $\mathcal{B}_1$ & $\mathcal{B}_2$ & $\mathcal{B}_3$ & $\mathcal{B}_4$ \\
        \midrule
        \multicolumn{7}{l}{\textit{Same families at every stage}} \\
        10 ($\geq$2 close-ups) & F@1\% & 74.4 & 79.0 & 81.5 & -- & -- \\
                               & F@2\% & 89.0 & 93.2 & 94.2 & -- & -- \\
                               & CD    & 1.06 & 0.88 & 0.83 & -- & -- \\
        6 ($\geq$3 close-ups)  & F@1\% & 77.3 & 80.1 & 81.9 & 83.2 & -- \\
                               & F@2\% & 91.4 & 92.6 & 92.9 & 94.4 & -- \\
                               & CD    & 0.91 & 0.86 & 0.83 & 0.76 & -- \\
        3 (4 close-ups)        & F@1\% & 81.6 & 81.9 & 83.8 & 84.8 & 86.5 \\
                               & F@2\% & 92.3 & 92.8 & 93.2 & 95.5 & 95.5 \\
                               & CD    & 0.85 & 0.84 & 0.80 & 0.73 & 0.70 \\
        \midrule
        \multicolumn{7}{l}{\textit{Without keeping the unmoved parts (every part regenerated)}} \\
        3 (4 close-ups)        & F@2\% & 92.3 & 93.0 & 92.1 & 95.8 & 96.2 \\
                               & CD    & 0.85 & 0.84 & 0.85 & 0.71 & 0.67 \\
        \midrule
        \multicolumn{7}{l}{\textit{Per-object change from the previous stage}} \\
        All objects            & $\Delta$F@2\% & -- & $+$4.2 & $+$1.0 & $+$1.5 & 0.0 \\
                               & Improved      & -- & 30/40 & 22/40 & 15/24 & 4/12 \\
                \midrule
        \multicolumn{7}{l}{\textit{Kept over later updates (F@2\%)}} \\
                               & & In $\mathcal{A}$ & \multicolumn{2}{c}{After its close-up} & \multicolumn{2}{c}{At the end} \\
        \cmidrule(l){3-3}\cmidrule(l){4-5}\cmidrule(l){6-7}
        Earlier reveals (76)   &       & 89.0 & \multicolumn{2}{c}{94.8} & \multicolumn{2}{c}{94.2} \\
        Unmoved exterior       &       & 97.1 & \multicolumn{2}{c}{--}   & \multicolumn{2}{c}{97.2} \\
        \bottomrule
    \end{tabular*}
\end{table}

\section{Real-World Examples}
\label{sec:supp_real}
This section describes the automatic part masks that all real-world examples use (\cref{sec:supp_automask}), updates of three objects captured with a phone (\cref{sec:supp_phone}), and the robot experiment (\cref{sec:supp_robot}).

\begin{figure*}[t]
  \centering
  \setlength{\tabcolsep}{0.3pt}
  \renewcommand{\arraystretch}{0.5}
  \newcommand{\rcell}[2]{\includegraphics[width=0.13\textwidth]{figures/real/#1_#2_sq.jpg}}
  \newcommand{\rrow}[3]{\rotatebox{90}{\parbox{0.13\textwidth}{\centering\scriptsize #3}} & \rcell{#1}{rest} & \rcell{#1}{A_open} & \rcell{#1}{closeup} & \rcell{#1}{overlay} & \rcell{#1}{#2}}
  \begin{tabular}{c@{\hspace{3pt}}ccccc}
    \rrow{homeoven_c2}{B_new}{Oven, close-up 1} \\
    \rrow{homeoven_c3}{B_new}{Oven, close-up 2} \\
    \rrow{homeoven_c4}{B_new}{Oven, close-up 3} \\
    \rrow{homeoven_c5}{B_new}{Oven, close-up 4} \\
    \rrow{homeoven_c6}{B_new}{Oven, close-up 5} \\
    \rrow{real3_pedestal}{B_open}{Drawer pedestal} \\
    \rrow{slidecab}{B_open}{Sliding doors} \\[1pt]
    & \scriptsize Rest & \scriptsize $\mathcal{A}$ opened & \scriptsize Close-up & \scriptsize Close-up $|$ $\mathcal{B}$ & \scriptsize $\mathcal{B}$ opened
  \end{tabular}
  \vspace{-2mm}
  \caption{Real-world updates with part masks from our automatic pipeline. The five oven rows start from the same rest photo and same $\mathcal{A}$ but update using different close-ups. The last column depicts the generated asset with the revealed part pulled out.}
  \label{fig:supp_real}
\end{figure*}

\begin{figure*}[t]
    \centering
    \includegraphics[width=\textwidth]{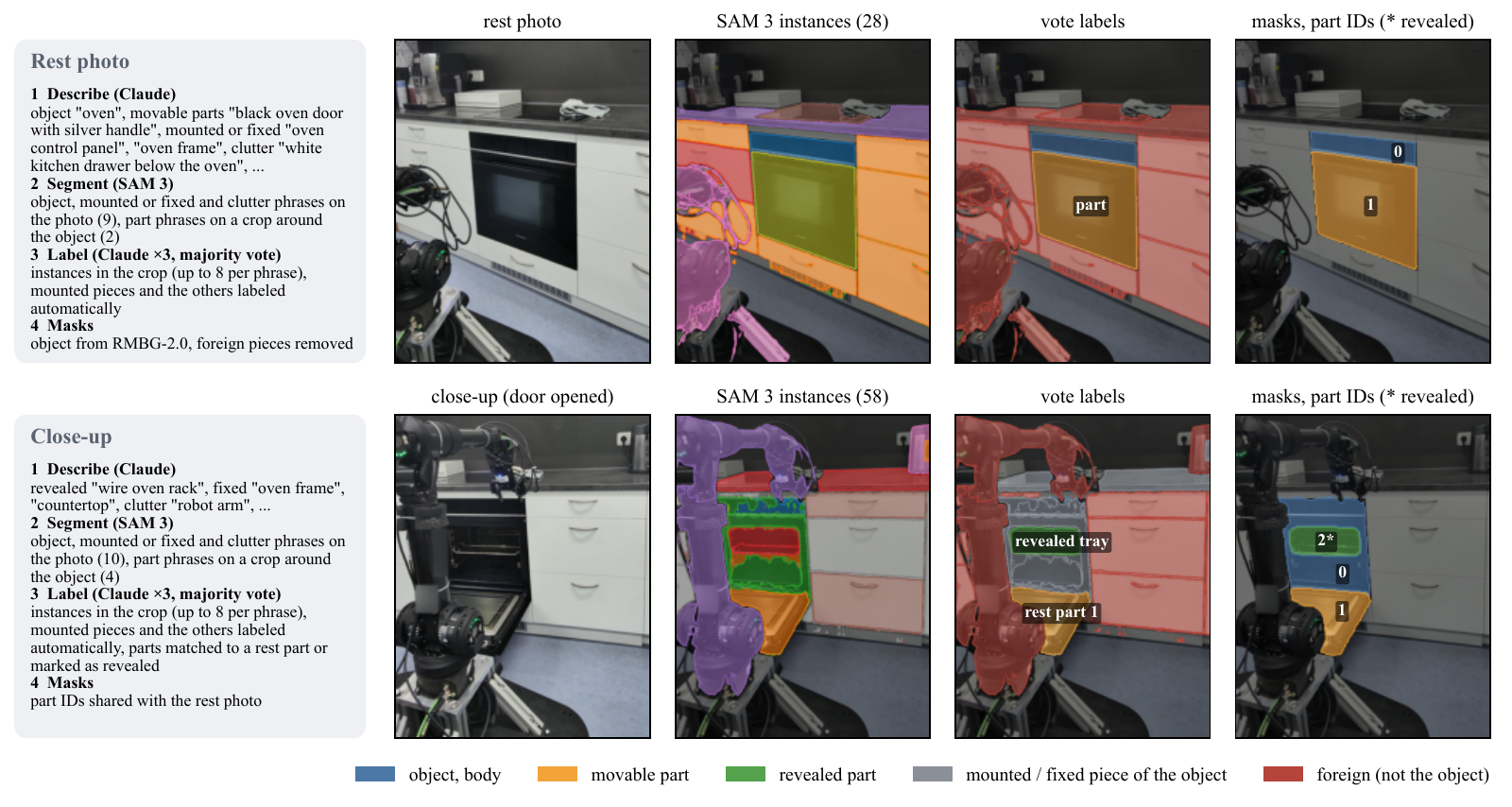}
   \caption{Automatic part masks on the oven of the robot experiment (\cref{sec:supp_robot}), computed during the run on the rest photo (top) and the close-up (bottom). From left, the photo, the instances of SAM~3 colored by phrase, the majority-vote labels, and the final masks with part IDs, where the close-up matches the opened door to rest part 1 and marks the wire rack as revealed. }
     \label{fig:supp_automask}
\end{figure*}

\begin{figure*}[t]
    \par\smallskip\noindent{\small\textbf{Description of the rest photo (once)}}
    {\tiny
        \begin{verbatim}
The photo shows an articulated object (furniture or appliance) at rest. Name things with short phrases that an
open-vocabulary segmenter understands.
For a BUILT-IN appliance (an oven, dishwasher, fridge or microwave built into kitchen units or a wall), THIS object is the
appliance only: the doors, drawers, fronts and panels of the units around, above or below it are not its parts and not part of
its description; list them under clutter (e.g. 'white kitchen drawer below the oven').
Answer with ONE JSON object and nothing else:
{"object": "<the object's category noun, e.g. cabinet, oven, microwave, dresser>",
 "description": "<one sentence that singles out THIS object in the photo, e.g. white two-door cabinet, the right one of two>",
 "parts": ["<phrase for each kind of part of THIS object that moves on its own and is visible in the photo, naming the object and
            its look, e.g. 'white oven door with handle', 'microwave door', 'cabinet door', 'drawer'; describe the WHOLE part, never a
            piece of it such as its glass window or handle>"],
 "part_nouns": ["<the plain noun of each of those parts, e.g. door, drawer, knob, lid>"],
 "mounted": ["<noun for each fixture permanently mounted ON the object that belongs to it, e.g. vessel sink, tap, hob, extractor
             hood; [] if none>"],
 "fixed": ["<short noun for each large FIXED piece of the object visible in the photo, e.g. countertop, oven frame, cabinet side
            panel; at most 2>"],
 "clutter": ["<noun for each of the (at most 6) largest things on, against or right next to the object that are not part of it and
             not mounted on it, e.g. cardboard box, lamp, cabinet (a neighbouring one), cart>"]}
\end{verbatim}
    }
    \par\smallskip\noindent{\small\textbf{Description of the close-up (once, after the rest photo is done)}}
    {\tiny
        \begin{verbatim}
Photo 1 is an earlier rest photo of an articulated object: {description} ({object}); its movable parts were named {parts}.
Photo 2 is a new close-up of the SAME object with one or more parts opened. Name things with short phrases that an
open-vocabulary segmenter understands. Answer with ONE JSON object and nothing else:
{"revealed": [{"phrases": ["<two different phrasings of one kind of item inside that is only visible in photo 2 AND moves on its
              own (slides out or rotates), e.g. 'wire oven rack', 'oven shelf rack' or 'glass turntable plate', 'round glass plate'>"],
               "type": "<one of drawer, shelf, tray, door: tray for racks, pans and turntables>"}],
              [] if none; fixed shelves, walls and rails are NOT listed
 "fixed": ["<short noun for each large FIXED piece of the object visible in photo 2, e.g. countertop, oven frame, cabinet side panel;
            at most 2>"],
 "clutter": ["<noun for each of the (at most 6) largest things on, against or right next to the object in photo 2 that are not
             part of it and not mounted on it, e.g. cardboard box, lamp, cabinet (a neighbouring one), cart>"]}
\end{verbatim}
    }
    \par\smallskip\noindent{\small\textbf{Labels of the rest photo (three answers)}}
    {\tiny
        \begin{verbatim}
TARGET object: {description} ({object}). The first image is the clean rest photo; the next images show the
instances an open-vocabulary segmenter found, one image per concept, each instance tinted and labelled with its ID.
Label EVERY ID ({ids}) on its own line as `<ID>: <LABEL>` with LABEL one of
  PART    - one whole movable part on the OUTSIDE of the TARGET (a door with its frame, a drawer front, a lid, a knob); things seen
            through a glass window or inside are SKIP; not a piece of a part (only the glass
            window, only the handle) and not two parts together; if several instances show the same part, choose the most
            complete one as PART and SKIP the others
  OBJECT  - exactly the whole TARGET object and nothing more (no neighbouring furniture, no items on it)
  FIXED   - a fixed piece of the TARGET (side panel, top, frame, fixed shelf)
  FOREIGN - anything that is not the TARGET: neighbouring furniture or its doors, items on or against it, room doors, walls, floor
  SKIP    - duplicates another ID, a fragment, or covers several things at once (e.g. the target plus its neighbour)
Then one last line `MOVABLE PARTS OF THE TARGET VISIBLE: <n>`.
\end{verbatim}
    }
    \par\smallskip\noindent{\small\textbf{Labels of the close-up (three answers)}}
    {\tiny
        \begin{verbatim}
TARGET object: {description} ({object}). Image 1 is the rest photo with the target's movable parts
R1..R{K} tinted. Image 2 is a later close-up photo of the same object with parts opened. The next images show the instances an
open-vocabulary segmenter found in the close-up, one image per concept, each tinted and labelled with its ID.
Label EVERY ID ({ids}) on its own line as `<ID>: <LABEL>` with LABEL one of
  R<k>       - the SAME physical part as rest part R<k> (opened or still closed), the whole part with its frame and border
               (not only its glass window); if several instances show it, choose the most complete one and SKIP the others
  NEW:<type> - a part NOT visible at rest that moves on its own (slides out or rotates), type one of {new_types}:
               racks, pans, baking trays and microwave turntables are NEW:tray; shelf only for a pull-out shelf; shelves resting on
               fixed supports inside a cabinet are FIXED
  OBJECT     - exactly the whole TARGET object and nothing more
  FIXED      - a fixed piece of the target (walls of the cavity, fixed shelves, bottom, top, frame, rails and runners)
  FOREIGN    - anything that is not the target (neighbouring furniture or its doors, room doors, items, walls, floor)
  SKIP       - duplicates another ID, a fragment, or covers several things at once
Use each R<k> at most once (the best instance). Then one last line `NEW PARTS: <n>`.
\end{verbatim}
    }
    \vspace{-2mm}
    \caption{Prompts of the automatic part masks. The pipeline fills in the fields in braces, such as \texttt{\{object\}}, and sends each prompt with the photos and instance overlays it describes.}
    \label{fig:supp_prompts}
\end{figure*}

\subsection{Automatic Part Masks}
\label{sec:supp_automask}
The pipeline has four stages (\cref{fig:supp_automask}), and its prompts are in \cref{fig:supp_prompts}. 
It runs on the rest photo first, then on the close-up. 
On the rest photo, a vision-language model (VLM, Claude Opus) names the object, its movable parts, and the items around it, SAM~3~\cite{carion2026sam} segments every named item, and the VLM labels each instance inside a crop around the object with the majority of three independent answers. 
On the close-up, the VLM also names the movable items that were hidden at rest, and the same segmentation and labeling follow. 
The labels are the following.

\begin{itemize}
    \setlength{\itemsep}{0pt}
    \item \texttt{PART}, one whole movable part of the object, such as a door with its frame or a drawer front (rest photo only).
    \item \texttt{R1}, \texttt{R2}, \ldots, the same physical part as the part of the rest photo with that number, opened or still closed (close-up only).
    \item \texttt{NEW}, a part hidden at rest that moves on its own, with its type drawer, shelf, tray, or door (close-up only).
    \item \texttt{OBJECT}, the whole object.
    \item \texttt{FIXED}, a fixed piece of the object, such as a side panel, the frame, or a fixed shelf.
    \item \texttt{FOREIGN}, anything that is not the object, such as neighboring furniture, items on the object, walls, or the floor.
    \item \texttt{SKIP}, a duplicate of another instance, a fragment, or several things at once.
\end{itemize}
The code labels some instances itself, without the VLM.
Pieces mounted on the object, such as a control panel, are \texttt{MOUNTED} and joined to the object, instances that lie mostly outside the crop are \texttt{FOREIGN}, and only the eight best-scoring instances of each phrase go to the VLM, while the remaining ones are \texttt{FOREIGN} or \texttt{SKIP}. 
The VLM only names and labels, and the masks are built from the outputs of SAM~3 and of the background removal of BRIA RMBG-2.0~\cite{BiRefNet}, with the body as the object minus its parts. 

We compare the automatic masks with the ground-truth part masks on all 598 cases of RevealBench (\cref{tab:supp_automask}). 
In each photo, the ground-truth parts are matched one to one to the predicted masks by IoU, and the mean IoU is averaged over the parts and then over the photos. 
The association measures whether a part keeps its ID from the rest photo to the close-up. 
Over the parts that are matched with an IoU of at least 0.5 in both photos, it is the share whose two matched masks have the same ID. 
A revealed part counts as found if its bounding box overlaps with an IoU of at least 0.5, since the ground truth of a wire rack covers only its wires. 
On 12 cases, the pipeline finds no object in one photo and stops, and all their parts count as missed. 
The close-up masks are less accurate than the rest masks, mostly in ovens, whose ground truth splits the interior into more racks, trays, and liners than the pipeline predicts.

\begin{table}[t]
    \centering
    \caption{Automatic part masks against the ground-truth masks on all 598 cases of RevealBench. Mean IoU over the parts of each photo, then over the photos. Association is over the parts found in both photos. Stopped cases count as missed.}
    \label{tab:supp_automask}
    \footnotesize
    \setlength{\tabcolsep}{4pt}
    \begin{tabular}{lccc}
        \toprule
        Metric                                 & ArtiVerse & Astra   & All      \\
        \midrule
        Cases (stopped)                        & 326 (6)   & 272 (6) & 598 (12) \\
        Mean IoU, rest photo                   & 0.82      & 0.88    & 0.85     \\
        Mean IoU, close-up                     & 0.74      & 0.72    & 0.73     \\
        Mean IoU, both photos                  & 0.78      & 0.80    & 0.79     \\
        Association                            & 96.7\%    & 92.9\%  & 94.8\%   \\
        Revealed parts found                   & 60.9\%    & 82.1\%  & 69.3\%   \\
        Predicted revealed parts that are real & 82.6\%    & 82.7\%  & 82.6\%   \\
        \bottomrule
    \end{tabular}
\end{table}

\subsection{Phone Captures}
\label{sec:supp_phone}
\cref{fig:supp_real} shows updates of three real objects captured with a phone, with camera poses from ARKit. They go through the registration of \cref{sec:supp_inference}, the standard generation and update, and the automatic part masks of \cref{sec:supp_automask}. 
The five oven updates come from one capture with five close-ups of different interiors, which we rearranged by hand during the capture. They all start from the same $\mathcal{A}$. The pixel-aligned features ground the revealed interior in the close-up, so that each oven update places the rack and the pan at the heights its close-up shows, and the cabinet update replaces the three shelves that $\mathcal{A}$ guesses with the single shelf of the close-up.

\subsection{Robot Experiment}
\label{sec:supp_robot}
\cref{fig:supp_robot} shows \methodname{} in the loop of a mobile manipulator with a 6-DoF arm that opens a real oven and pulls out the rack that is hidden until the door is open. Motion planning is done solely based on the oven asset generated by \methodname{}. \methodname{} runs online on a workstation with one RTX 5090 that is connected to the robot. All three photos, two of the closed oven and one of the opened oven, come from the left and right cameras of a calibrated rig at the rear of the robot, so their poses are known.

Before the oven is opened, both cameras take a rest-state photo. The right photo is used to generate $\mathcal{A}$, and both photos are used by the registration of \cref{sec:supp_inference}, which places $\mathcal{A}$ in the robot's coordinate frame at metric scale. The part masks of all three photos come from the automatic pipeline of \cref{sec:supp_automask} running online during the experiment. Each motion is then planned solely based on the registered asset. The motion planner identifies the part with a revolute joint as the door and places the grasp using a heuristic as the protrusion on its geometry.
Once the opening motion is finished, the right camera takes a close-up of the opened oven, and the close-up updates $\mathcal{A}$ to $\mathcal{B}$. The rack, which $\mathcal{A}$ does not contain, is then generated as a new part with a prismatic joint, and the pull is planned along its predicted slide axis. Both motions are executed open loop under impedance control~\cite{pro2026crisp}. 
The robot video on the project page shows the full run, including the denoising of $\mathcal{A}$ and $\mathcal{B}$ in real time, the registration, and the planned motions.

\section{RevealTrain}
\label{sec:supp_revealtrain}

\begin{figure*}[t]
    \centering
    \includegraphics[width=\textwidth]{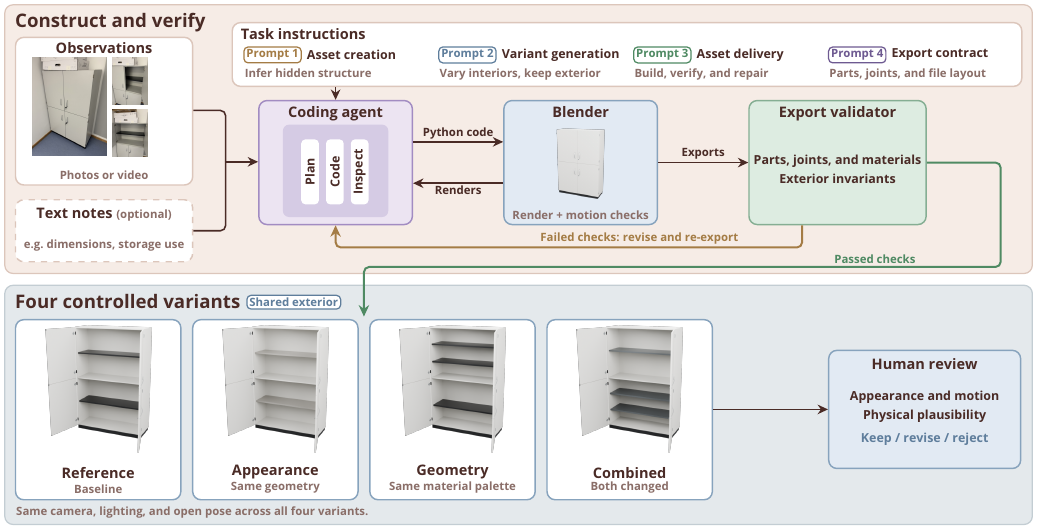}
    \caption{Agentic modeling pipeline. A coding agent receives the reference observations and the task instructions, writes Blender code, inspects the renders, and revises the asset until its exported parts, joints, and materials pass the validator. It then creates variants that share the exterior and differ in the interior, and a human reviews every family.
    The cabinet shown is an example capture and not part of RevealTrain.}
    \label{fig:supp_agent}
\end{figure*}

\begin{figure*}[t]
    \centering
    \includegraphics[width=\textwidth]{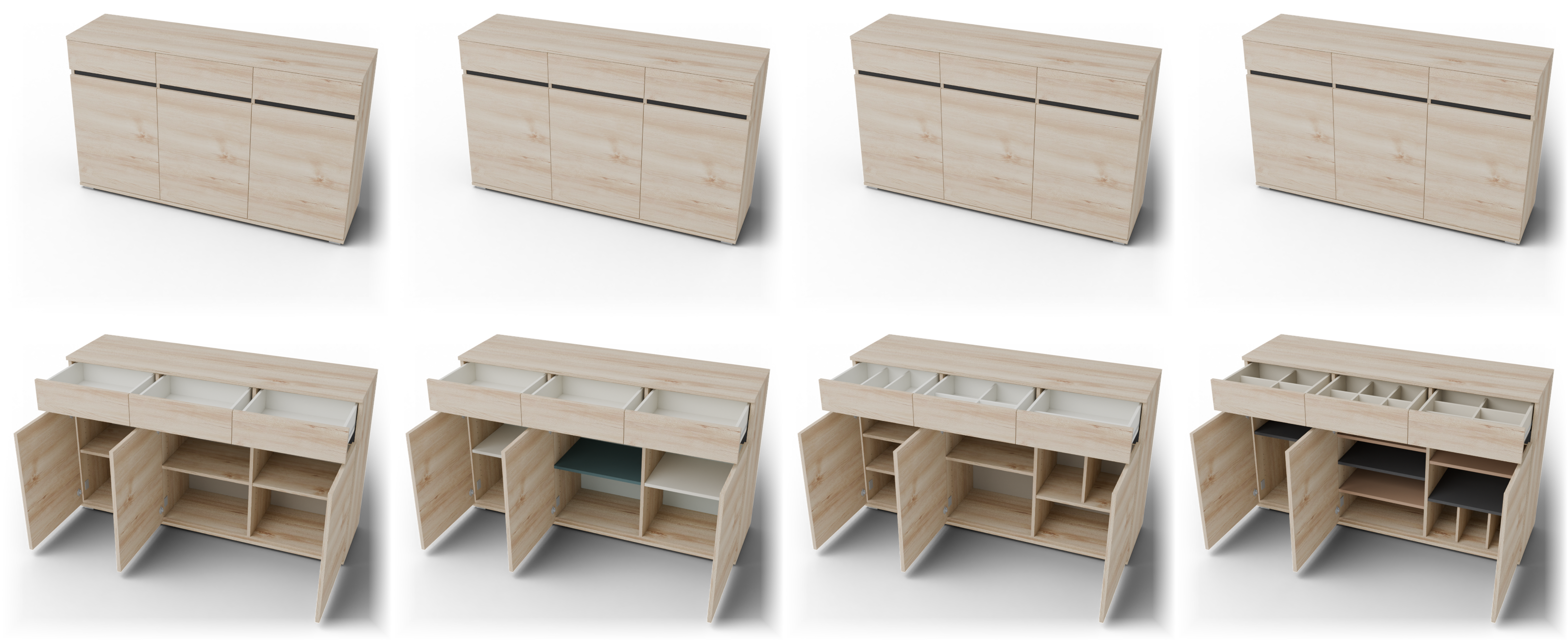}
    \caption{Variants 0 to 3 of a RevealTrain family, closed (top) and opened (bottom).}
    \label{fig:supp_family}
\end{figure*}

\paragraph{Agentic modeling pipeline.}
RevealTrain objects are modeled procedurally in Blender by a coding agent (\cref{fig:supp_agent}) from reference images of Amazon Berkeley Objects (ABO)~\cite{collins2022abo} and PartNet-Mobility (PM)~\cite{xiang2020sapien}. 
An ABO family, one object with its interior variants, starts from 1 to 12 photos (median 8), and a PM family from four rest-state renders. 
No source mesh is imported, and for PM the agent sees only the renders and, as hints, the part annotations. 
A human shortlists the ABO objects. Families whose ABO object also appears in ArtiVerse~\cite{iliash2026artiverse} or ACD~\cite{iliash2026s2o} go to RevealBench-Astra, so RevealTrain shares no object with either dataset. 
The category is one of the seven PM categories of SINGAPO~\cite{liu2025singapo}, of which the training split covers six (\cref{tab:supp_stats}). 
For each family, the agent (GPT-6 Astra in the Codex CLI) receives the images and a task description with the export format and the variant rules, together with an example builder of a template object, whose helper functions most builders reuse. 
It writes a single Blender Python script that builds the object with its parts, joints, and materials, runs it headless, compares renders of the result with the reference images, and exports per-part meshes and textures together with a part tree that stores the type, axis, and range of every joint.
Textures are procedural, made with the agent's built-in image generation, or cropped from the ABO photos, and no assets were downloaded during the jobs.
The agent also writes a check specific to the asset that sweeps the parts through their articulation range for collisions. A validator checks the part hierarchy, the joint types and ranges, the texture paths and brightness, and that the exterior meshes, materials, and textures are byte-identical across the variants of a family, and it requires the collision check of the agent to pass. 
The agent runs the validator during its session until it passes, typically four times, and fixes a failure in 34 of the 100 training families. Every family then passes the final validation without further repair, and a human reviews every family and rejected none.
A family takes 15 to 88 minutes of agent time (median 35).

\paragraph{Families and variants.}
A family has four variants, or twelve for 25 of the training families. Variant 0 matches the reference images, and the others keep the exterior and change the interior, such as the shelves, dividers, drawer partitions, and interior finishes. Some variants also change the movable parts behind the same closed front, for example a dishwasher whose pull-out drawer becomes a drop-down door. The training split thus has 600 objects. \cref{fig:supp_family} shows an example family.

\paragraph{Statistics and splits.}
The training split contains 100 families, 91 modeled from ABO references and 9 from PM, with 63 families of storage furniture, 16 tables, 7 microwaves, 5 ovens, 5 dishwashers, and 4 refrigerators. 
RevealTrain and RevealBench-Astra share no family, no ABO object, and no PM object.
The 600 training objects have 2678 parts, 600 bodies and 2078 movable parts with 921 revolute and 1157 prismatic joints (\cref{tab:supp_stats}).
They are rendered in 12000 rest-state views, 48590 opened close-ups of 4859 states, and 3216 reveal close-ups of 201 states.

\begin{table}[t]
    \centering
    \caption{Training split of RevealTrain, with families (Fam.), objects (Obj.), parts, parts per object, and revolute (Rev.) and prismatic (Pris.) joints. Parts count the body and every movable part, and each movable part has one joint. Parts per object is the median, with the range in parentheses.}
    \label{tab:supp_stats}
    \footnotesize
    \setlength{\tabcolsep}{3.5pt}
    \begin{tabular}{lrrrcrr}
        \toprule
        Category      & Fam. & Obj. & Parts & Per obj.  & Rev. & Pris. \\
        \midrule
        Storage furn. & 63   & 396  & 1712  & 4 (2--7)  & 655  & 661   \\
        Table         & 16   & 120  & 502   & 4 (2--6)  & 47   & 335   \\
        Microwave     & 7    & 28   & 130   & 5 (4--5)  & 91   & 11    \\
        Oven          & 5    & 20   & 121   & 6 (4--7)  & 59   & 42    \\
        Dishwasher    & 5    & 20   & 82    & 4 (2--5)  & 25   & 37    \\
        Refrigerator  & 4    & 16   & 131   & 8 (4--14) & 44   & 71    \\
        \midrule
        All           & 100  & 600  & 2678  & 4 (2--14) & 921  & 1157  \\
        \bottomrule
    \end{tabular}
\end{table}

\section{Full Rest-State Generation Results}
\label{sec:supp_rest}

\subsection{Metrics and Baselines}
\label{sec:supp_rest_protocol}
\cref{tab:supp_pm} and \cref{tab:supp_acd} give all metrics of the rest-state generation results summarized in \cref{tab:main_rest}, including the centroid distance dcDist, the overlap ratio between moving sibling parts (AOR), and the CLIP similarity of the rendered appearance, which is not available for URDF-Anything+ and SPARK since they output no texture.

\label{sec:supp_baselines}
\paragraph{Rest-state metrics.}
We use the official metric code of SINGAPO~\cite{liu2025singapo} unchanged, where AS averages over the states at 0, 25, 50, 75, and 100\% of every joint range and RS is the first of them. Since only rigidly moving parts matter, parts connected by a fixed joint are merged into one part, both in the ground truth, as in PAct~\cite{liu2026pact}, and in the prediction of every baseline.

\paragraph{Baselines.}
All methods receive the same input renders, and we describe only where we depart from their released defaults. SINGAPO receives the topology of the ground-truth part graph, and we sample it once instead of five times. PWM-ArtGen receives the ground-truth graph restricted to the given parts and the part masks. Articulate-Anything uses GPT-6 Astra as its vision-language model and runs as in its released image-input example. URDF-Anything+ has its manual orientation step answered automatically from the camera azimuth. SPARK uses GPT-6 Astra in place of its OpenAI and Gemini calls and skips texturing. Particulate, paired with Hunyuan3D in its paper, is paired with TRELLIS. SPARK fails on 2 ACD inputs, which are left out of its means.

\begin{table*}[t]
    \centering

    \caption{Rest-state image-to-3D articulated object generation results on PM~\cite{xiang2020sapien} with SINGAPO~\cite{liu2025singapo} test split. Geometry metrics are given $\times 100$.}
    \label{tab:supp_pm}
    \scalebox{0.85}{%
        \begin{tabular}{lcccccccc}
            \toprule
                                                                                      & \multicolumn{7}{c}{Geometry} & \multicolumn{1}{c}{Appear.}                                                                                                                                     \\
            \cmidrule(lr){2-8} \cmidrule(lr){9-9}
            Method                                                                    & RS-dgIoU $\downarrow$        & AS-dgIoU $\downarrow$       & RS-dcDist $\downarrow$ & AS-dcDist $\downarrow$ & RS-dCD $\downarrow$ & AS-dCD $\downarrow$ & AOR $\downarrow$ & CLIP $\uparrow$  \\
            \midrule
            \multicolumn{9}{l}{\textit{Retrieval-based}}                                                                                                                                                                                                                               \\
            SINGAPO~\cite{liu2025singapo}                                             & \rthird{17.12}               & \rthird{19.41}              & \rthird{1.78}          & \rsecond{3.89}         & \rthird{0.49}       & \rthird{2.20}       & \rsecond{0.23}   & 0.9034           \\
            Articulate-Anything~\cite{le2025articulate}                               & 28.93                        & 31.22                       & 4.43                   & 8.65                   & 2.99                & 6.68                & 0.59             & 0.9318           \\
            PWM-ArtGen~\cite{zheng2026pwm}                                            & 48.58                        & 49.58                       & 6.59                   & 10.08                  & 1.88                & 4.86                & 1.98             & 0.9172           \\
            \midrule
            \multicolumn{9}{l}{\textit{Image-only generation}}                                                                                                                                                                                                                         \\
            URDF-Anything+~\cite{wu2026urdf}                                          & 61.38                        & 62.05                       & 13.42                  & 19.76                  & 5.63                & 11.72               & 12.99            & --               \\
            PhysX-Anything~\cite{cao2025physxanything}                                & 45.53                        & 46.56                       & 7.97                   & 13.98                  & 8.98                & 14.64               & 0.46             & 0.9105           \\
            TRELLIS~\cite{xiang2025structured} + Particulate~\cite{li2026particulate} & 51.94                        & 52.93                       & 12.45                  & 16.55                  & 7.40                & 11.33               & 2.74             & 0.9287           \\
            SPARK~\cite{he2026spark}                                                  & 34.43                        & 35.56                       & 4.82                   & 18.77                  & 1.96                & 11.45               & 8.07             & --               \\
            \midrule
            \multicolumn{9}{l}{\textit{Part-conditioned generation}}                                                                                                                                                                                                                   \\
            \multicolumn{9}{l}{\textit{(image + part masks)}}                                                                                                                                                                                                                          \\
            PAct~\cite{liu2026pact}                                                   & 18.65                        & 20.53                       & 2.29                   & \rthird{4.79}          & 0.82                & 2.82                & 3.65             & \rthird{0.9491}  \\
            \methodname{}-PM                                                          & \rsecond{15.01}              & \rfirst{16.70}              & \rsecond{1.38}         & \rfirst{3.18}          & \rsecond{0.39}      & \rsecond{1.96}      & \rfirst{0.13}    & \rfirst{0.9622}  \\
            \methodname{} (Ours)                                                      & \rfirst{14.97}               & \rsecond{16.92}             & \rfirst{1.27}          & \rfirst{3.18}          & \rfirst{0.38}       & \rfirst{1.79}       & \rthird{0.37}    & \rsecond{0.9617} \\
            \bottomrule
        \end{tabular}
    }

    \vspace{10pt}

    \caption{Rest-state image-to-3D articulated object generation results on ACD Dataset~\cite{iliash2026s2o} with SINGAPO~\cite{liu2025singapo} test split. Geometry metrics are given $\times 100$.}
    \label{tab:supp_acd}
    \scalebox{0.85}{%
        \begin{tabular}{lcccccccc}
            \toprule
                                                                                      & \multicolumn{7}{c}{Geometry} & \multicolumn{1}{c}{Appear.}                                                                                                                                     \\
            \cmidrule(lr){2-8} \cmidrule(lr){9-9}
            Method                                                                    & RS-dgIoU $\downarrow$        & AS-dgIoU $\downarrow$       & RS-dcDist $\downarrow$ & AS-dcDist $\downarrow$ & RS-dCD $\downarrow$ & AS-dCD $\downarrow$ & AOR $\downarrow$ & CLIP $\uparrow$  \\
            \midrule
            \multicolumn{9}{l}{\textit{Retrieval-based}}                                                                                                                                                                                                                               \\
            SINGAPO~\cite{liu2025singapo}                                             & 62.67                        & 63.60                       & 9.78                   & 12.00                  & 4.78                & 6.78                & 1.93             & 0.8689           \\
            Articulate-Anything~\cite{le2025articulate}                               & 76.19                        & 77.19                       & 13.00                  & 17.19                  & 7.06                & 10.46               & \rthird{1.17}    & 0.8810           \\
            PWM-ArtGen~\cite{zheng2026pwm}                                            & 67.29                        & 68.43                       & 8.32                   & 10.96                  & 6.20                & 7.23                & 3.06             & 0.8720           \\
            \midrule
            \multicolumn{9}{l}{\textit{Image-only generation}}                                                                                                                                                                                                                         \\
            URDF-Anything+~\cite{wu2026urdf}                                          & 75.05                        & 75.60                       & 14.58                  & 18.54                  & 7.75                & 11.52               & 9.99             & --               \\
            PhysX-Anything~\cite{cao2025physxanything}                                & 67.90                        & 68.50                       & 13.60                  & 19.66                  & 15.63               & 24.69               & \rsecond{1.11}   & 0.8972           \\
            TRELLIS~\cite{xiang2025structured} + Particulate~\cite{li2026particulate} & 75.69                        & 76.29                       & 16.61                  & 20.94                  & 8.98                & 13.58               & 2.63             & 0.9361           \\
            SPARK~\cite{he2026spark}                                                  & 69.31                        & 70.18                       & 10.98                  & 18.48                  & 5.27                & 11.00               & 6.59             & --               \\
            \midrule
            \multicolumn{9}{l}{\textit{Part-conditioned generation}}                                                                                                                                                                                                                   \\
            \multicolumn{9}{l}{\textit{(image + part masks)}}                                                                                                                                                                                                                          \\
            PAct~\cite{liu2026pact}                                                   & \rthird{44.19}               & \rthird{45.91}              & \rthird{5.65}          & \rthird{8.95}          & \rthird{2.91}       & \rthird{4.51}       & 2.75             & \rthird{0.9423}  \\
            \methodname{}-PM                                                          & \rfirst{32.84}               & \rfirst{35.14}              & \rsecond{2.54}         & \rsecond{5.73}         & \rsecond{1.19}      & \rsecond{2.09}      & 1.30             & \rsecond{0.9474} \\
            \methodname{} (Ours)                                                      & \rsecond{33.82}              & \rsecond{35.57}             & \rfirst{2.49}          & \rfirst{4.73}          & \rfirst{1.14}       & \rfirst{1.80}       & \rfirst{0.82}    & \rfirst{0.9497}  \\
            \bottomrule
        \end{tabular}
    }

\end{table*}

\begin{table}[t]
    \centering
    \caption{Binding parts to masks in the sparse-structure stage, with the masks in the default and in a random order. Duplicates is the number of inputs with two parts at an F-score above 0.8. Changed is the percentage of masks whose generated part differs between the two orders. Two parts count as the same if they match with an F-score of at least 0.5, where voxels within 1 or 3 voxels of each other count as matched.}
    \label{tab:supp_binding}
    \footnotesize
    \setlength{\tabcolsep}{4pt}
    \begin{tabular}{llcccc}
        \toprule
                &                         & \multicolumn{2}{c}{Duplicates$\downarrow$} & \multicolumn{2}{c}{Changed (\%)$\downarrow$} \\
        \cmidrule(lr){3-4} \cmidrule(lr){5-6}
        Set     & Method                  & default & random & 1 vox. & 3 vox. \\
        \midrule
        PM      & PAct~\cite{liu2026pact} & 9 & 7 & 27.7 & 25.1 \\
                & \methodname{}           & \textbf{0} & \textbf{0} & \textbf{4.9} & \textbf{0.9} \\
        \midrule
        ACD     & PAct~\cite{liu2026pact} & 40 & 43 & 48.8 & 45.6 \\
                & \methodname{}           & \textbf{0} & \textbf{0} & \textbf{3.6} & \textbf{0.4} \\
        \bottomrule
    \end{tabular}
\end{table}

\begin{figure}[t]
    \centering
    \setlength{\tabcolsep}{0.3pt}
\renewcommand{\arraystretch}{0.5}
\begin{tabular}{ccccc}
  & \multicolumn{2}{c}{\scriptsize\renewcommand{\arraystretch}{0.9}\begin{tabular}[b]{@{}c@{}}normal order\end{tabular}} & \multicolumn{2}{c}{\scriptsize\renewcommand{\arraystretch}{0.9}\begin{tabular}[b]{@{}c@{}}shuffled order\end{tabular}} \\[2pt]
  & \scriptsize\renewcommand{\arraystretch}{0.9}\begin{tabular}[b]{@{}c@{}}input (IDs)\end{tabular} & \scriptsize\renewcommand{\arraystretch}{0.9}\begin{tabular}[b]{@{}c@{}}output\end{tabular} & \scriptsize\renewcommand{\arraystretch}{0.9}\begin{tabular}[b]{@{}c@{}}input (IDs)\end{tabular} & \scriptsize\renewcommand{\arraystretch}{0.9}\begin{tabular}[b]{@{}c@{}}output\end{tabular} \\[2pt]
  \rotatebox[origin=c]{90}{\scriptsize Ours}\hspace{2pt} & \raisebox{-0.5\height}{\includegraphics[width=0.23\linewidth]{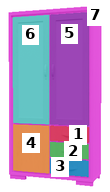}} & \raisebox{-0.5\height}{\includegraphics[width=0.23\linewidth]{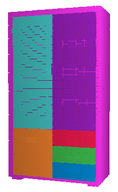}} & \raisebox{-0.5\height}{\includegraphics[width=0.23\linewidth]{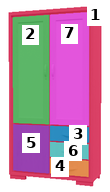}} & \raisebox{-0.5\height}{\includegraphics[width=0.23\linewidth]{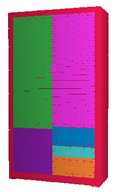}} \\[2pt]
  \rotatebox[origin=c]{90}{\scriptsize PAct}\hspace{2pt} & \raisebox{-0.5\height}{\includegraphics[width=0.23\linewidth]{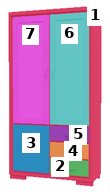}} & \raisebox{-0.5\height}{\includegraphics[width=0.23\linewidth]{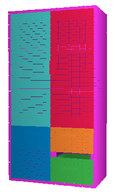}} & \raisebox{-0.5\height}{\includegraphics[width=0.23\linewidth]{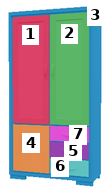}} & \raisebox{-0.5\height}{\includegraphics[width=0.23\linewidth]{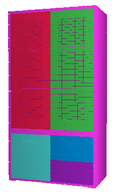}} \\[2pt]
\end{tabular}

    \caption{Binding parts to masks on an ACD bookcase, with the masks in the default and in a random order. For each method and order, the masks (left) and the generated parts (right) are colored by part ID. The parts of \methodname{} follow the masks in both orders, while those of PAct do not.}
    \label{fig:supp_binding}
\end{figure}

\subsection{Binding Parts to Masks}
\label{sec:supp_binding}
PAct also conditions on the part masks, yet \methodname{} outperforms it clearly (\cref{tab:main_rest}). We found that PAct sometimes duplicates parts. In the final meshes of the rest-state results (\cref{tab:supp_pm} and \cref{tab:supp_acd}), PAct generates two near-identical parts on 7 of the 154 PM and 33 of the 270 ACD inputs, while neither of our models does. A likely cause is a weak binding between parts and masks. We therefore run the sparse-structure stage of both methods twice on every input with the same seed, once with the masks in the default order and once in a random order (\cref{tab:supp_binding} and \cref{fig:supp_binding}). When the masks are reordered, the parts should move with them. For PAct, 27.7\% of the PM masks and 48.8\% of the ACD masks get a different part, so its parts do not follow the masks. For \methodname{}, only 4.9\% and 3.6\% do, almost all of them the same part shifted by a few voxels in depth.

\section{Ablations}
\label{sec:supp_ablations}

\subsection{Runtime}
\label{sec:supp_runtime}
We report the median time per input on one RTX 5090 with model loading excluded in \cref{tab:supp_runtime_rest} and \cref{tab:supp_runtime_update}.
In rest-state generation, \methodname{} and PAct take about as long to generate and decode, 12.4 and 11.2~s on PM and 19.3 and 14.6~s on ACD.
The articulation head takes about 1~ms.
The update is timed on 8 RevealBench cases, the same for every method.
We re-timed the updates of all methods on the same workstation, so these medians differ slightly from \cref{sec:exp}.
An update of \methodname{}, which generates $\mathcal{A}$, registers it to the cameras, updates it, and exports $\mathcal{B}$, takes a median of 114~s, and the registration, a mask fit followed by RoMa and PnP, takes 5.6~s.
The update sampling grows with the number of parts, from 12~s for three parts to 112~s for sixteen.
The timed update cases have more parts than RevealBench on average (6.5 against 4.9).

\begin{table}[!tb]
    \centering
    \caption{Rest-state generation runtime in seconds. Gen.\ includes decoding and articulation prediction, Bake is the texture bake, and Total also includes I/O. $^\dagger$Decoder on an A100, with loading. Articulate-Anything and SPARK are dominated by API calls.}
    \label{tab:supp_runtime_rest}
    \footnotesize
    \setlength{\tabcolsep}{4pt}
    \resizebox{\columnwidth}{!}{%
        \begin{tabular}{lcccccc}
            \toprule
                                                                 & \multicolumn{3}{c}{PM} & \multicolumn{3}{c}{ACD}                               \\
            \cmidrule(lr){2-4} \cmidrule(lr){5-7}
            Method                                               & Gen.                   & Bake                    & Total & Gen. & Bake & Total \\
            \midrule
            SINGAPO~\cite{liu2025singapo}                        & -- & -- & \rsecond{2.8} & -- & -- & \rsecond{3.0}  \\
            Articulate-Anything~\cite{le2025articulate}          & -- & -- & 498 & -- & -- & 507  \\
            PWM-ArtGen~\cite{zheng2026pwm}                       & -- & -- & \rfirst{2.0} & -- & -- & \rfirst{2.4}  \\
            \midrule
            URDF-Anything+~\cite{wu2026urdf}                     & 37.8 & -- & 39.0 & 38.9 & -- & 40.2  \\
            PhysX-Anything$^\dagger$~\cite{cao2025physxanything} & -- & -- & 425 & -- & -- & 548  \\
            TRELLIS + Particulate                                & \rfirst{6.5} & \rfirst{11.5} & \rthird{21.0} & \rfirst{6.6} & \rfirst{11.9} & \rthird{21.1}  \\
            SPARK~\cite{he2026spark}                             & -- & -- & 471 & -- & -- & 642  \\
            \midrule
            PAct~\cite{liu2026pact}                          & \rsecond{11.2} & \rthird{38.7} & 51.7 & \rsecond{14.6} & \rthird{48.7} & 65.1  \\
            \methodname{}                                    & \rthird{12.4} & \rsecond{14.0} & 27.5 & \rthird{19.3} & \rsecond{17.9} & 36.9  \\
            \bottomrule
        \end{tabular}%
    }
\end{table}

\begin{table}[!tb]
    \centering
    \caption{Update runtime on RevealBench in seconds. $\mathcal{A}$ is the generation of the initial asset, Reg.\ the registration, Update the SS and SLat flow sampling, Bake the texture bakes of $\mathcal{A}$ and $\mathcal{B}$, and Total also includes decoding and I/O. The inpainting baselines are given the opened part and need no registration.}
    \label{tab:supp_runtime_update}
    \footnotesize
    \setlength{\tabcolsep}{4pt}
    \resizebox{\columnwidth}{!}{%
        \begin{tabular}{lccccc}
            \toprule
            Method                              & $\mathcal{A}$ & Reg. & Update & Bake & Total \\
            \midrule
            RePaint~\cite{lugmayr2022repaint}   & \rsecond{33.3} & --   & \rsecond{27.3} & 52.0 & \rthird{122.5}  \\
            VoxHammer~\cite{li2026voxhammer}    & \rsecond{33.3} & --   & \rfirst{26.8} & \rsecond{50.2} & \rsecond{121.0}  \\
            VoxHammer+KV~\cite{li2026voxhammer} & \rthird{33.4} & --   & 60.5 & \rthird{50.7} & 156.1  \\
            \midrule
            \methodname{}                       & \rfirst{31.4} & 5.6  & \rthird{28.8} & \rfirst{48.2} & \rfirst{114.4}  \\
            \bottomrule
        \end{tabular}%
    }
\end{table}

\subsection{Aggregating the Articulation-Head Input}
\label{sec:supp_agg}
The articulation head reads the part tokens of the last sampling step (\cref{sec:base}). 
PAct~\cite{liu2026pact} instead averages them over the last 20 of its 25 sampling steps. \cref{tab:supp_agg} feeds the head the part tokens averaged over the last $N$ steps of a single sampling run, so the geometry is identical for every $N$ and only the predicted joints change. Averaging halves the median axis error, but the joint types are essentially unchanged, and the gains in the part-level metrics are small, so we keep $N=1$ in all other experiments.

\begin{table}[!tb]
    \centering
    \caption{Joint accuracy with the articulation-head input averaged over the last $N$ of 25 sampling steps. $^\dagger$ marks PAct's setting.}
    \label{tab:supp_agg}
    \footnotesize
    \setlength{\tabcolsep}{6pt}
    \begin{tabular}{llcccc}
        \toprule
              &              & Type                 & \multicolumn{2}{c}{Axis err.\,(\textdegree)$\downarrow$} & Pivot                    \\
        \cmidrule(lr){4-5}
        Split & $N$          & acc.\,(\%)$\uparrow$ & median                                                   & mean  & err.$\downarrow$ \\
        \midrule
        PM    & 1            & 97.6                 & 0.31                                                     & 2.41  & 0.0284           \\
        PM    & 2            & 97.6                 & 0.24                                                     & 2.30  & 0.0290           \\
        PM    & 3            & 97.9                 & 0.22                                                     & 2.33  & 0.0296           \\
        PM    & 5            & 97.9                 & 0.18                                                     & 2.38  & 0.0276           \\
        PM    & 10           & 97.9                 & 0.16                                                     & 2.32  & 0.0255           \\
        PM    & 20$^\dagger$ & 98.1                 & 0.14                                                     & 2.24  & 0.0241           \\
        PM    & 25           & 98.1                 & 0.14                                                     & 2.08  & 0.0235           \\
        \midrule
        ACD   & 1            & 95.4                 & 0.34                                                     & 4.25  & 0.0535           \\
        ACD   & 2            & 95.4                 & 0.28                                                     & 3.73  & 0.0510           \\
        ACD   & 3            & 95.4                 & 0.24                                                     & 3.51  & 0.0492           \\
        ACD   & 5            & 95.3                 & 0.20                                                     & 3.40  & 0.0482           \\
        ACD   & 10           & 95.2                 & 0.18                                                     & 3.37  & 0.0488           \\
        ACD   & 20$^\dagger$ & 95.2                 & 0.15                                                     & 3.21  & 0.0498           \\
        ACD   & 25           & 95.3                 & 0.16                                                     & 3.12  & 0.0495           \\
        \bottomrule
    \end{tabular}
\end{table}

\FloatBarrier
\renewcommand{\dbltopfraction}{0.95}\renewcommand{\textfraction}{0.05}\setcounter{dbltopnumber}{2}  %
\section{Additional Results}
\label{sec:supp_results}
\begin{figure*}[tp]
    \centering
    \setlength{\tabcolsep}{0.3pt}
\renewcommand{\arraystretch}{0.5}
\begin{tabular}{cccccccccc}
   & \scriptsize\renewcommand{\arraystretch}{0.9}\begin{tabular}[b]{@{}c@{}}Input\end{tabular} & \scriptsize\renewcommand{\arraystretch}{0.9}\begin{tabular}[b]{@{}c@{}}SINGAPO\end{tabular} & \scriptsize\renewcommand{\arraystretch}{0.9}\begin{tabular}[b]{@{}c@{}}Articulate-\\Anything\end{tabular} & \scriptsize\renewcommand{\arraystretch}{0.9}\begin{tabular}[b]{@{}c@{}}PWM-\\ArtGen\end{tabular} & \scriptsize\renewcommand{\arraystretch}{0.9}\begin{tabular}[b]{@{}c@{}}PhysX-\\Anything\end{tabular} & \scriptsize\renewcommand{\arraystretch}{0.9}\begin{tabular}[b]{@{}c@{}}TRELLIS+\\Particulate\end{tabular} & \scriptsize\renewcommand{\arraystretch}{0.9}\begin{tabular}[b]{@{}c@{}}PAct\end{tabular} & \scriptsize\renewcommand{\arraystretch}{0.9}\begin{tabular}[b]{@{}c@{}}Ours\end{tabular} & \scriptsize\renewcommand{\arraystretch}{0.9}\begin{tabular}[b]{@{}c@{}}GT\end{tabular} \\[2pt]
   & \raisebox{-0.5\height}{\includegraphics[width=0.08\textwidth]{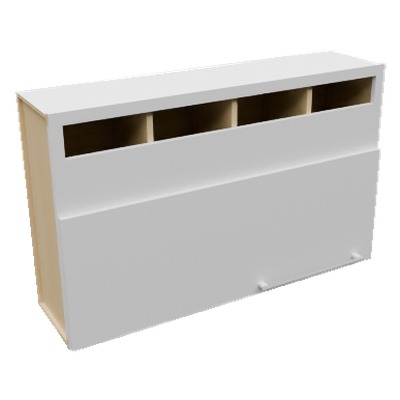}} & \raisebox{-0.5\height}{\includegraphics[width=0.08\textwidth]{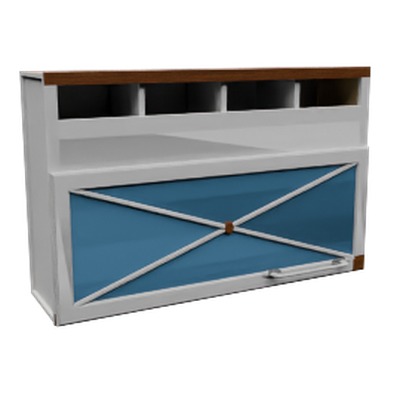}} & \raisebox{-0.5\height}{\includegraphics[width=0.08\textwidth]{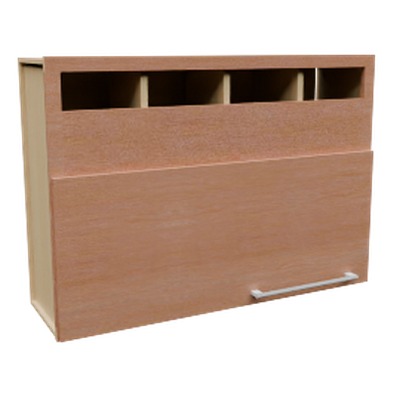}} & \raisebox{-0.5\height}{\includegraphics[width=0.08\textwidth]{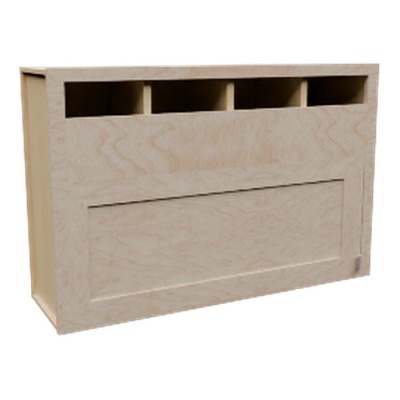}} & \raisebox{-0.5\height}{\includegraphics[width=0.08\textwidth]{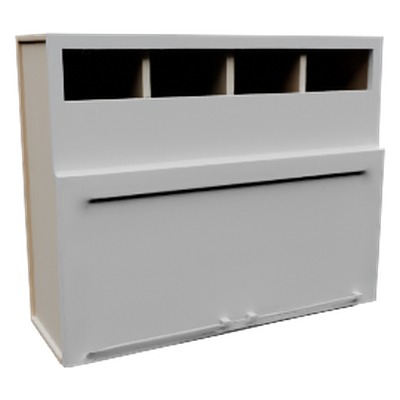}} & \raisebox{-0.5\height}{\includegraphics[width=0.08\textwidth]{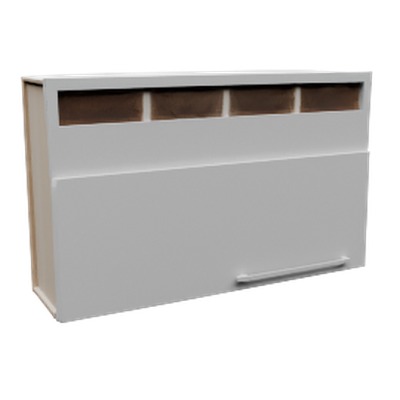}} & \raisebox{-0.5\height}{\includegraphics[width=0.08\textwidth]{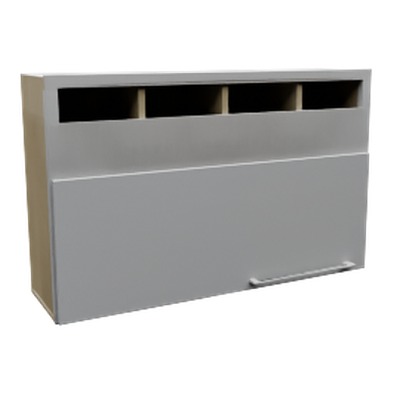}} & \raisebox{-0.5\height}{\includegraphics[width=0.08\textwidth]{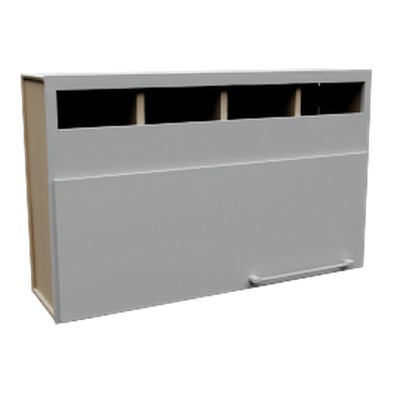}} & \raisebox{-0.5\height}{\includegraphics[width=0.08\textwidth]{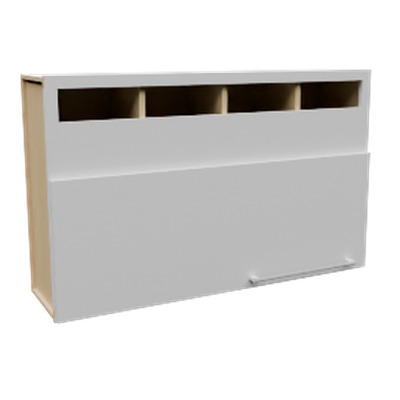}} \\
  \rotatebox[origin=c]{90}{\scriptsize PM}\hspace{2pt} & \raisebox{-0.5\height}{\includegraphics[width=0.08\textwidth]{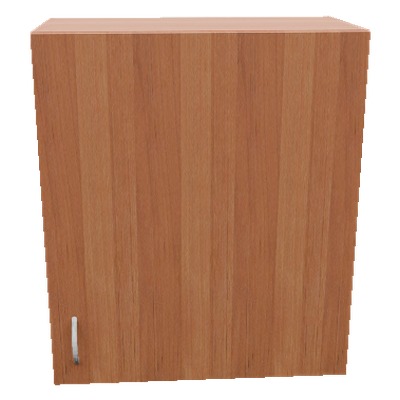}} & \raisebox{-0.5\height}{\includegraphics[width=0.08\textwidth]{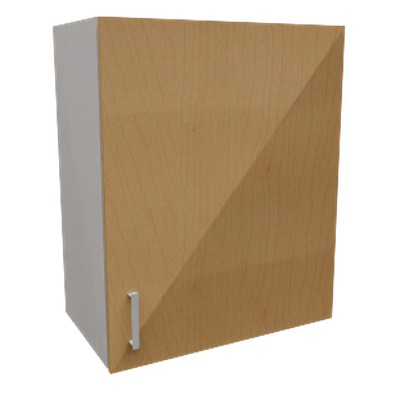}} & \raisebox{-0.5\height}{\includegraphics[width=0.08\textwidth]{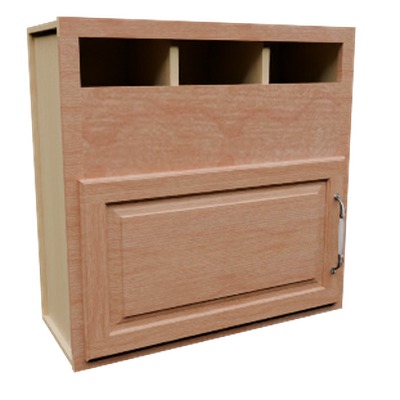}} & \raisebox{-0.5\height}{\includegraphics[width=0.08\textwidth]{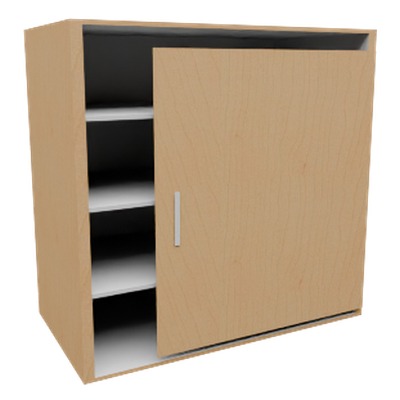}} & \raisebox{-0.5\height}{\includegraphics[width=0.08\textwidth]{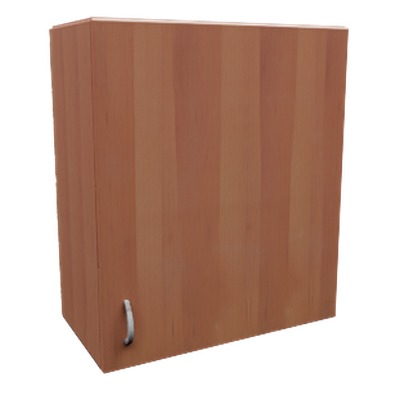}} & \raisebox{-0.5\height}{\includegraphics[width=0.08\textwidth]{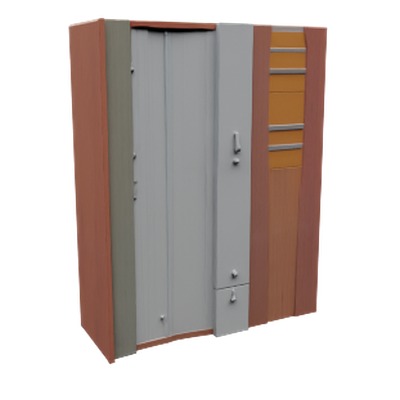}} & \raisebox{-0.5\height}{\includegraphics[width=0.08\textwidth]{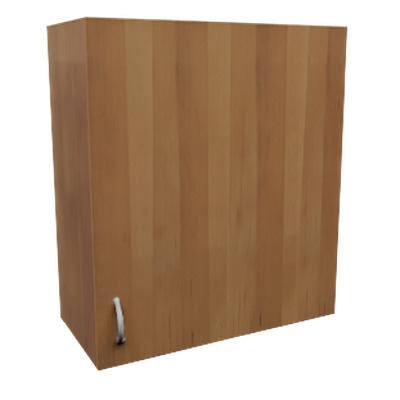}} & \raisebox{-0.5\height}{\includegraphics[width=0.08\textwidth]{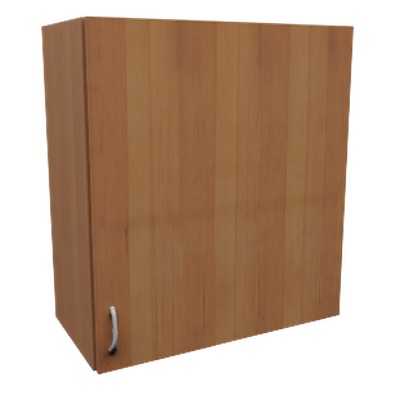}} & \raisebox{-0.5\height}{\includegraphics[width=0.08\textwidth]{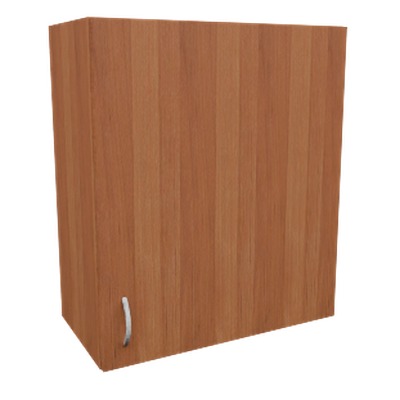}} \\
   & \raisebox{-0.5\height}{\includegraphics[width=0.08\textwidth]{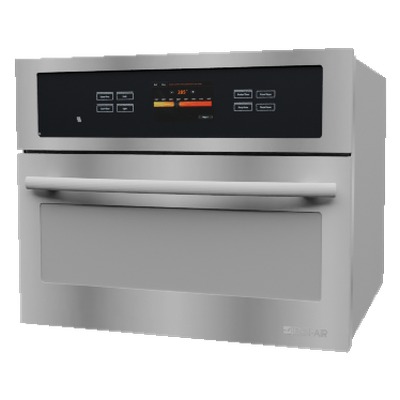}} & \raisebox{-0.5\height}{\includegraphics[width=0.08\textwidth]{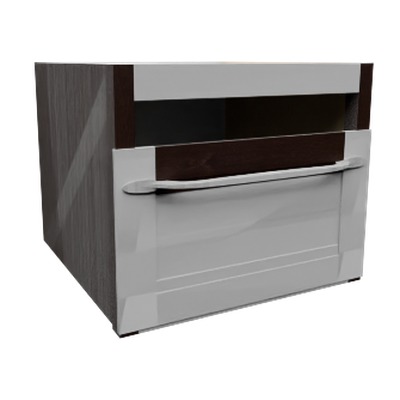}} & \raisebox{-0.5\height}{\includegraphics[width=0.08\textwidth]{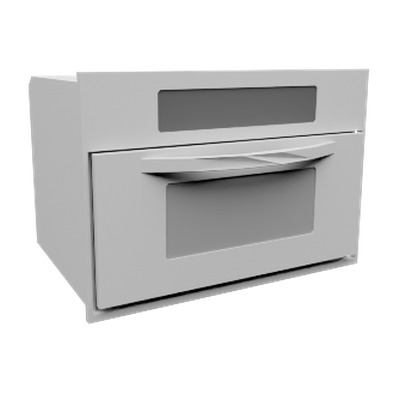}} & \raisebox{-0.5\height}{\includegraphics[width=0.08\textwidth]{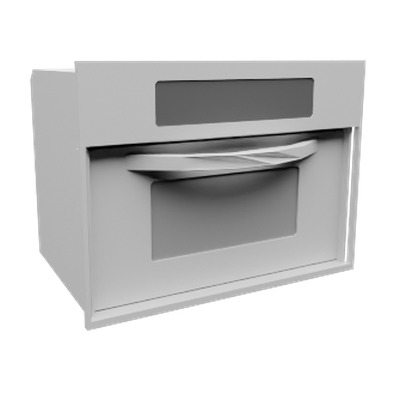}} & \raisebox{-0.5\height}{\includegraphics[width=0.08\textwidth]{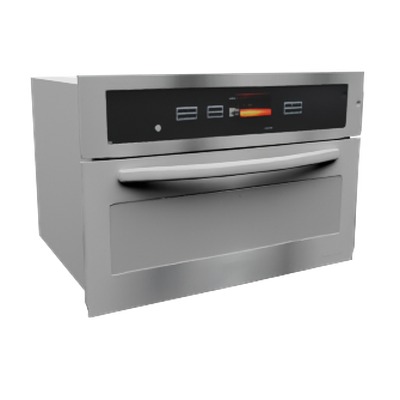}} & \raisebox{-0.5\height}{\includegraphics[width=0.08\textwidth]{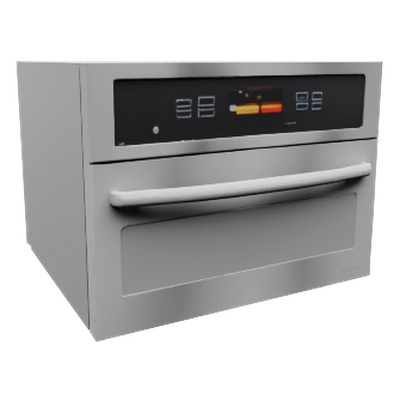}} & \raisebox{-0.5\height}{\includegraphics[width=0.08\textwidth]{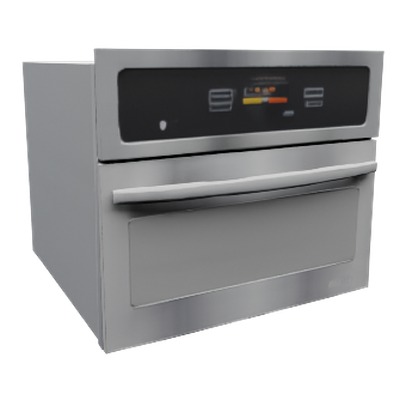}} & \raisebox{-0.5\height}{\includegraphics[width=0.08\textwidth]{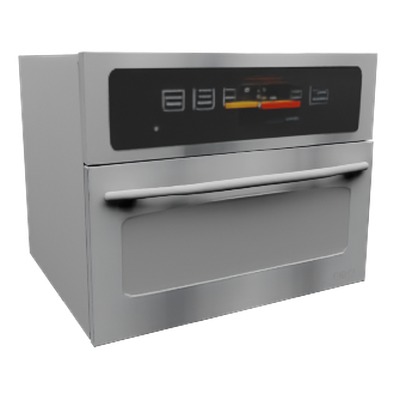}} & \raisebox{-0.5\height}{\includegraphics[width=0.08\textwidth]{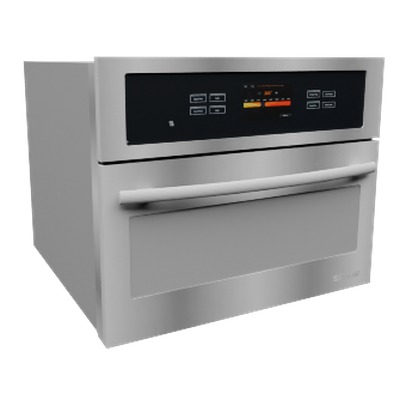}} \\[4pt]
   & \raisebox{-0.5\height}{\includegraphics[width=0.08\textwidth]{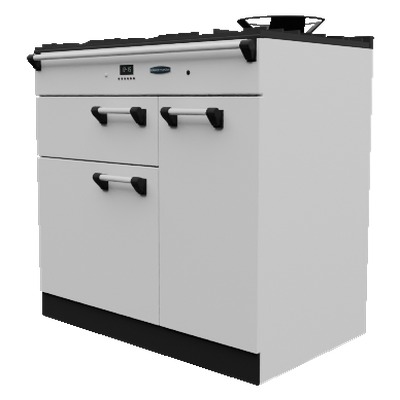}} & \raisebox{-0.5\height}{\includegraphics[width=0.08\textwidth]{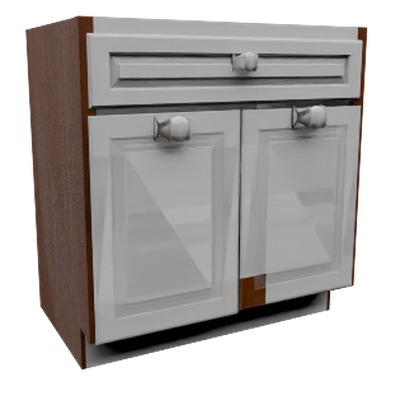}} & \raisebox{-0.5\height}{\includegraphics[width=0.08\textwidth]{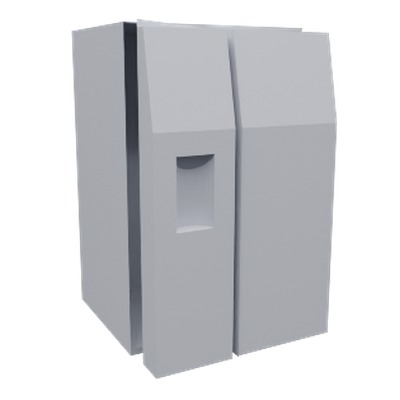}} & \raisebox{-0.5\height}{\includegraphics[width=0.08\textwidth]{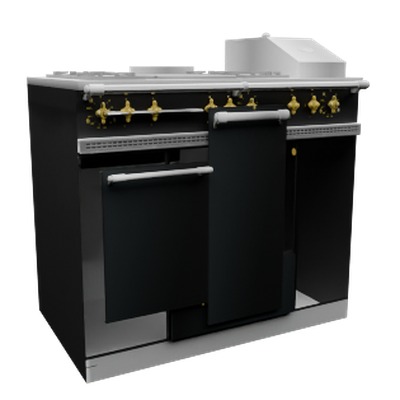}} & \raisebox{-0.5\height}{\includegraphics[width=0.08\textwidth]{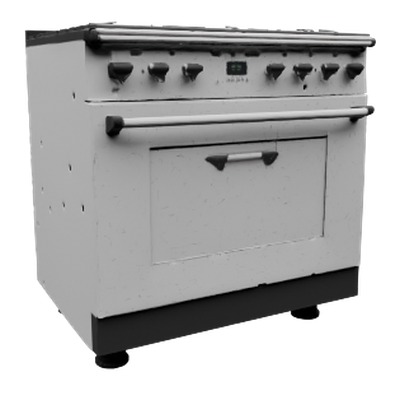}} & \raisebox{-0.5\height}{\includegraphics[width=0.08\textwidth]{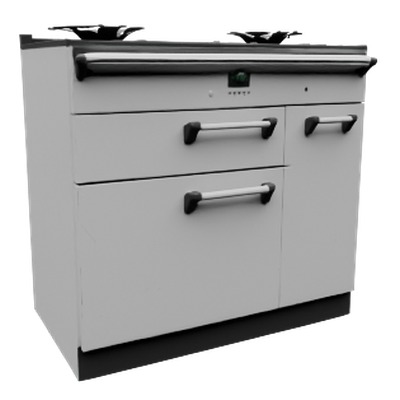}} & \raisebox{-0.5\height}{\includegraphics[width=0.08\textwidth]{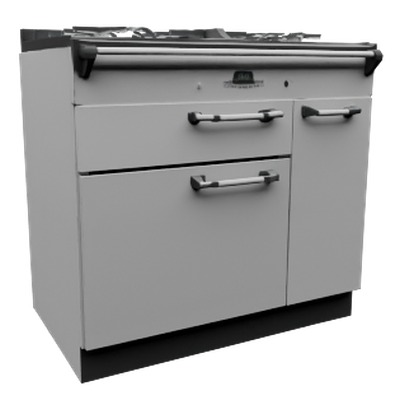}} & \raisebox{-0.5\height}{\includegraphics[width=0.08\textwidth]{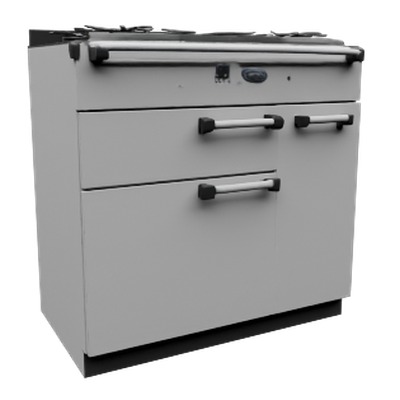}} & \raisebox{-0.5\height}{\includegraphics[width=0.08\textwidth]{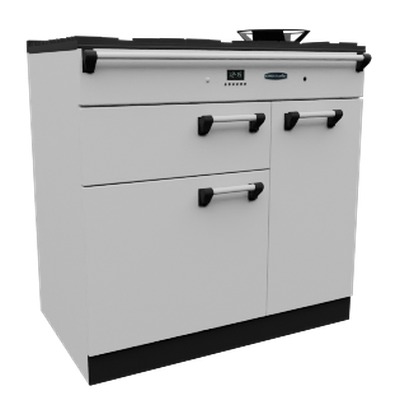}} \\
  \rotatebox[origin=c]{90}{\scriptsize ACD}\hspace{2pt} & \raisebox{-0.5\height}{\includegraphics[width=0.08\textwidth]{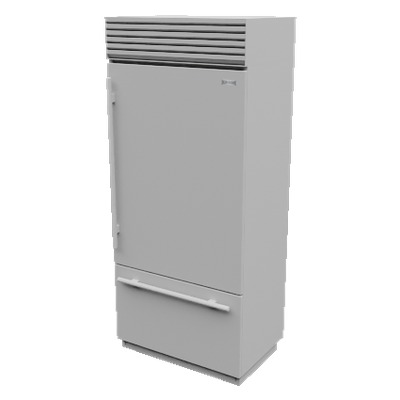}} & \raisebox{-0.5\height}{\includegraphics[width=0.08\textwidth]{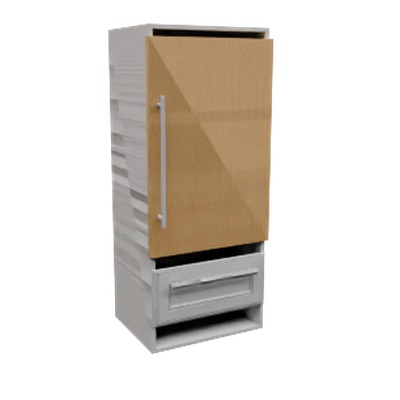}} & \raisebox{-0.5\height}{\includegraphics[width=0.08\textwidth]{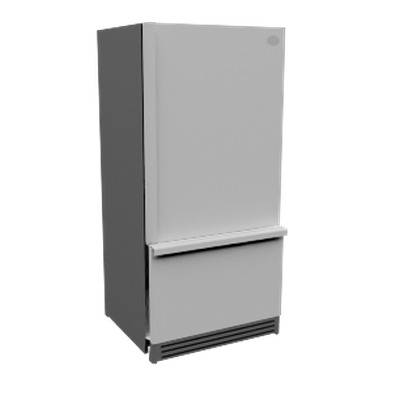}} & \raisebox{-0.5\height}{\includegraphics[width=0.08\textwidth]{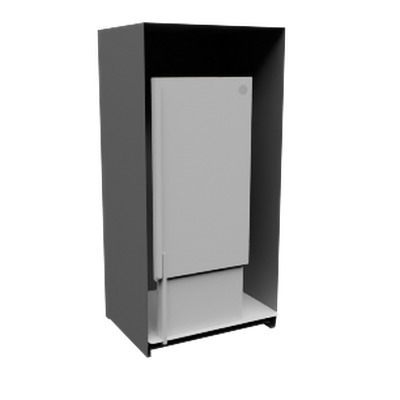}} & \raisebox{-0.5\height}{\includegraphics[width=0.08\textwidth]{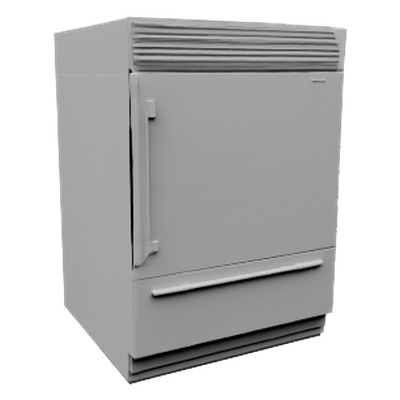}} & \raisebox{-0.5\height}{\includegraphics[width=0.08\textwidth]{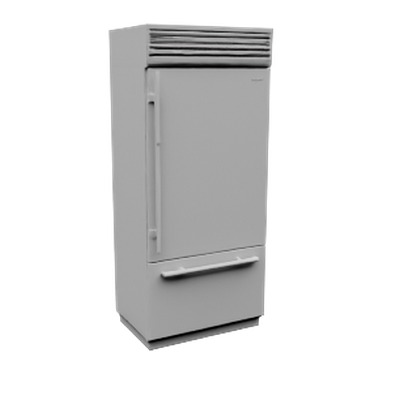}} & \raisebox{-0.5\height}{\includegraphics[width=0.08\textwidth]{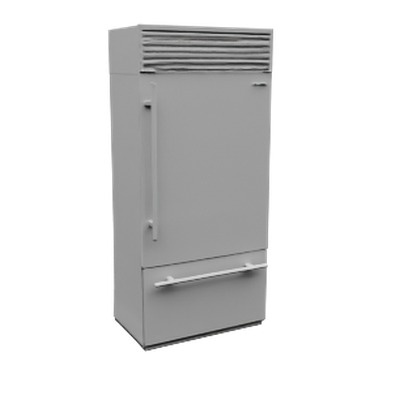}} & \raisebox{-0.5\height}{\includegraphics[width=0.08\textwidth]{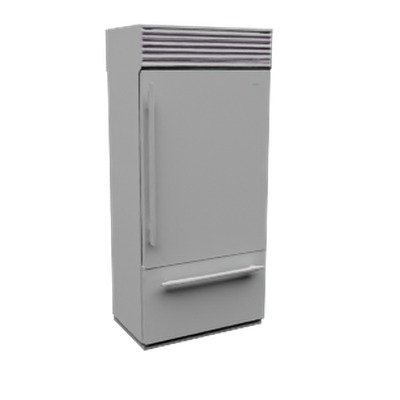}} & \raisebox{-0.5\height}{\includegraphics[width=0.08\textwidth]{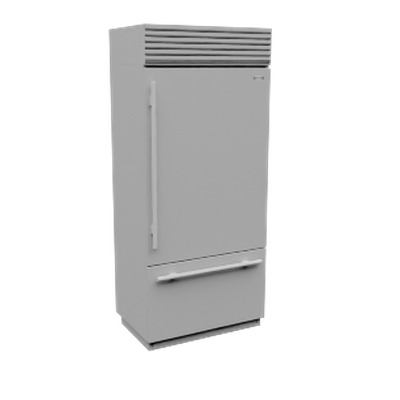}} \\
   & \raisebox{-0.5\height}{\includegraphics[width=0.08\textwidth]{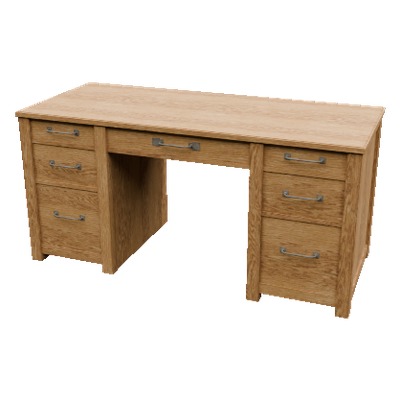}} & \raisebox{-0.5\height}{\includegraphics[width=0.08\textwidth]{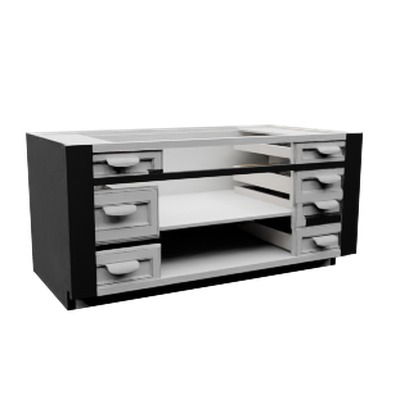}} & \raisebox{-0.5\height}{\includegraphics[width=0.08\textwidth]{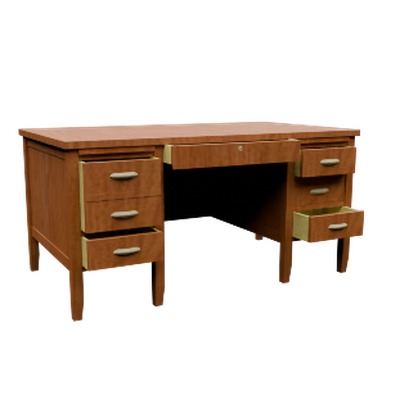}} & \raisebox{-0.5\height}{\includegraphics[width=0.08\textwidth]{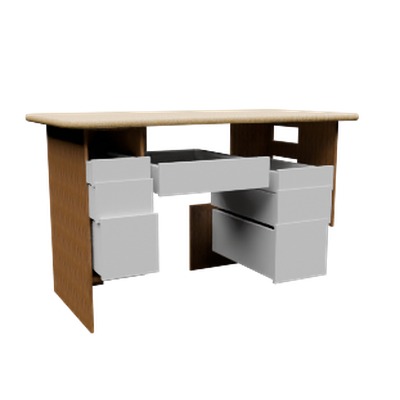}} & \raisebox{-0.5\height}{\includegraphics[width=0.08\textwidth]{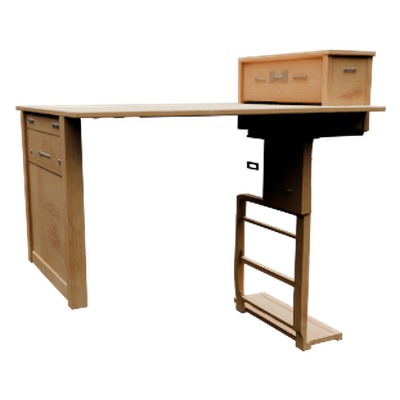}} & \raisebox{-0.5\height}{\includegraphics[width=0.08\textwidth]{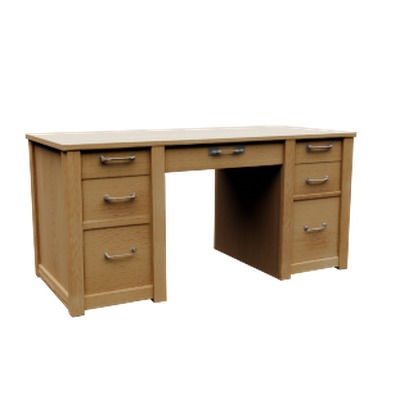}} & \raisebox{-0.5\height}{\includegraphics[width=0.08\textwidth]{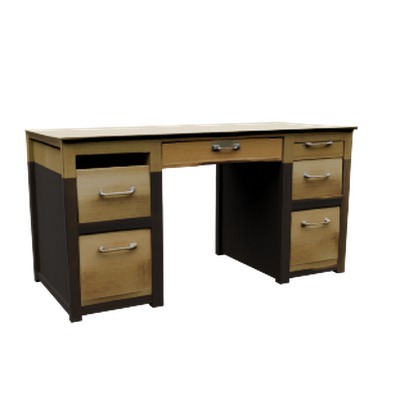}} & \raisebox{-0.5\height}{\includegraphics[width=0.08\textwidth]{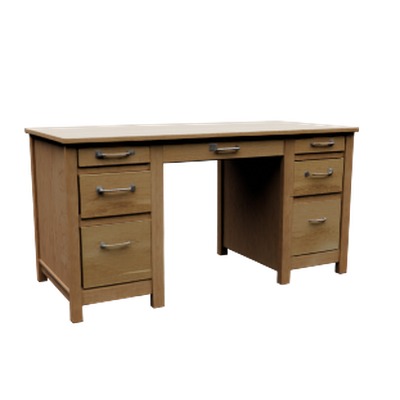}} & \raisebox{-0.5\height}{\includegraphics[width=0.08\textwidth]{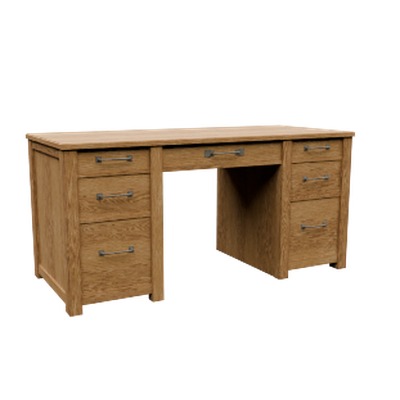}} \\
\end{tabular}

    \caption{Rest-state generation on three more inputs of PM (top) and ACD (bottom), one per object. All results are shown at rest from the same view as the ground truth.}
    \label{fig:supp_rest}
\end{figure*}

\begin{figure*}[tp]
    \centering
    \setlength{\tabcolsep}{0.3pt}
\renewcommand{\arraystretch}{0.5}
\begin{tabular}{cccccccc}
  \scriptsize\renewcommand{\arraystretch}{0.9}\begin{tabular}[b]{@{}c@{}}Rest\end{tabular} & \scriptsize\renewcommand{\arraystretch}{0.9}\begin{tabular}[b]{@{}c@{}}Close-up\end{tabular} & \scriptsize\renewcommand{\arraystretch}{0.9}\begin{tabular}[b]{@{}c@{}}DIPO\end{tabular} & \scriptsize\renewcommand{\arraystretch}{0.9}\begin{tabular}[b]{@{}c@{}}FreeArt3D$^\dagger$\end{tabular} & \scriptsize\renewcommand{\arraystretch}{0.9}\begin{tabular}[b]{@{}c@{}}VoxHammer$^\ddagger$\end{tabular} & \scriptsize\renewcommand{\arraystretch}{0.9}\begin{tabular}[b]{@{}c@{}}$\mathcal{A}$ (before)\end{tabular} & \scriptsize\renewcommand{\arraystretch}{0.9}\begin{tabular}[b]{@{}c@{}}Ours\end{tabular} & \scriptsize\renewcommand{\arraystretch}{0.9}\begin{tabular}[b]{@{}c@{}}Ground truth\end{tabular} \\[2pt]
  \includegraphics[width=0.12\textwidth]{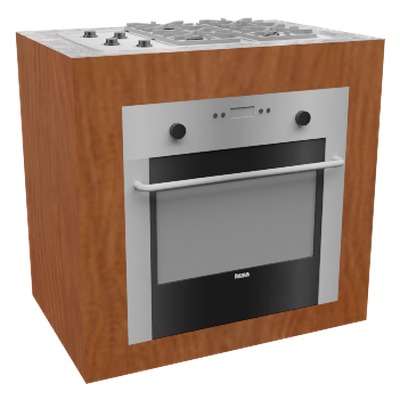} & \includegraphics[width=0.12\textwidth]{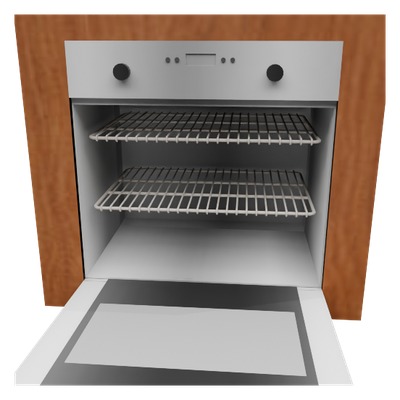} & \includegraphics[width=0.12\textwidth]{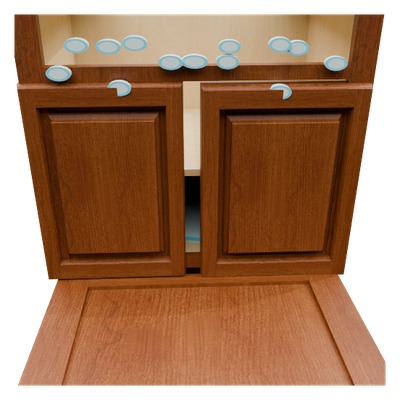} & \includegraphics[width=0.12\textwidth]{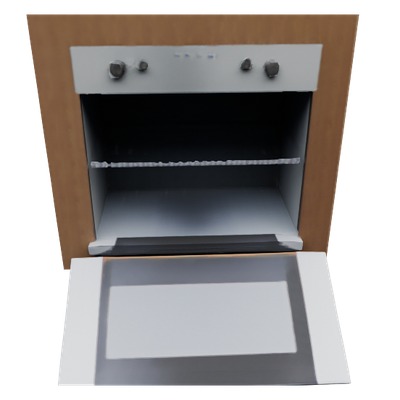} & \includegraphics[width=0.12\textwidth]{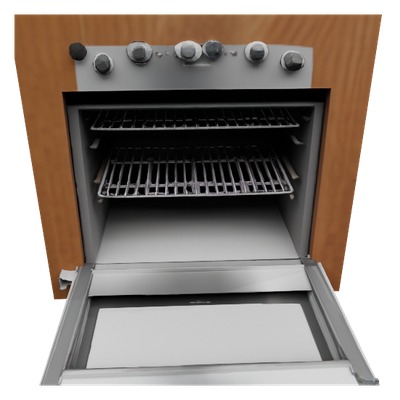} & \includegraphics[width=0.12\textwidth]{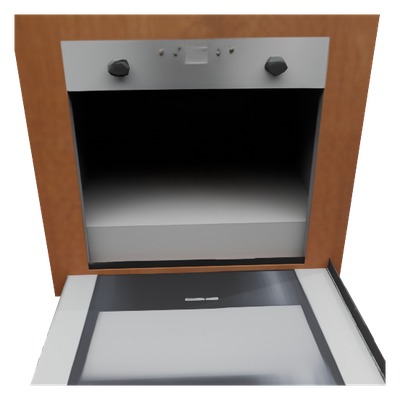} & \includegraphics[width=0.12\textwidth]{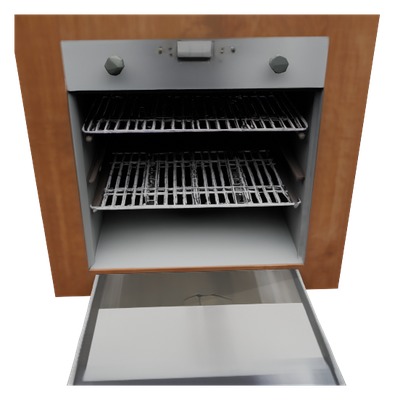} & \includegraphics[width=0.12\textwidth]{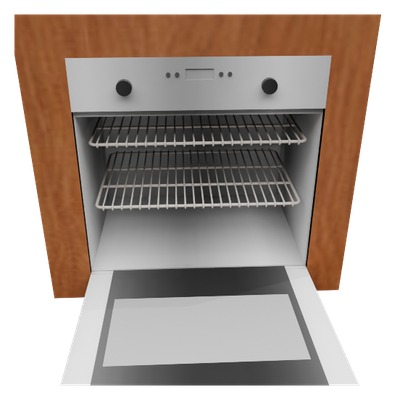} \\
  \includegraphics[width=0.12\textwidth]{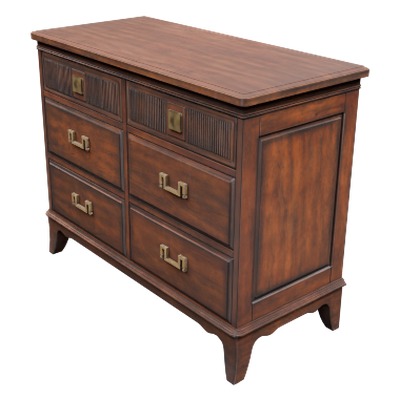} & \includegraphics[width=0.12\textwidth]{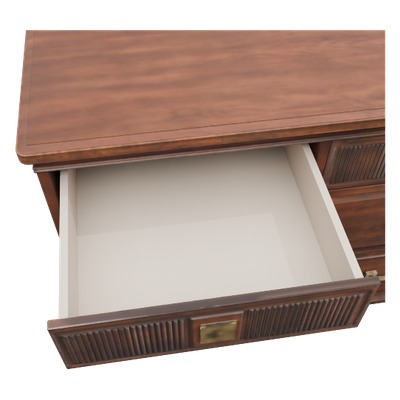} & \includegraphics[width=0.12\textwidth]{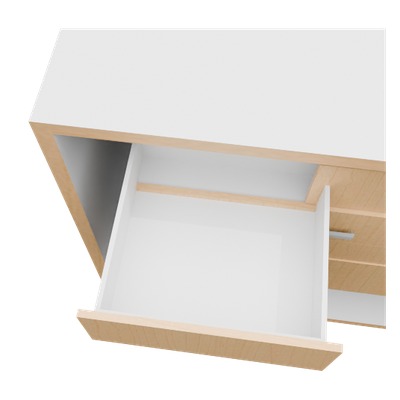} & \includegraphics[width=0.12\textwidth]{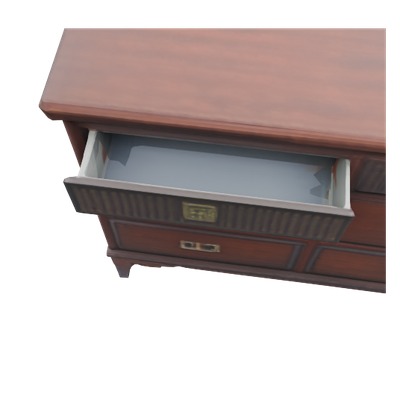} & \includegraphics[width=0.12\textwidth]{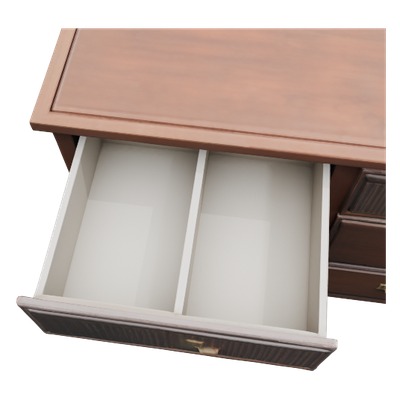} & \includegraphics[width=0.12\textwidth]{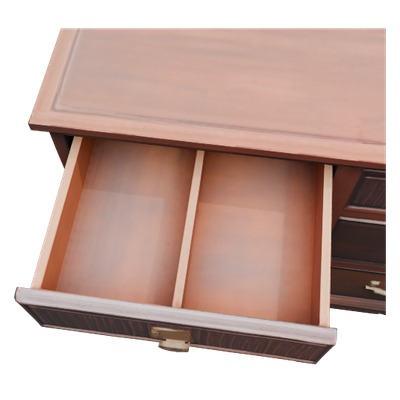} & \includegraphics[width=0.12\textwidth]{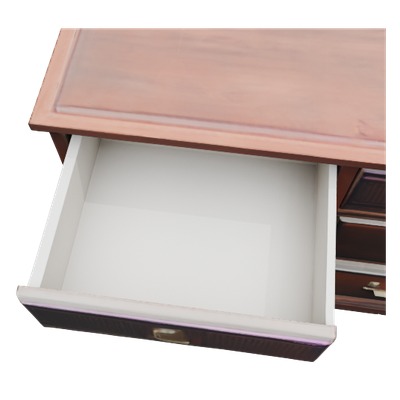} & \includegraphics[width=0.12\textwidth]{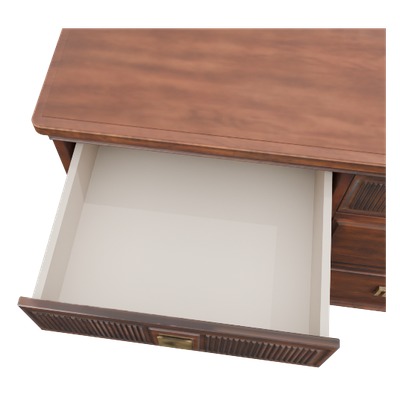} \\
  \includegraphics[width=0.12\textwidth]{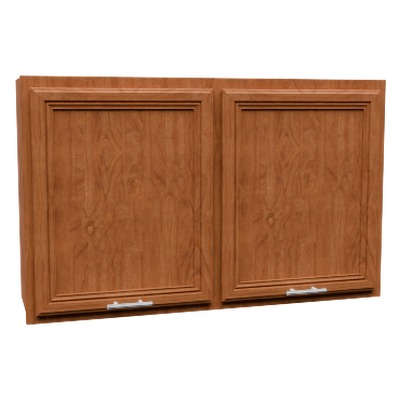} & \includegraphics[width=0.12\textwidth]{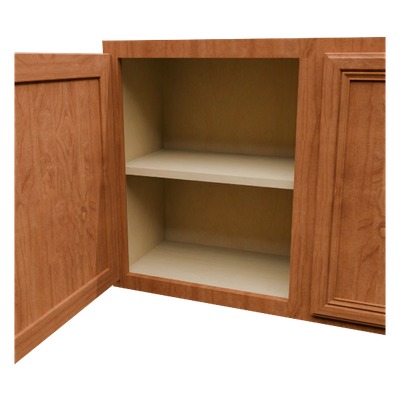} & \includegraphics[width=0.12\textwidth]{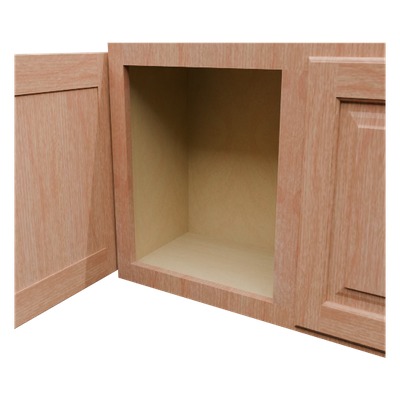} & \includegraphics[width=0.12\textwidth]{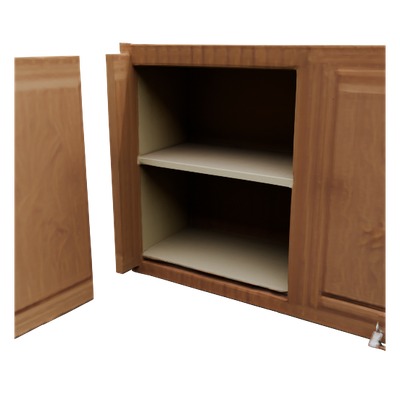} & \includegraphics[width=0.12\textwidth]{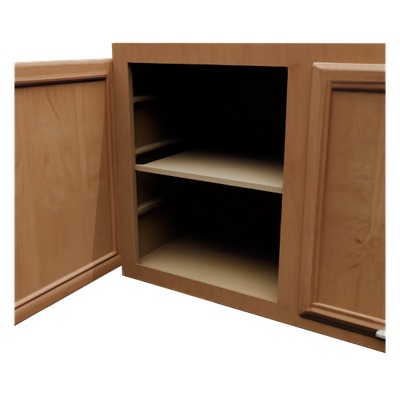} & \includegraphics[width=0.12\textwidth]{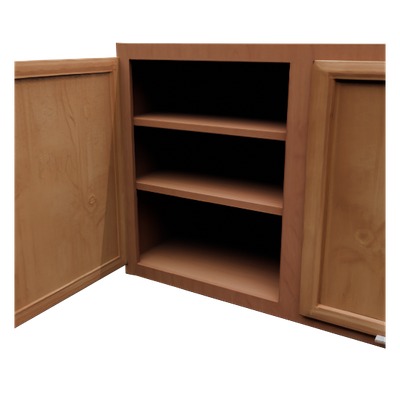} & \includegraphics[width=0.12\textwidth]{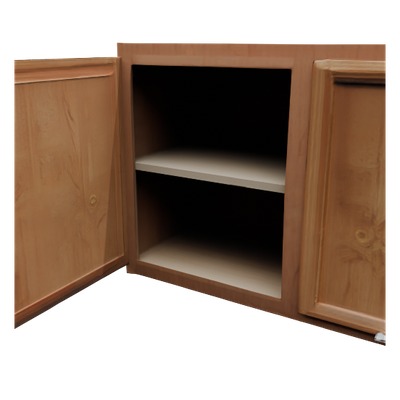} & \includegraphics[width=0.12\textwidth]{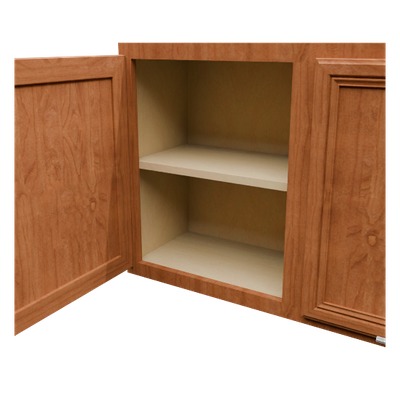} \\
\end{tabular}

    \caption{Three more updates on RevealBench-ArtiVerse. All results are posed with the ground-truth joint at the true opening and rendered from a novel view. $^\dagger$ privileged inputs, $^\ddagger$ opened part given.}
    \label{fig:supp_updates}
\end{figure*}

\begin{figure*}[tp]
    \centering
    \setlength{\tabcolsep}{0.3pt}
\renewcommand{\arraystretch}{0.5}
\begin{tabular}{ccccc}
  \scriptsize\renewcommand{\arraystretch}{0.9}\begin{tabular}[b]{@{}c@{}}Rest\end{tabular} & \scriptsize\renewcommand{\arraystretch}{0.9}\begin{tabular}[b]{@{}c@{}}Close-up\end{tabular} & \scriptsize\renewcommand{\arraystretch}{0.9}\begin{tabular}[b]{@{}c@{}}$\mathcal{A}$ (before)\end{tabular} & \scriptsize\renewcommand{\arraystretch}{0.9}\begin{tabular}[b]{@{}c@{}}Ours\end{tabular} & \scriptsize\renewcommand{\arraystretch}{0.9}\begin{tabular}[b]{@{}c@{}}Ground truth\end{tabular} \\[2pt]
  \includegraphics[width=0.13\textwidth]{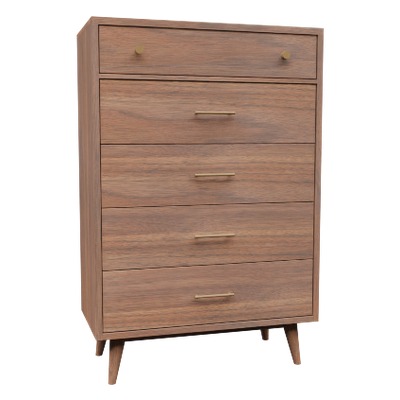} & \includegraphics[width=0.13\textwidth]{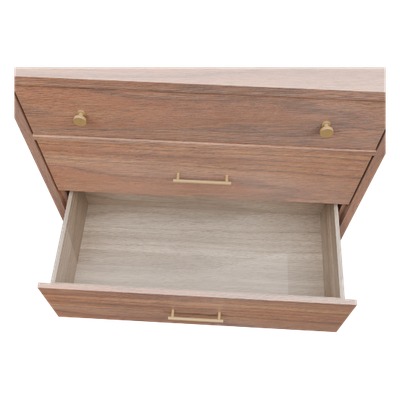} & \includegraphics[width=0.13\textwidth]{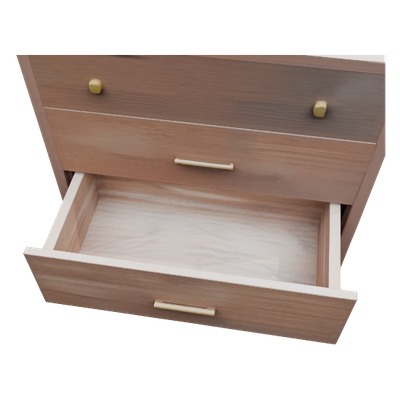} & \includegraphics[width=0.13\textwidth]{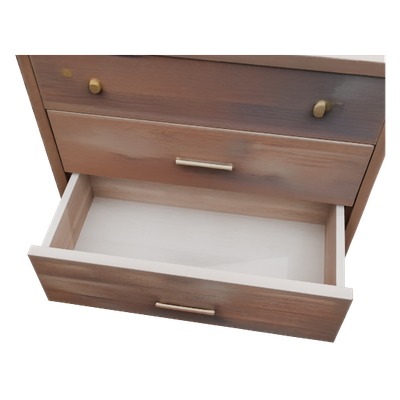} & \includegraphics[width=0.13\textwidth]{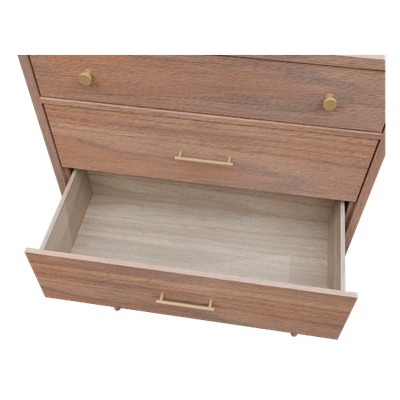} \\
  \hspace{0.13\textwidth} & \includegraphics[width=0.13\textwidth]{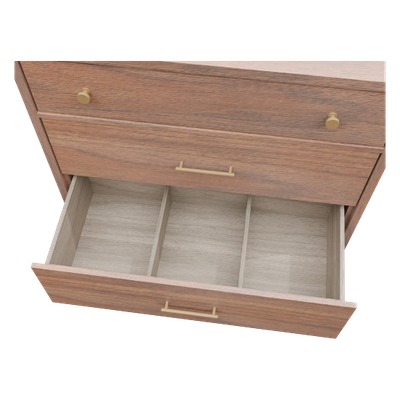} & \includegraphics[width=0.13\textwidth]{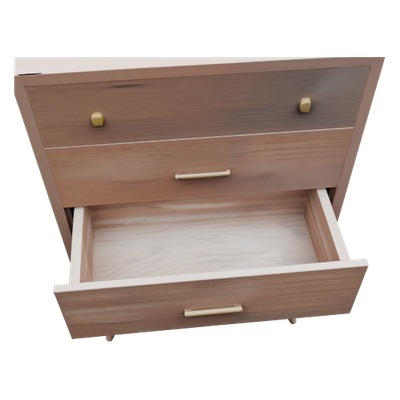} & \includegraphics[width=0.13\textwidth]{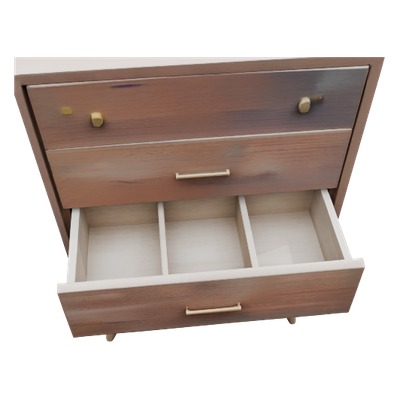} & \includegraphics[width=0.13\textwidth]{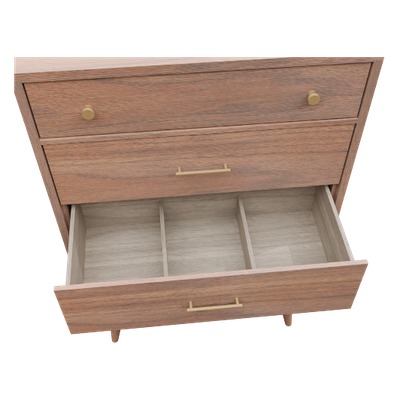} \\
  \hspace{0.13\textwidth} & \includegraphics[width=0.13\textwidth]{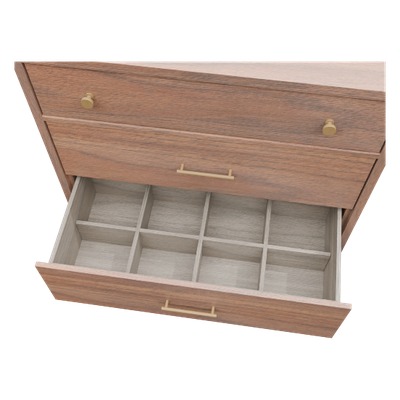} & \includegraphics[width=0.13\textwidth]{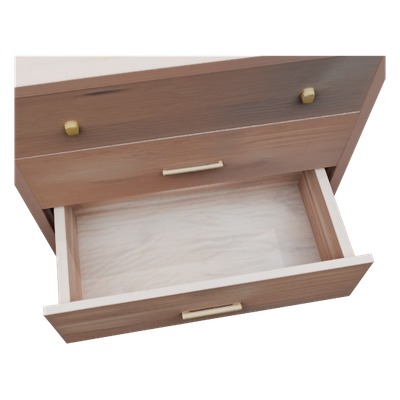} & \includegraphics[width=0.13\textwidth]{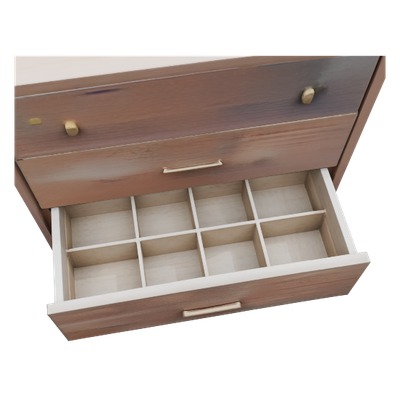} & \includegraphics[width=0.13\textwidth]{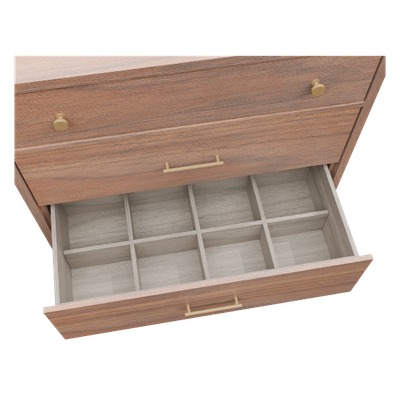} \\
  \hspace{0.13\textwidth} & \includegraphics[width=0.13\textwidth]{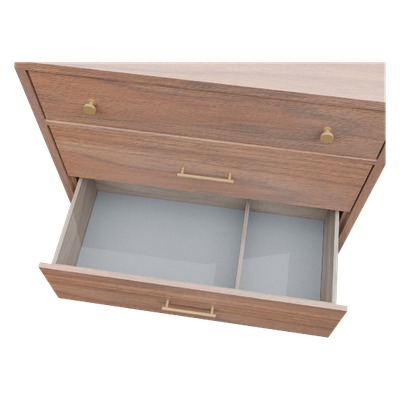} & \includegraphics[width=0.13\textwidth]{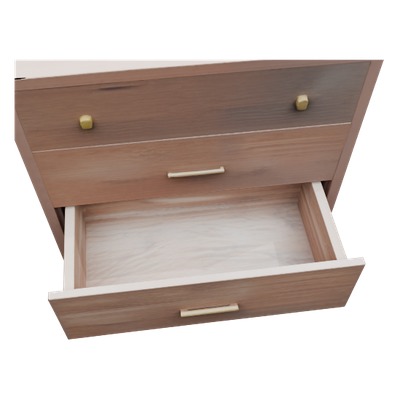} & \includegraphics[width=0.13\textwidth]{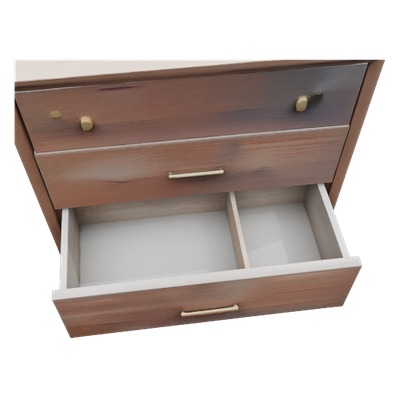} & \includegraphics[width=0.13\textwidth]{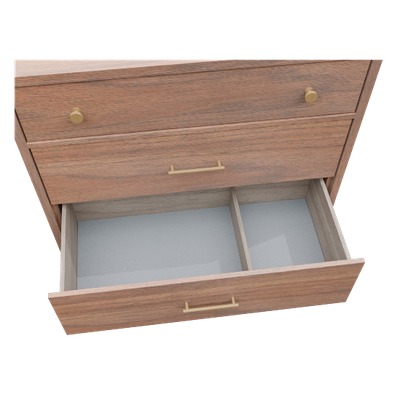} \\[4pt]
  \includegraphics[width=0.13\textwidth]{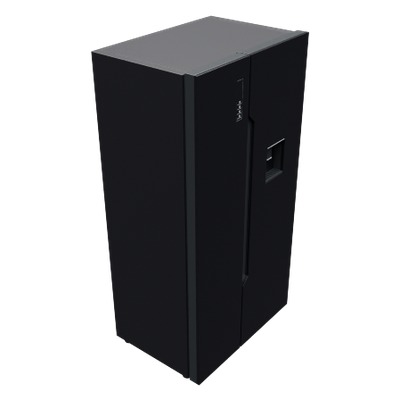} & \includegraphics[width=0.13\textwidth]{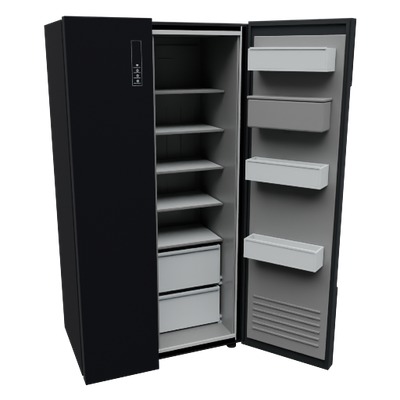} & \includegraphics[width=0.13\textwidth]{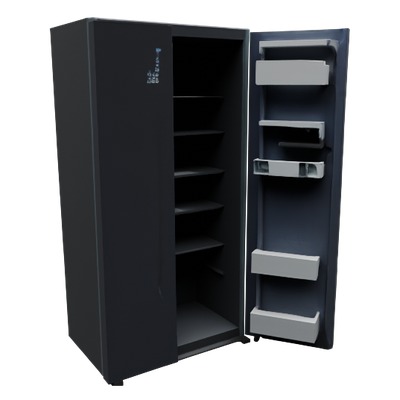} & \includegraphics[width=0.13\textwidth]{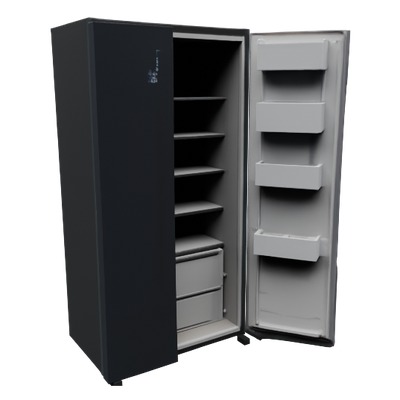} & \includegraphics[width=0.13\textwidth]{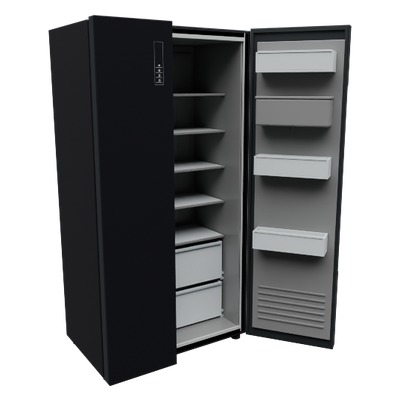} \\
  \hspace{0.13\textwidth} & \includegraphics[width=0.13\textwidth]{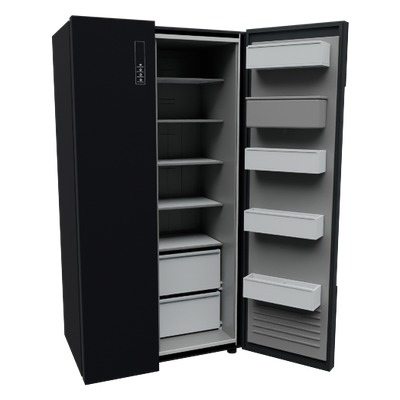} & \includegraphics[width=0.13\textwidth]{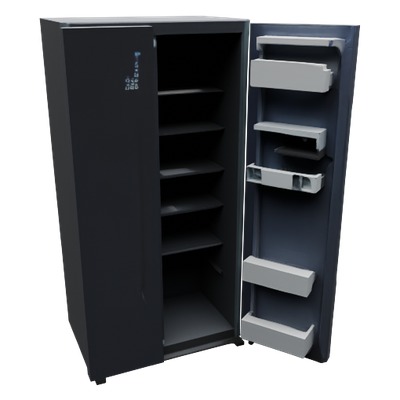} & \includegraphics[width=0.13\textwidth]{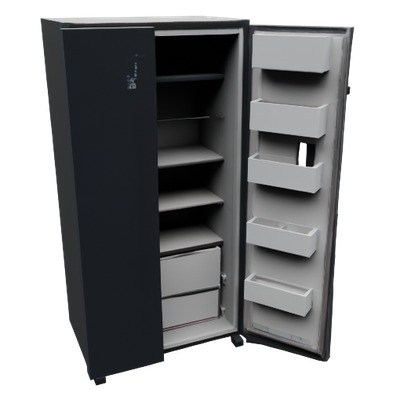} & \includegraphics[width=0.13\textwidth]{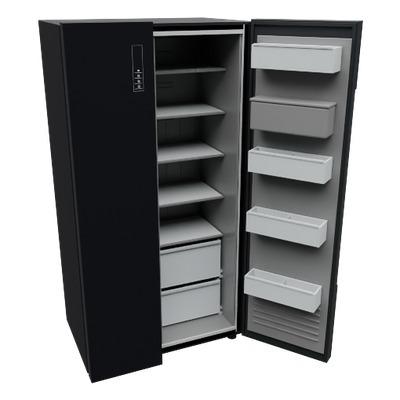} \\
  \hspace{0.13\textwidth} & \includegraphics[width=0.13\textwidth]{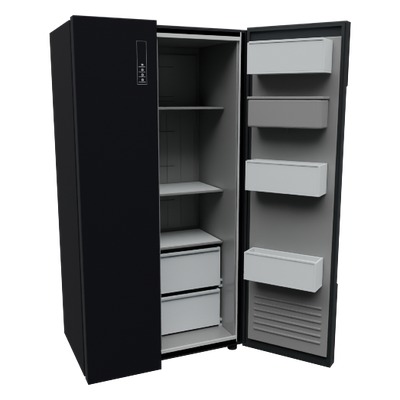} & \includegraphics[width=0.13\textwidth]{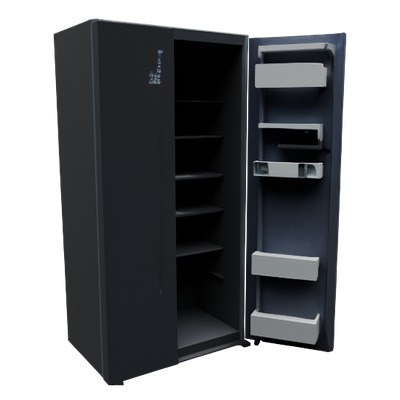} & \includegraphics[width=0.13\textwidth]{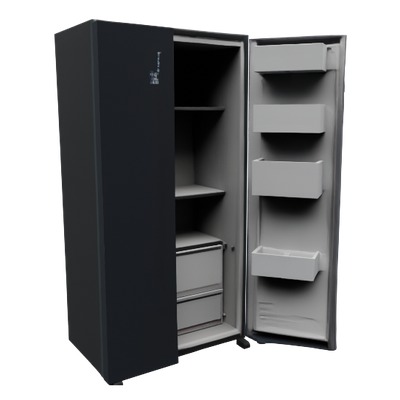} & \includegraphics[width=0.13\textwidth]{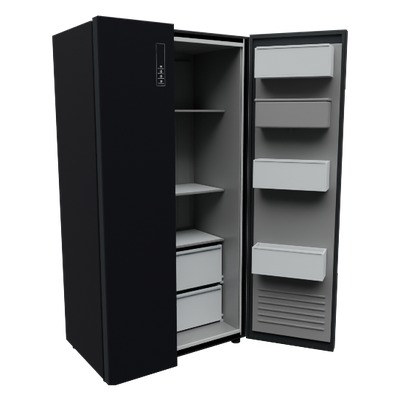} \\
  \hspace{0.13\textwidth} & \includegraphics[width=0.13\textwidth]{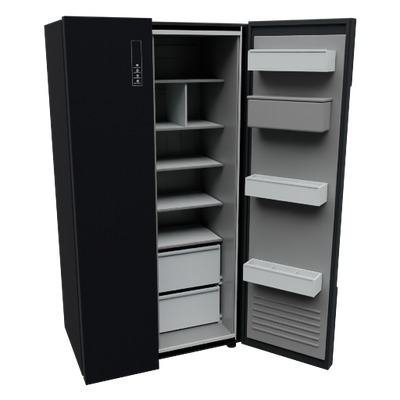} & \includegraphics[width=0.13\textwidth]{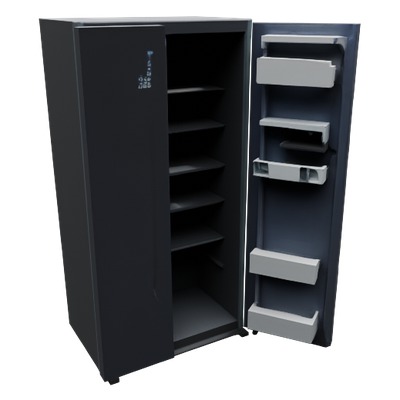} & \includegraphics[width=0.13\textwidth]{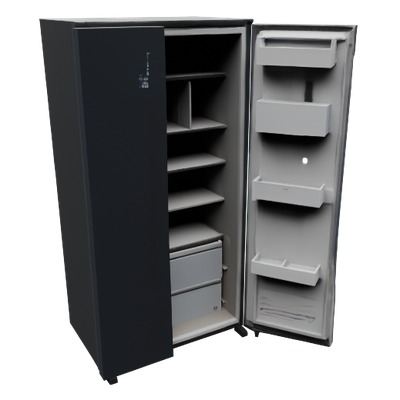} & \includegraphics[width=0.13\textwidth]{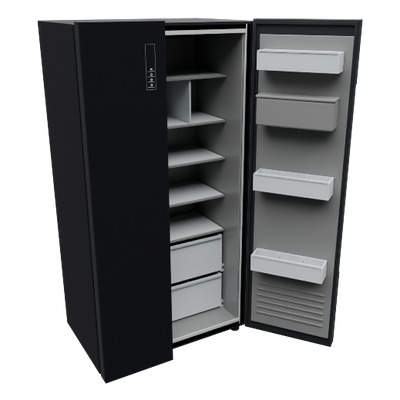} \\
\end{tabular}

    \caption{All four variants of two RevealBench-Astra families, a chest of drawers whose variants differ in the partitions of the opened drawer (top) and a refrigerator whose variants differ in the shelves and door racks (bottom). The rest image is the same for the four variants of a family, so the initial asset $\mathcal{A}$ cannot tell them apart, while the update follows each close-up (revealed F@2\% of 99.0 to 100 and 97.7 to 99.6).}
    \label{fig:supp_siblings}
\end{figure*}

\begin{figure*}[p]
    \centering
    \setlength{\tabcolsep}{0.3pt}
\renewcommand{\arraystretch}{0.5}
\begin{tabular}{ccccc}
  \scriptsize\renewcommand{\arraystretch}{0.9}\begin{tabular}[b]{@{}c@{}}Rest\end{tabular} & \scriptsize\renewcommand{\arraystretch}{0.9}\begin{tabular}[b]{@{}c@{}}Close-up\end{tabular} & \scriptsize\renewcommand{\arraystretch}{0.9}\begin{tabular}[b]{@{}c@{}}$\mathcal{A}$ (before),\\own joint\end{tabular} & \scriptsize\renewcommand{\arraystretch}{0.9}\begin{tabular}[b]{@{}c@{}}Ours,\\own joint\end{tabular} & \scriptsize\renewcommand{\arraystretch}{0.9}\begin{tabular}[b]{@{}c@{}}Ground truth\end{tabular} \\[2pt]
  \includegraphics[width=0.17\textwidth]{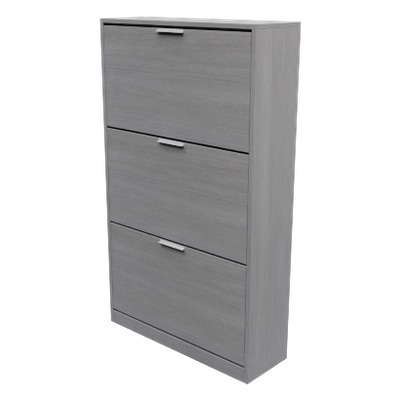} & \includegraphics[width=0.17\textwidth]{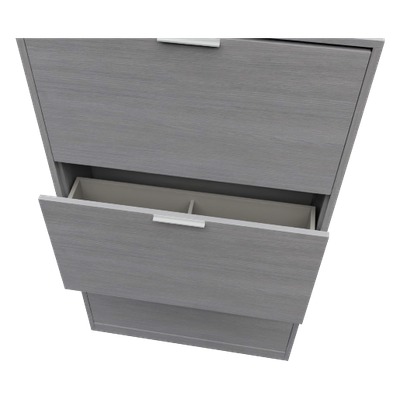} & \includegraphics[width=0.17\textwidth]{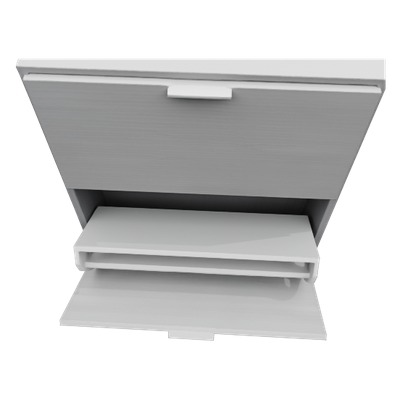} & \includegraphics[width=0.17\textwidth]{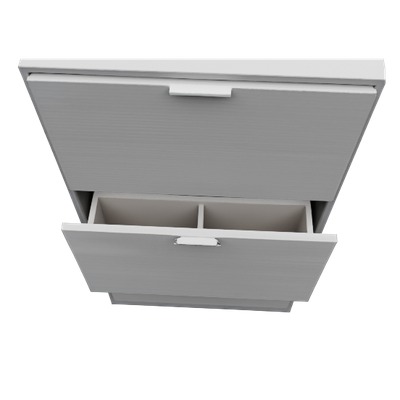} & \includegraphics[width=0.17\textwidth]{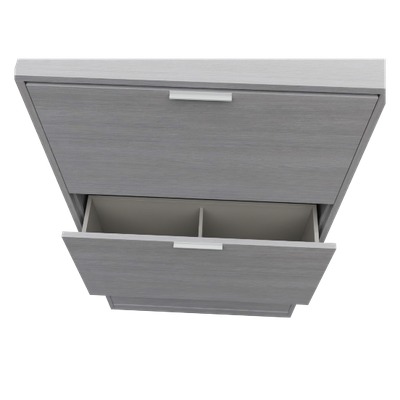} \\
  \includegraphics[width=0.17\textwidth]{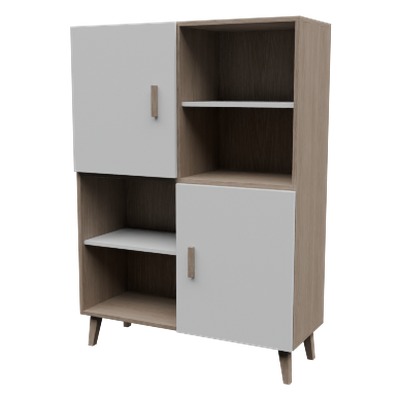} & \includegraphics[width=0.17\textwidth]{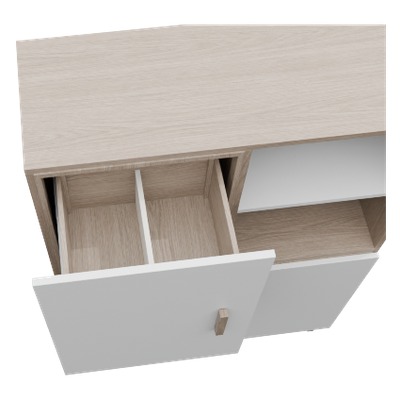} & \includegraphics[width=0.17\textwidth]{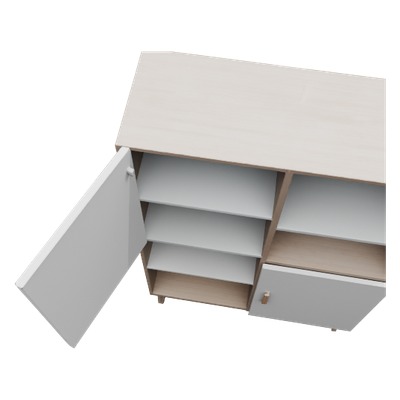} & \includegraphics[width=0.17\textwidth]{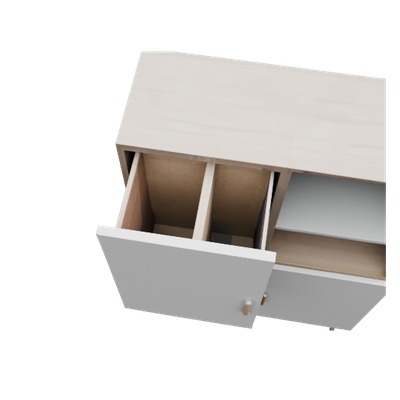} & \includegraphics[width=0.17\textwidth]{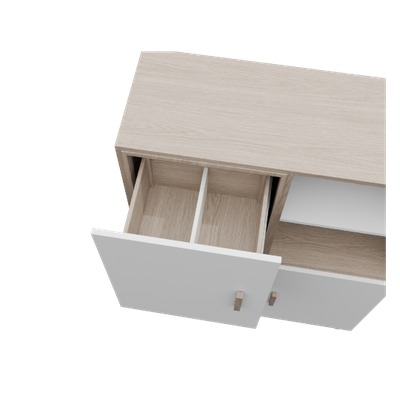} \\
  \includegraphics[width=0.17\textwidth]{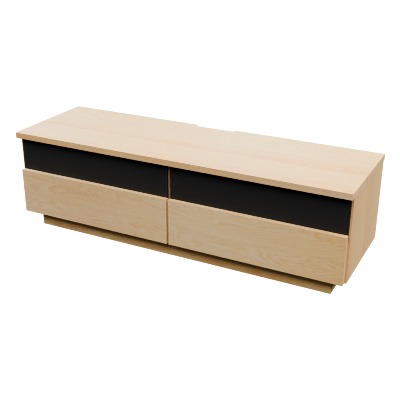} & \includegraphics[width=0.17\textwidth]{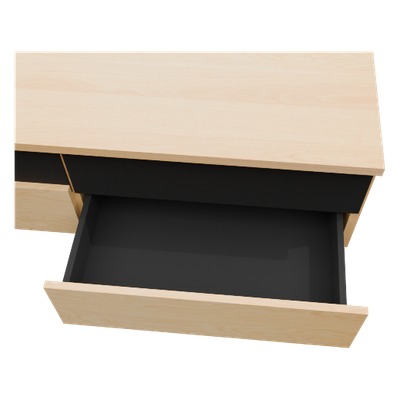} & \includegraphics[width=0.17\textwidth]{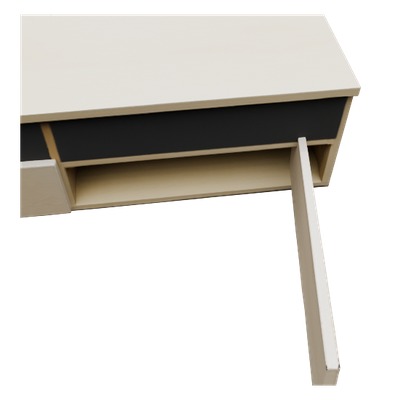} & \includegraphics[width=0.17\textwidth]{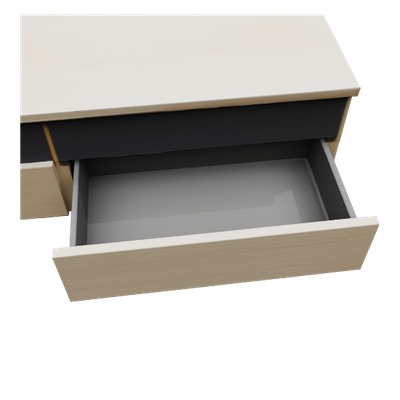} & \includegraphics[width=0.17\textwidth]{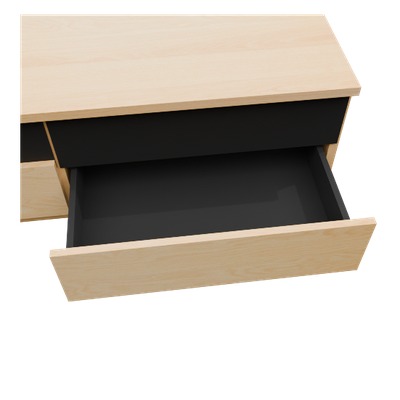} \\
  \includegraphics[width=0.17\textwidth]{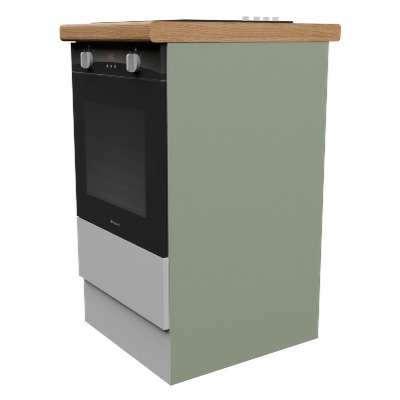} & \includegraphics[width=0.17\textwidth]{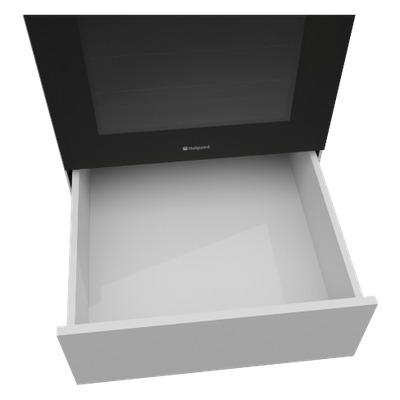} & \includegraphics[width=0.17\textwidth]{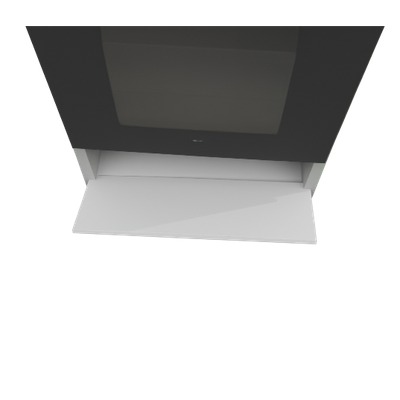} & \includegraphics[width=0.17\textwidth]{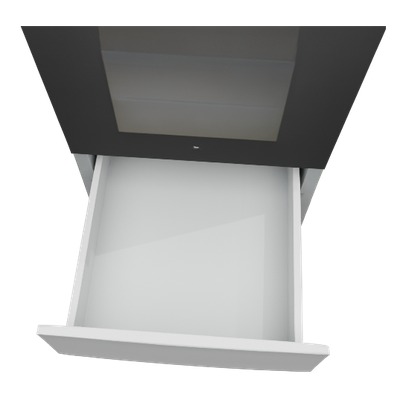} & \includegraphics[width=0.17\textwidth]{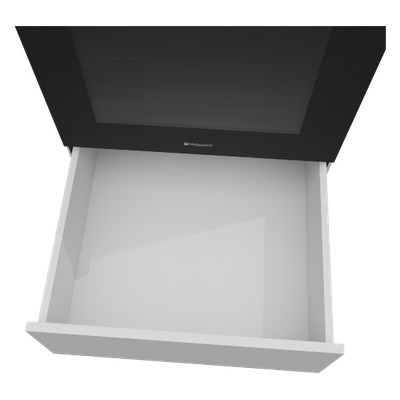} \\
  \includegraphics[width=0.17\textwidth]{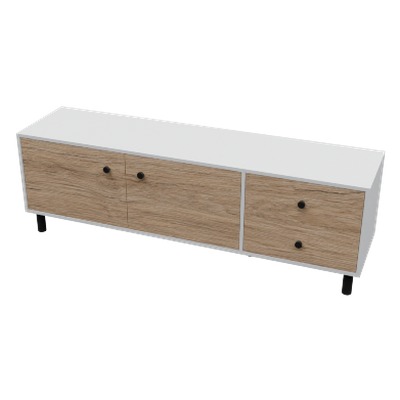} & \includegraphics[width=0.17\textwidth]{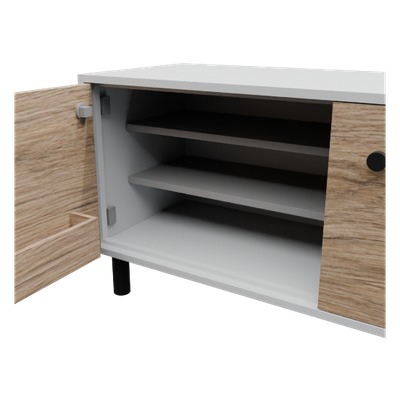} & \includegraphics[width=0.17\textwidth]{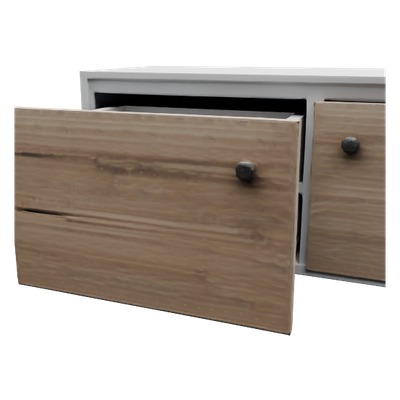} & \includegraphics[width=0.17\textwidth]{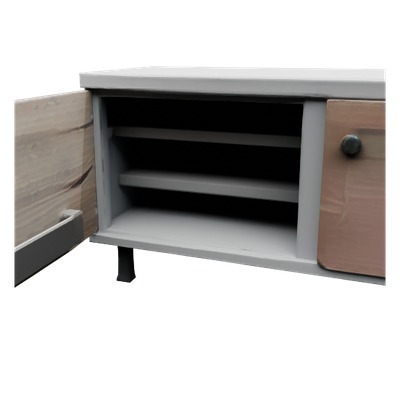} & \includegraphics[width=0.17\textwidth]{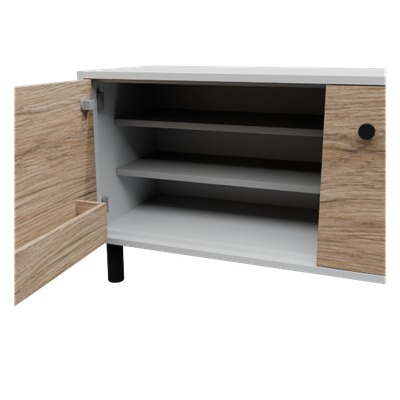} \\
\end{tabular}

    \caption{Joint correction. Before the update ($\mathcal{A}$) and after it, the opened part is posed at the true opening with the joint that the method predicts. In the first four rows, $\mathcal{A}$ predicts a hinge for a drawer, with axis errors of 87.9 to 89.8$^\circ$, and the update predicts a sliding joint with an axis error of at most 0.3$^\circ$. In the last row, $\mathcal{A}$ predicts a sliding joint for a door (88.5$^\circ$), and the update predicts its hinge (0.3$^\circ$).}
    \label{fig:supp_joint}
\end{figure*}

\begin{figure*}[p]
    \centering
    \setlength{\tabcolsep}{0.3pt}
\renewcommand{\arraystretch}{0.5}
\begin{tabular}{ccccc}
  \scriptsize\renewcommand{\arraystretch}{0.9}\begin{tabular}[b]{@{}c@{}}Rest\end{tabular} & \scriptsize\renewcommand{\arraystretch}{0.9}\begin{tabular}[b]{@{}c@{}}Close-up\end{tabular} & \scriptsize\renewcommand{\arraystretch}{0.9}\begin{tabular}[b]{@{}c@{}}$\mathcal{A}$ (before),\\own joint\end{tabular} & \scriptsize\renewcommand{\arraystretch}{0.9}\begin{tabular}[b]{@{}c@{}}Ours,\\own joint and state\end{tabular} & \scriptsize\renewcommand{\arraystretch}{0.9}\begin{tabular}[b]{@{}c@{}}Ground truth\end{tabular} \\[2pt]
  \includegraphics[width=0.17\textwidth]{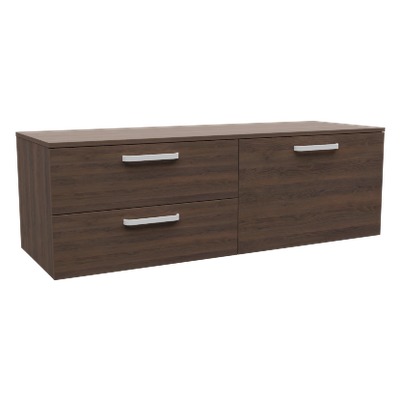} & \includegraphics[width=0.17\textwidth]{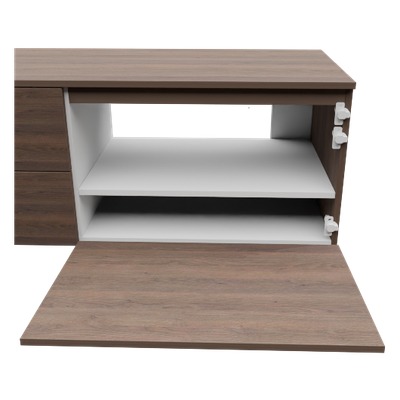} & \includegraphics[width=0.17\textwidth]{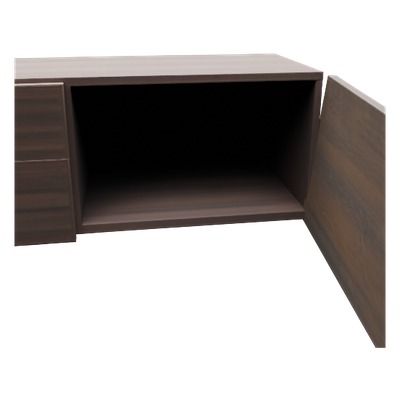} & \includegraphics[width=0.17\textwidth]{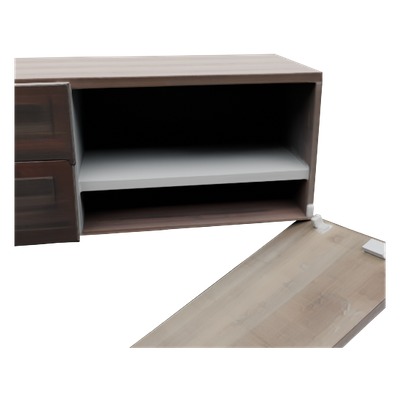} & \includegraphics[width=0.17\textwidth]{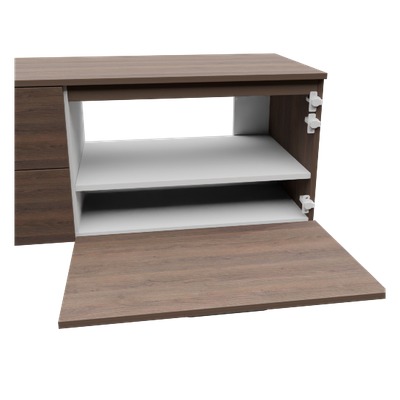} \\
  \includegraphics[width=0.17\textwidth]{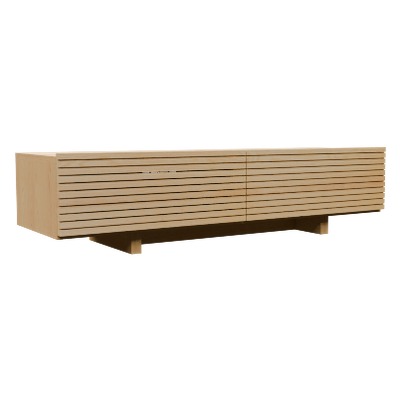} & \includegraphics[width=0.17\textwidth]{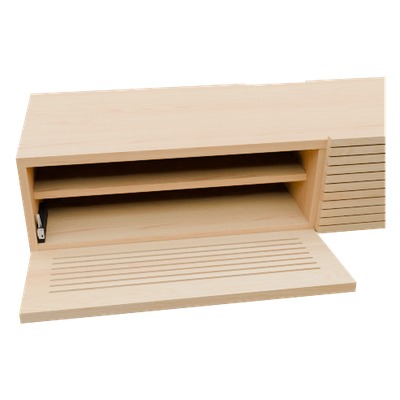} & \includegraphics[width=0.17\textwidth]{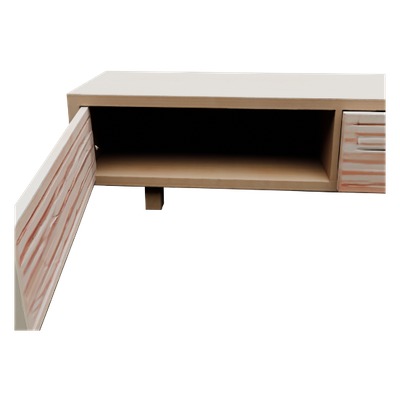} & \includegraphics[width=0.17\textwidth]{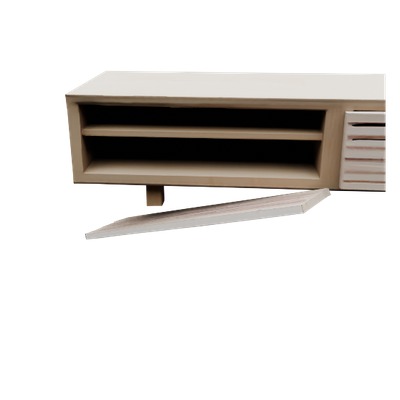} & \includegraphics[width=0.17\textwidth]{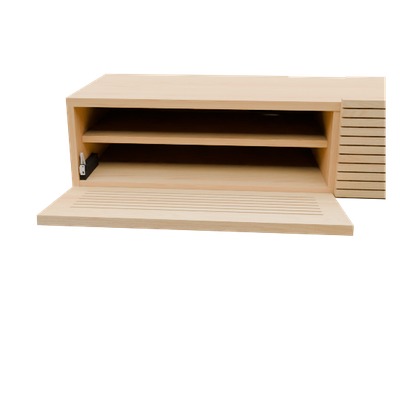} \\
  \includegraphics[width=0.17\textwidth]{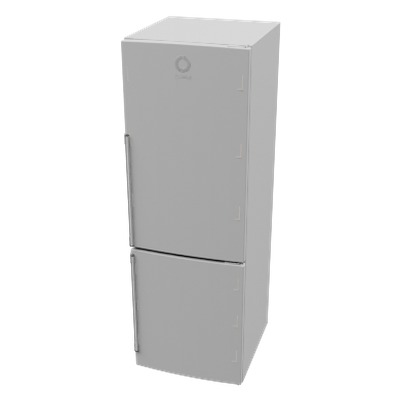} & \includegraphics[width=0.17\textwidth]{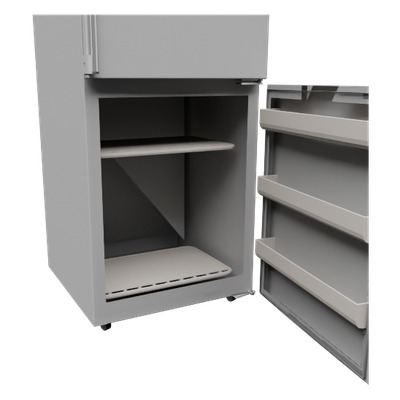} & \includegraphics[width=0.17\textwidth]{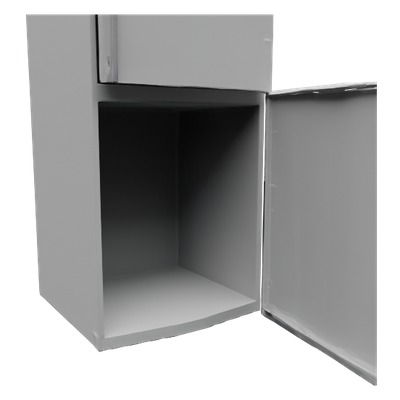} & \includegraphics[width=0.17\textwidth]{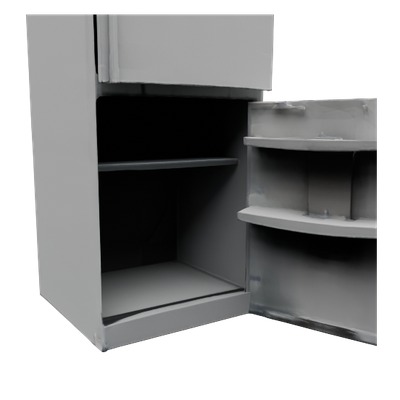} & \includegraphics[width=0.17\textwidth]{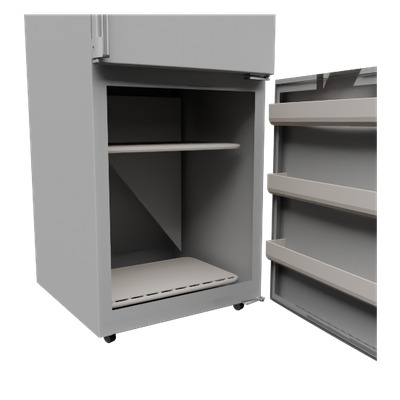} \\
  \includegraphics[width=0.17\textwidth]{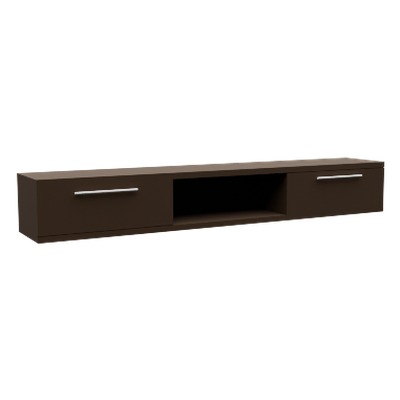} & \includegraphics[width=0.17\textwidth]{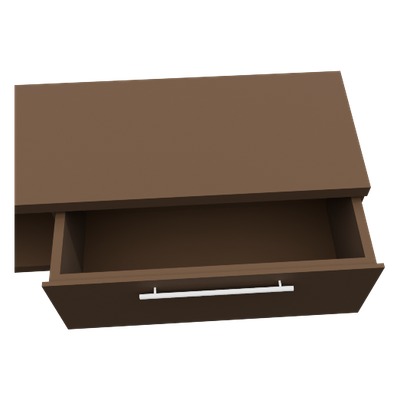} & \includegraphics[width=0.17\textwidth]{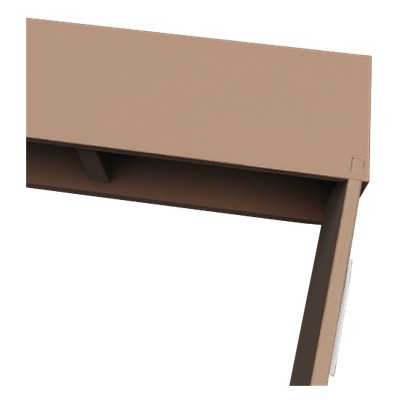} & \includegraphics[width=0.17\textwidth]{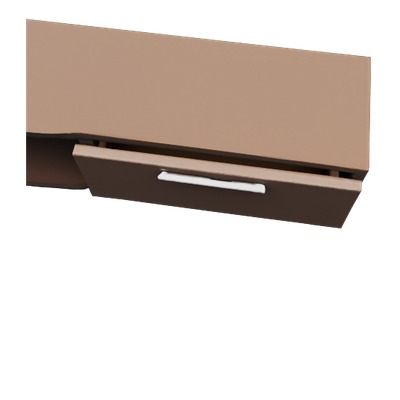} & \includegraphics[width=0.17\textwidth]{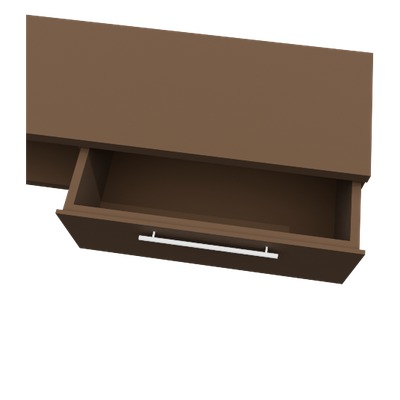} \\
\end{tabular}

    \caption{Failure cases. $\mathcal{A}$ is shown at the true opening. In the first two rows, $\mathcal{A}$ hinges a drop-down flap at its side (axis errors of 85.4 and 87.3\textdegree{}), and the update moves the hinge to the bottom edge but leaves it tilted by 22\textdegree{}, although the revealed geometry is close to the ground truth (revealed F@2\% of 99.2 and 99.5). Training on more varied hinges, such as these flaps, would likely help the open branch correct them fully. In the third row, the opening of the fridge door is estimated 29\textdegree{} off, and the bins seen on the inside of the door come out bent and incomplete, with a revealed F@2\% of 28.6 (18.6 for $\mathcal{A}$). In the fourth row, the update corrects the joint type of the drawer but predicts an opening of about 14\% of the true one, an error of 82\% of the travel, although the revealed geometry is correct (97.3).}
    \label{fig:supp_failures}
\end{figure*}

\cref{fig:supp_rest} compares the rest-state generation on PM and ACD.
\cref{fig:supp_updates} shows more updates on RevealBench-ArtiVerse, and \cref{fig:supp_siblings} all four variants of two RevealBench-Astra families.
\cref{fig:supp_joint} shows joints that the update corrects, and \cref{fig:supp_failures} failure cases.

\section{Limitations and Future Work}
\label{sec:limitation}
\methodname{} relies on known camera poses and on registration to a generated $\mathcal{A}$ by rendering its textured mesh (\cref{sec:align} and \cref{sec:supp_inference}). 
Partial occlusion, such as by the robot arm in \cref{fig:supp_robot}, is tolerated, but interiors that hold several objects, such as a shelf full of books or items stored in a cabinet, are not evaluated, and the update may degrade on them.
\methodname{} also estimates prismatic states less accurately than FoundationPose, although FoundationPose requires the ground-truth mesh, depth, and rest pose (\cref{tab:main_rb_state}).
Both $\mathcal{A}$ and $\mathcal{B}$ are bounded by the shape prior of TRELLIS and inherit errors of the part masks.
Over a sequence of close-ups, the regenerated parts can drift in appearance, since every update denoises their anchored latents again.
The open branch can also correct the joint of an unusual part only partly, such as a drop-down flap whose side hinge it moves to a tilted bottom hinge (\cref{fig:supp_failures}).
A stronger prior such as TRELLIS.2~\cite{xiang2026native}, native 2D--3D correspondences as in CUPID~\cite{huang2026cupid}, and more diverse opening ranges and hinges in RevealTrain would address these, and a segmentation model trained on articulated objects could give more reliable masks.
Keeping the anchored latents fixed during sampling or training on sequences of updates could prevent the drift over chained updates.
Procedural generators~\cite{joshi2026procedural} could add variants in RevealTrain, and few-step distillation~\cite{zhou2026few,lai2025unleashing} could make updates faster.

\FloatBarrier

\end{document}